\documentclass[journal]{vgtc}

\usepackage[export]{adjustbox}
\usepackage{algorithm}
\usepackage{algpseudocode}
\usepackage{amsfonts}
\usepackage{amsmath}
\usepackage{amssymb}
\usepackage{array}
\usepackage[english]{babel}
\usepackage{bm}
\usepackage{booktabs}
\usepackage{caption}
\usepackage{ccicons}
\usepackage{cite}
\usepackage{comment}
\usepackage{enumitem}
\usepackage{epstopdf}
\usepackage{float}
\usepackage{graphicx}
\usepackage{hyperref}
\usepackage[utf8]{inputenc}
\usepackage{marginnote}
\usepackage{mathptmx}
\usepackage{ microtype }
\usepackage{multicol}
\usepackage{multirow}
\usepackage{orcidlink}
\usepackage{pifont}
\usepackage{soul}
\usepackage{subcaption}
\usepackage{tabu}
\usepackage{tcolorbox}
\usepackage{textcomp}
\usepackage{tikz}
\usepackage{times}
\usepackage{todonotes}
\usepackage{wasysym}
\usepackage{wrapfig}
\usepackage{xspace}

\newcommand{\adv}{ADV\xspace}
\newcommand{\advmodel}{ADV\xspace}
\newcommand{\circledchar}[1]{ \protect\tikz[baseline=(char.base)]{
\protect\node[shape=circle, draw=none, fill=gray!20, inner sep=0pt, minimum size=1em] (char) {\textcolor{black}{#1}}; }}
\newcommand{\cmX}{\protect\tikz[baseline={(0,-0.5ex)}]{\definecolor{customXColor}{HTML}{B03F18}\protect\draw[line width=1.5pt, draw=customXColor] (-0.7ex, 0.7ex) -- (0.7ex, -0.7ex);
\protect\draw[line width=1.5pt, draw=customXColor] (-0.7ex, -0.7ex) -- (0.7ex, 0.7ex); }\xspace }
\newcommand{\cnnn}{CNNs\xspace}
\newcommand{\cnns}{CNN\xspace}
\newcommand{\cov}{COV\xspace}
\newcommand{\covmodel}{COV\xspace}
\newcommand{\customLabelZigzag}[1]{\protect\tikz[baseline={(0,-0.6ex)}]{\definecolor{customColor}{HTML}{#1}\protect\draw[line width=1pt, draw=customColor] (0, 0.6ex) -- (0.2, 0.6ex); \protect\draw[line width=1pt, draw=customColor] (0.2, 0.6ex) -- (0.2, -0.3ex); \protect\draw[line width=1pt, draw=customColor] (0.2, -0.3ex) -- (0.4, -0.3ex); }}
\newcommand{\cyanHorizontalBox}{\drawHollowRect{cyan}{0.8pt}{6pt}{3pt}}
\newcommand{\defhighlighter}[3][]{\tikzset{every highlighter/.style={color=#2, fill opacity=#3, #1}}}
\newcommand{\distanceName}{training-test distance\xspace}
\newcommand{\drawHollowRect}[4]{\protect\tikz{\protect\draw[#1, line width=#2] (0, 0) rectangle (#3, #4);}}
\newcommand{\eg}{e.g.}
\newcommand{\googleDriveLink}{\href{https://drive.google.com/drive/folders/19Ybm-wpsaY-dL5OM-Ee_mf7ntw3WpHCs?usp=sharing}{\texttt{Google Drive}}\xspace}
\newcommand{\grayCell}{\textcolor[HTML]{c8c8c8}{$\ssquare[1]$}\xspace}
\newcommand{\grayDiamond}{\textcolor[HTML]{9C9C9C}{$\sdiamond[0.6]$}\xspace}
\newcommand{\greenDiamond}{\textcolor[HTML]{328724}{$\sdiamond[0.8]$}\xspace}
\newcommand{\haehnX}{\protect\tikz[baseline={(0,-0.5ex)}]{\definecolor{customXColor}{HTML}{086373}\protect\draw[line width=1.5pt, draw=customXColor] (-0.7ex, 0.7ex) -- (0.7ex, -0.7ex);
\protect\draw[line width=1.5pt, draw=customXColor] (-0.7ex, -0.7ex) -- (0.7ex, 0.7ex); }\xspace }
\newcommand{\ie}{i.\,e.}
\newcommand{\iid}{IID\xspace}
\newcommand{\iidmodel}{IID\xspace}
\newcommand{\jc}[1]{\textcolor{jccolor}{#1}}
\newcommand{\jcc}[1]{\textcolor{jcccolor}{#1}}
\newcommand{\mini}{OOD\xspace}
\newcommand{\mytitle}{ A Rigorous Behavior Assessment of \cnnn Using \\ a Data-Domain Sampling Regime}
\newcommand{\mytitlea}{ A Rigorous Behavior Assessment of \cnnn Using a Data-Domain Sampling Regime}
\newcommand{\oodmodel}{OOD\xspace}
\newcommand{\osflink} {\href{https://osf.io/gfqc3} {\texttt{osf\discretionary{}{.}{.}io\discretionary{/}{}{/}gfqc3}}\xspace}
\newcommand{\purpleDiamond}{\textcolor[HTML]{BB94BE}{$\sdiamond[0.8]$}\xspace}
\newcommand{\redVerticalBox}{\hspace{1.5pt}\drawHollowRect{red}{0.8pt}{3pt}{6pt}\hspace{1.5pt}}
\newcommand{\resnet}{ResNet50\xspace}
\newcommand{\rvision}[1]{#1}
\newcommand{\sbullet}[1][.5]{\mathbin{\vcenter{\hbox{\scalebox{#1}{$\bullet$}}}}}
\newcommand{\sdiamond}[1][.5]{\mathbin{\vcenter{\hbox{\scalebox{#1}{\ding{117}}}}}}
\newcommand{\shumanaititle}{Human-AI comparisons\xspace}
\newcommand{\siititle}{Stability\xspace}
\newcommand{\sititle}{Sampling effect\xspace}
\newcommand{\sm}{Apdx.\xspace}
\newcommand{\ssquare}[1][.5]{\mathbin{\vcenter{\hbox{\scalebox{#1}{$\blacksquare$}}}}}
\newcommand{\talbotX}{\protect\tikz[baseline={(0,-0.5ex)}]{\definecolor{customXColor}{HTML}{B68328}\protect\draw[line width=1.5pt, draw=customXColor] (-0.7ex, 0.7ex) -- (0.7ex, -0.7ex);
\protect\draw[line width=1.5pt, draw=customXColor] (-0.7ex, -0.7ex) -- (0.7ex, 0.7ex); }\xspace }
\newcommand{\tochange}[1]{\textcolor{red}{#1}}
\newcommand{\vgg}{VGG19\xspace}

\definecolor{customXColor}{HTML}{086373}
\definecolor{jcccolor}{RGB}{100, 200, 150}
\definecolor{jccolor}{RGB}{200, 150, 100}
\definecolor{lcolor}{RGB}{255, 229, 204}
\definecolor{mycolor_blue}{RGB}{142, 182, 215}
\definecolor{mycolor_red}{RGB}{240, 129, 131}
\definecolor{sjcolor}{RGB}{79, 200, 51}
\definecolor{NavyBlue}{RGB}{8,111,189}

\onlineid{2070}

\vgtccategory{Research}

\vgtcpapertype{Theory/model}
\usetikzlibrary{calc}
\usetikzlibrary{decorations.pathmorphing, shapes.geometric, positioning}

\newif\ifdetail
\let\ifdetail\iffalse

\newif\ifjccom
\let\ifjccom\iffalse

\defhighlighter{yellow}{.5}

\DeclareRobustCommand*\highlight[1][]{\tikzset{this highlighter/.style={#1}}\SOUL@setup
\def\SOUL@preamble{\begin{tikzpicture}[overlay, remember picture]\ highlight@BeginHighlight \highlight@EndHighlight\end{tikzpicture}}

\def\SOUL@postamble{\begin{tikzpicture}[overlay, remember picture]\highlight@EndHighlight \highlight@DoHighlight\end{tikzpicture}}

\def\SOUL@everyhyphen{\discretionary{\SOUL@setkern\SOUL@hyphkern \SOUL@sethyphenchar \tikz[overlay, remember picture]{\highlight@EndHighlight ;}}{}{\SOUL@setkern\SOUL@charkern }}

\def\SOUL@everyexhyphen##1{\SOUL@setkern\SOUL@hyphkern \hbox{##1}\discretionary{\tikz[overlay, remember picture]{\highlight@EndHighlight ;}}{}{\SOUL@setkern\SOUL@charkern }}
\def\SOUL@everysyllable{\begin{tikzpicture}[overlay, remember picture]\path let \p0 = (begin highlight), \p1 = (0,0) in \pgfextra{\global}\highlight@previous=\y0 \global\highlight@current =\y1 \endpgfextra (0,0) ; \ifdim\highlight@current < \highlight@previous \highlight@DoHighlight \highlight@BeginHighlight \fi\end{tikzpicture}\the\SOUL@syllable \tikz[overlay, remember picture]{\highlight@EndHighlight ;}}\SOUL@ }
\makeatother

\setlist[itemize]{labelindent=0pt, leftmargin=*, itemindent=0pt, listparindent=\parindent}

\title{\mytitle}

\author{
\authororcid{Shuning~Jiang}{0000-0002-6706-2818},
\authororcid{Wei-Lun~Chao}{0000-0003-1269-7231},
\authororcid{Daniel~Haehn}{0000-0001-9144-3461},
\authororcid{Hanspeter~Pfister}{0000-0002-3620-2582},
\authororcid{Jian~Chen}{0000-0002-1599-0831} }

\authorfooter{
\item S. Jiang, W. Chao, and J. Chen are with the Department of Computer Science and Engineering, The Ohio State University, Columbus, OH, USA. E-mail: \{jiang.2126, chao.209, chen.8028\}@osu.edu.
\item D. Haehn is with the Department of Computer Science, University of Massachusetts Boston, Boston, MA, USA. E-mail: daniel.haehn@umb.edu.
\item Hanspeter Pfister is with SEAS, Harvard University, Cambridge, MA, USA. E-mail: pfister@seas.harvard.edu. }

\abstract{We present a data-domain sampling regime for quantifying \cnnn' graphic perception behaviors. This regime lets us evaluate \cnnn' ratio estimation ability in bar charts from three perspectives: \textit{sensitivity to training-test distribution discrepancies}, \textit{stability to limited samples}, and \textit{relative expertise to human observers}. After analyzing 16 million trials from 800 \cnns models and 6,825 trials from 113 human participants, we arrived at a simple and actionable conclusion: \cnnn can outperform humans and their biases simply depend on the \distanceName.
\rvision{We show evidence of this simple, elegant behavior of the machines when they interpret visualization images.} \osflink~provides registration, the code for our sampling regime, and experimental results. }


\teaser{
\centering
\includegraphics[width=\linewidth]{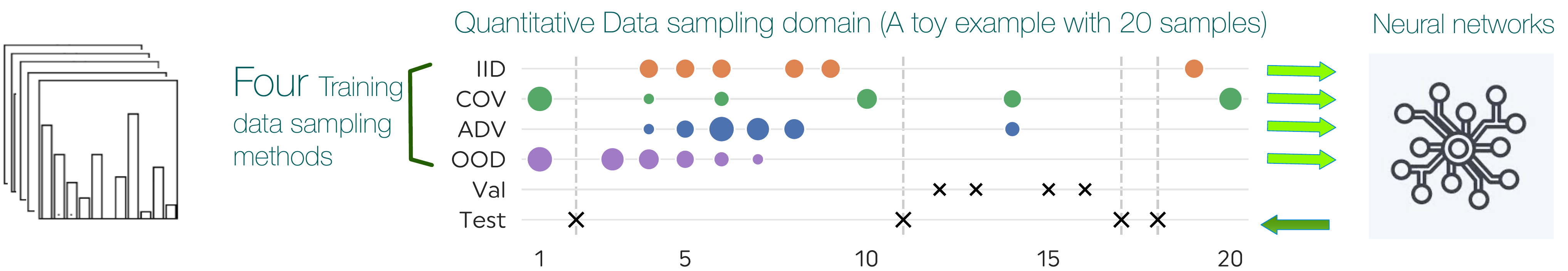}
\caption{\textbf{A toy example of a training data sampling regime.} We quantify behaviors of convolutional neural networks (\cnnn) in response to training inputs. Four data-domain sampling methods are used to supervise \cnnn before sending them to take the same test. }
\label{fig:teaser} }

\graphicspath{{figs/}{figures/}{pictures/}{images/}{./}}
\begin{document}
\firstsection{Introduction}
\label{sec:intro}

\maketitle

{A} crucial consideration when using machine learning models, such as a convolutional neural network (\cnns) or a large-language model, for visualization tasks is understanding when to trust their results and how to effectively train them~\cite{Hoque2024HarderBF}. Assessing their behavior quantitatively is essential for evaluating their reliability and identifying potential limitations.
\rvision{While most quantification studies emphasize algorithmic accuracy, highlighting vulnerability~\cite{haehn2018evaluating, wang2020cnn}, deficiencies in visual literacy~\cite{bendeck2024empirical, wang2024aligned, hong2025llms, Guo2023EvaluatingLL}, and appearance-based robustness (\eg, changes in background colors or line widths~\cite{Cui2024GeneralizationOC}), evaluations explicitly addressing data-domain issues remain rare~\cite{Yang2024ThinkingIS}. Haehn et al.~\cite{haehn2018evaluating} examined data-domain concerns, but their training, test, and validation datasets were restricted to the image ratios from Cleveland and McGill~\cite{cleveland1984graphical}, which are essentially the \textit{test} samples for \textit{human} observers. To rigorously evaluate both unsupervised and supervised algorithms, the training data must be carefully defined, as its selection can give observers advantages or disadvantages when tested and compared~\cite{chollet2019measure}. }

\rvision{Furthermore, the quantification of algorithmic performance can facilitate alignment with human observers' ability.} This is important as even for the most popular visualizations (\eg, bar charts and scatter plots), the ability to perform chart-reading tasks is considerably challenging to human observers who rely on the coordination of several mental processes, including rapid perceptual computations~\cite{cleveland1984graphical} with respect to a visualization schema~\cite{pinker2014theory}, ability to perform abstraction~\cite{munzner2014visualization}, working memory~\cite{Padilla2018DecisionMW}, domain knowledge~\cite{Wilkinson2012}, and interpretation processes~\cite{Tversky2011VisualizingT}. However, it is important to recognize that the behavior constraints of \cnnn differ from those of humans~\cite{geirhos2018imagenettrained}.

The goal of this work is to fill in the knowledge gaps to critically quantify supervised discriminative \cnnn by focusing on data-domain sampling. Especially, we study model behaviors by manipulating training-test data distribution shift
\rvision{to answer the following question:
\textit{How do sampled training data values influence the model behaviors on the test set?} Through multiple experiments, we assess algorithmic behaviors from several perspectives: models' robustness to out-of-distribution data samples (\textit{sensitivity}), resilience to limited input resources (\textit{stability}), and when they surpass human (\textit{relative expertise)}.}

\textbf{Method}. To systematically answer our research question and quantify \cnnn' behaviors in visualization contexts, we followed reliable tests used in human studies that focus on data targets such as bar heights~\cite{Bearfield2023WhatDT, talbot2014four} or ratios~\cite{cleveland1984graphical,heer2010crowdsourcing}. We then designed a training data sampling regime to enable direct behavior quantification during the test phase, after training models from shifting training-test data sampling distributions (see a toy example in~\autoref{fig:teaser}):
\textbf{(1) Independent and identical data distribution (\iid)}: simple random sampling from the same data domain as the test set, as the baseline.
\textbf{(2) Coverage (\cov)}: sampling values that are distant from each other in the \textit{training} set;
\textbf{(3) Adversarial (\adv)}: sampling values that are positioned away from the \textit{test} set values;
\textbf{(4) Out-of-distribution (\mini)}: sampling values concentrated in the left portion of the data domain. Through three controlled and two exploratory studies, we systematically quantify the reading behavior changes in response to these training data samples, after sending observers to take the ``\textit{same}'' tests in each of the multiple runs, so that behaviors are comparable.

\begin{figure}[!t]
    \centering
    \begin{subfigure}
        [T]{\columnwidth}
        \includegraphics[width=\columnwidth]{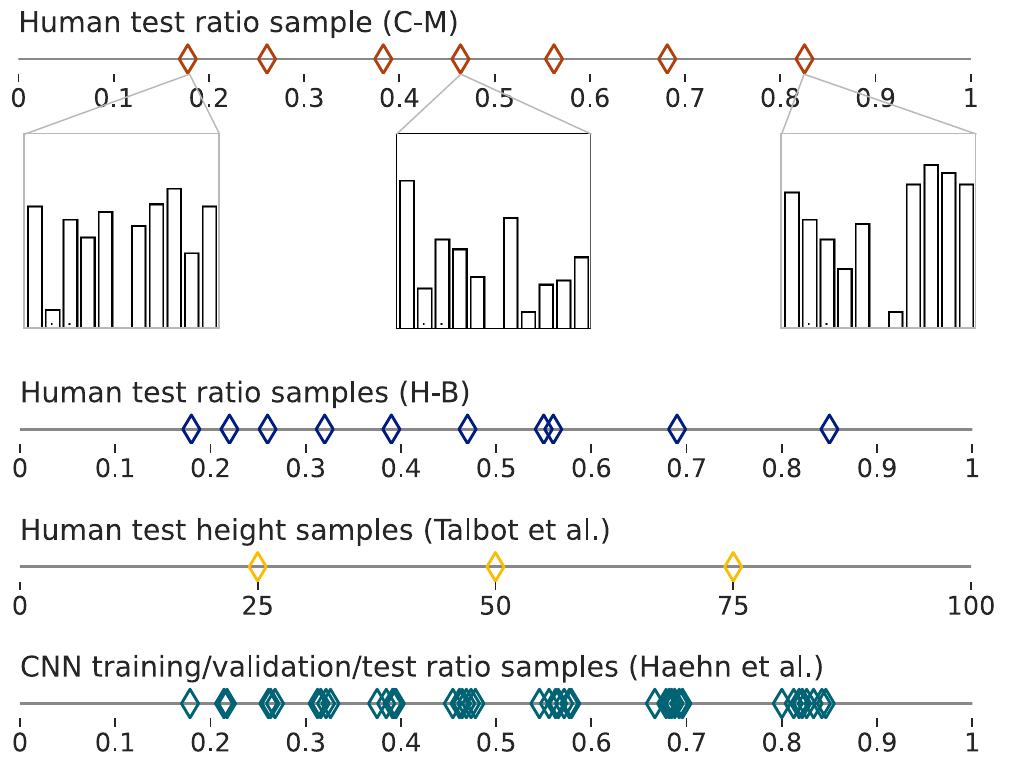}
    \end{subfigure}
    \caption{\rvision{\textbf{Data sampling configurations in published studies.} Human studies (\textit{$1^{st}$ to $3^{rd}$ rows}):
    Ratio or height samples as seen by human observers in Cleveland and McGill~\cite{cleveland1984graphical}, Heer and Bostock~\cite{heer2010crowdsourcing}, and Talbot et al.~\cite{talbot2014four}.         An algorithmic study (\textit{Last row}):
    Haehn et al.~\cite{haehn2018evaluating}     reused test samples in the human experiments as  the  data-domain to sample the training, test, and validation sets     for \cnnn.}
    }
    \vspace{-10pt}
    \label{fig:othersSamplingExample}
\end{figure}

\textbf{Results.} We have new findings to interpret \cnnn' ability to decode charts. First, \cnnn achieved higher accuracy and lower uncertainty with \iid and \cov in the ratio estimation task (\autoref{sec:study.outofdistribution}). Second, reducing the number of training samples increases both inference error and uncertainty, which explains the underperformed \cnnn in Hanehn et al.'s experiment~\cite{haehn2018evaluating}. Third, an unexpected finding is the benefits of the coverage sampling \cov: increasing training data span and evenly distributing samples within the span leads to higher stability than the most widely used \iid-supervised models, especially when the total number of unique values is limited. Fourth, we also show an adversarial perspective when the training data are in the same domain as \iid but introduced lower reliability. This means that \cnnn, when used in visualizations, can easily be fooled~(\autoref{sec:study.stability}). Finally, \cnnn have much lower errors and uncertainty than people in some of the testing conditions~(\autoref{sec:study.humanai}). These results emphasize the importance of formalizing the training sampling domain of quantitative data when evaluating \cnnn in visualizations.

Our work makes the following \textbf{contributions}: (1) A fundamental understanding of \cnnn' behaviors by focusing on \textit{quantitative} data sampling unique to the visualization context; (2) A reusable controlled study method for comparing observers, either humans or \cnnn. (3) New knowledge in \cnnn' behaviors through a set of experiments to quantify aspects of train-test discrepancies. We hope our research will both inspire new systematic AI behavior experiments in visualizations and support a cost-efficient interpretation strategy to understand complex algorithmic behaviors.
\section{Background and Related Work}
\label{sec:relatedwork}

\noindent This section reviews the work that inspired ours: chart perception experiments in visualizations and quantification studies of \cnnn.

\subsection{Sampling Methods for Quantifying Visual Perception}
\label{sec:relatedwork_sampling}
\noindent Sampling is crucial in evaluating both human and algorithmic observers particularly in visualization tasks such as chart reading. In human studies, controlled sampling helps isolate the factors that influence reading. Cleveland and McGill in their 1984 seminal work (C-M)~\cite{cleveland1984graphical} sampled 7 out of 9 ratios ($1^{st}$ row in~\autoref{fig:othersSamplingExample}) derived from 10 heights: 10, 12.1, 14.7, 17.8, 21.5, 26.1, 31.6, 38.3, 46.4, 56.2 to spaced ratio samples seen by human observers.
\rvision{These samples were spaced evenly on a logarithmic scale to represent the power-law distribution of human length perception~\cite{Householder1940WeberLT}.} This work has been replicated and extended to crowdsourcing studies by Heer and Bostock~\cite{heer2010crowdsourcing}, who chose a slightly different set of ratios ($2^{nd}$ row in~\autoref{fig:othersSamplingExample}) from C-M's original selection. Talbot et al.~\cite{talbot2014four} adopted C-M's ratios but instead sampled bar heights ($3^{rd}$ row in~\autoref{fig:othersSamplingExample}) corresponding to those ratios.

Progress toward machine systems that quantify human-like understanding of visualizations has incorporated
\rvision{training data distribution as these} human studies. For example, Haehn et al.\cite{haehn2018evaluating} rounded Cleveland and McGill’s original 10 heights to the nearest integers, yielding 43 unique ratios (last row in \autoref{fig:othersSamplingExample}), which were randomly allocated for model training, validation, and test.
\rvision{This method assessed model inference accuracy under a specific human test data preparation paradigm, which did not consider that humans and models may have distinct decision mechanisms~\cite {geirhos2018imagenettrained}. In this work, we explicitly separated training and test data to systematically assess algorithmic behaviors. This allows us to draw new conclusions and offer deeper insight into when and why \cnnn succeed. }

\begin{figure}
    \centering
    \begin{subfigure}
        {0.19\columnwidth}
        \centering
        \small{Type 1. \\Adjacent bars}\par
        \smallskip
        \includegraphics[width=\columnwidth]{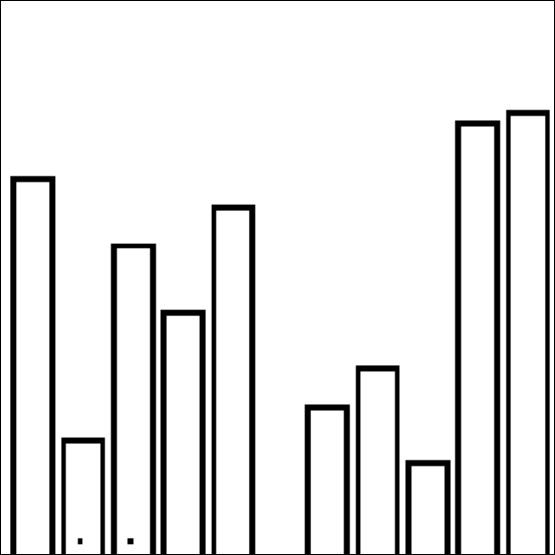}
    \end{subfigure}
    \begin{subfigure}
        {0.19\columnwidth}
        \centering
        \small{Type 2. \\Aligned stacked bars}\par
        \smallskip
        \includegraphics[width=\columnwidth]{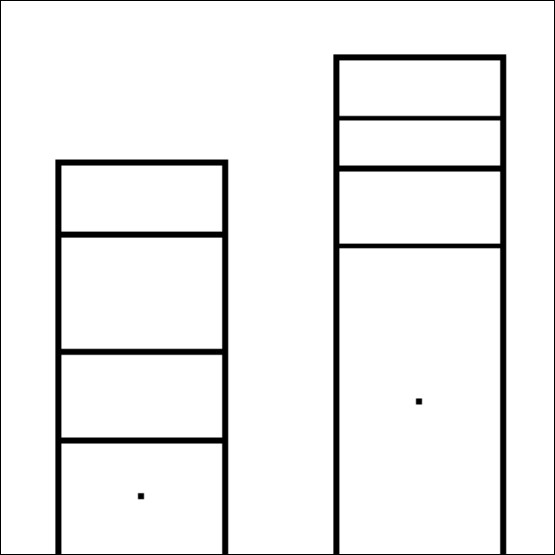}
    \end{subfigure}
    \begin{subfigure}
        {0.19\columnwidth}
        \centering
        \small{Type 3.\\ Separated bars}\par
        \smallskip
        \includegraphics[width=\columnwidth]{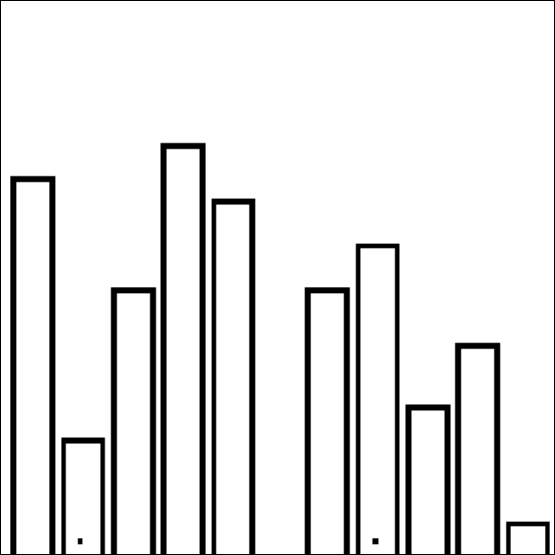}
    \end{subfigure}
    \begin{subfigure}
        {0.19\columnwidth}
        \centering
        \small{Type 4. \\Unaligned stacked bars}\par
        \smallskip
        \includegraphics[width=\columnwidth]{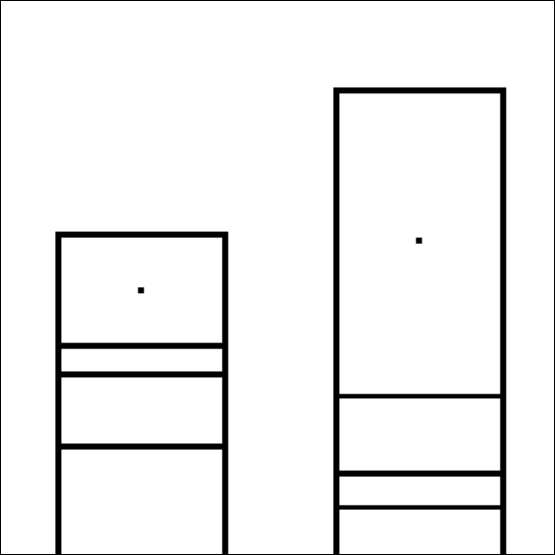}
    \end{subfigure}
    \begin{subfigure}
        {0.19\columnwidth}
        \centering
        \small{Type 5.\\ Divided bars}\par
        \smallskip
        \includegraphics[width=\columnwidth]{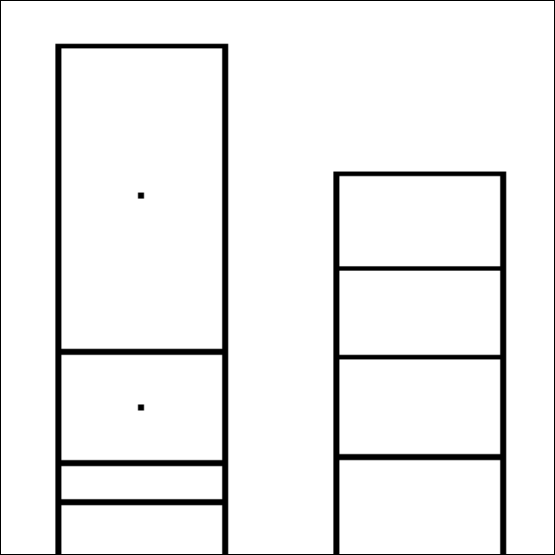}
    \end{subfigure}
    \caption{Our study reused the five bar types from Cleveland and McGill~\cite{cleveland1984graphical}.}
    \vspace{-10px}
    \label{fig:fiveTypesExample}
\end{figure}

\subsection{Understanding Machine Learning Algorithm Behaviors}

\noindent Our work is related to understanding the behavior of \cnnn and aligns with three broad research directions. The first method concerns \textit{explainable} approaches that introduce additional computational models~\cite{lake2015deep} or visualizations~\cite{tzeng2005opening, zintgraf2017visualizing,zhou2016learning,strobelt2017lstmvis,liu2016towards} to interpret black-box behaviors. We avoided introducing new models, as this process itself is error-prone~\cite{rudin2019stop}. We also avoided visualization-based explainable methods (\eg,~\cite{wang2020cnn} and also see the excellent review~\cite{liu2017towards,Choo2018VisualAF,Wang2023VisualAF}), as they are not quantitative~\cite{Borowski2021ExemplaryNI} and may not be consistently perceived by human observers~\cite{geirhos2023don}. The second method involves human-algorithm alignment, where algorithms are used to explain human decision-making. This approach has only recently led to significant advances, offering a broad, global understanding of observers' decisions~\cite{conwell2024large, mistry2023learning}, primarily for categorical data in computer vision. The third method aims to trace algorithmic responses back to their training data by analyzing \cnnn' characteristics to interpret model behaviors~\cite{Geirhos2020ShortcutLI,Wichmann2023AreDN,Borowski2021ExemplaryNI}. This method has also been used in visualizations by posing different questions to an algorithm and observing its responses to interpret its behavior~\cite{bendeck2024empirical, wang2024aligned, hong2025llms, Guo2023EvaluatingLL}.

In our work, we adopt the third approach to examine \cnnn' behaviors by probing their outputs through controlled alterations in input distributions. By analyzing the outputs in response to different training inputs, we aim to identify optimal strategies for training \cnns models in a visualization context. To our knowledge, our method is the first attribution experiment designed to quantify \cnns observers' behaviors.

\begin{figure*}[!t]
    \centering
    \includegraphics[width=\textwidth]{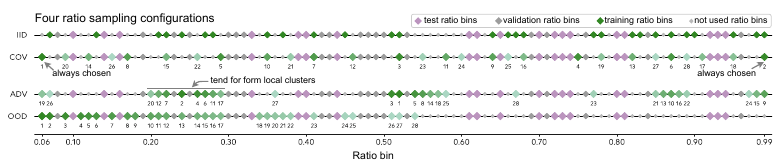}
    \caption{\textbf{The ratio bin data sampling configurations in our sampling regime.} The axis represents 94 unique ratio bins $\in[0.06, 0.99]$. The numbers below the training samples \greenDiamond and the shading show the order in which the samples are chosen for each of the sampling methods. The smaller the number (or the darker the \greenDiamond color), the earlier a ratio bin is chosen.
    \textbf{\ul{Observations.}} For the training samples
    \greenDiamond, \cov bins are spread out;
    \adv bins form clusters;
    \mini bins do not fully represent the test distribution. (\sm ~\autoref{fig:robusness3RunsCNN} has the full list of all runs for ratio sampling.) }
    \label{fig:ratioHeightSampling}
\end{figure*}

\subsection{The Cost of Data Diversity}
\noindent Producing large training sets is often a slow and data wrangling acquisition, and cleaning can be prohibitively expensive~\cite{kandel2011wrangler}. Therefore, we examine the reliability of the model under resource constraints to evaluate how models perform with varying numbers of unique inputs. An adversarial condition in vision is defined as ``carefully crafted, seemingly normal inputs that have been slightly modified with imperceptible perturbations, causing the network to misclassify the image''~\cite{kurakin2018adversarial, goodfellow2014explaining}. We define adversarial cases in the quantitative visualization domain as the training samples that share the same domain as test samples but may impair \cnnn' graphical perception ability. We leveraged an adversarial condition where the training samples are further from the \textit{test} samples, and can limit the reliability of the models.
\rvision{ }

\section{A Sampling Regime: Measuring Model Response to Training-Test Sampling Distribution Variations}
\label{sec:samplingRegime}

\noindent This section presents the first contribution of this work: a training data sampling regime that quantifies model behavior under varying training-test discrepancies in the ratio or height data sampling domain.

\subsection{The Problem Domain: Tasks and Visualization}

\noindent We used a ratio estimation task, ``\textit{judge what percent the smaller is of the larger}''~\cite{cleveland1984graphical}, because it required CNNs to reason about spatial relationships and had the highest inference errors among all tasks~\cite{haehn2018evaluating}.

\textbf{Ratio estimation} task uses five bar chart types shown in \autoref{fig:fiveTypesExample}. Types 1--3 are bottom-aligned and thus judgment is by perceiving \textit{position} along the common scale; and Types 4 and 5 must be compared by perceiving different \textit{bar lengths}. Two target bars in each image are marked by a black dot and are adjacent to each other in all but Type 3. The ratio to be estimated is calculated as $\mathrm{ratio}=\mathrm{shorter}\; (h)/\mathrm{taller\;bar\;height}(H)$. In the human studies of position-length~\cite{cleveland1984graphical,heer2010crowdsourcing,talbot2014four}, people preferred position over length, and Types 1--3 outperformed Types 4--5.

\subsection{The Target Data Sampling Domains}
\label{sec:samplingRegimeTasks}

\noindent In our experiment, we used either the ratio or the taller bar height as the sampling target. \autoref{tab:datarange} shows the range and count in the data domain.

\begin{figure*}
    \centering
    \includegraphics[width=\textwidth]{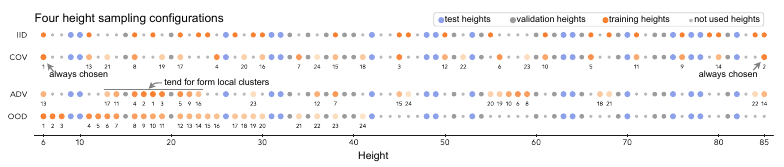}
    \vspace{-10px}
    \caption{\textbf{The bar height data sampling configurations in our sampling regime.} The axis represents 80 unique taller bar heights $\in[6, 85]$. The numbers below the training samples \textcolor[HTML]{F67F3B}{$\sbullet[1]$} and the shading show the order in which the samples are chosen for each of the sampling methods. The smaller the number (or the darker the \textcolor[HTML]{F67F3B}{$\sbullet[1]$} color), the earlier a height is chosen.
    \textbf{\ul{Observations.}} For the training set \textcolor[HTML]{F67F3B}{$\sbullet[1]$}, \cov samples are spread out. \adv tends to form local clusters, and \mini partially represents the test distribution. (\sm~\autoref{fig:robusness3RunsCNN} has the full list of all runs for height sampling.) }
    \vspace{-10px}
    \label{fig:heightSamplingSM}
\end{figure*}

Bar heights are integers: the taller bar height $H$ ranges from 6 to 85 (80 values), and the shorter bar height $h$ ranges from 5 to 84 (80 values), resulting in 3,240 unique ($h, H$) pairs. From these pairs, 2,081 unique ratios were derived, spanning from 5/85=0.059 to 84/85=0.988. We grouped these ratios into 94 evenly spaced bins ($rb$): $rb_{1}:[0.055, 0.065)$, $rb_{2}:[0.065, 0.075)$, ..., and $rb_{94}:[0.985, 0.995)$. For simplicity, each ratio bin is referred to by its midpoint value, such that $rb{_1}: 0.06$, $rb{_2}: 0.07$, ..., $rb{_{94}}: 0.99$, with $rb\in[0.06, 0.99]$. For Type 5, where two target bars are stacked on the same side, the constraint $h+H\le90$ reduced the sampling domain slightly, limiting the ratio bins to $rb \in[0.06, 0.98]$.
\autoref{fig:ratioHeightSampling} and \autoref{fig:heightSamplingSM} illustrate the ratio samples $\sdiamond[0.8]$ and height samples $\sbullet[1]$ along the axis, respectively.

\begin{table}[!t]
    \small
    \centering
    \caption{{Data range and the total number of unique samples.}
    }
    \vspace{-10pt}
    \label{tab:datarange}
    \resizebox{\columnwidth}{!}{\addtolength{\tabcolsep}{-0.4em}
    \begin{tabular}{llll}\\\toprule
        \textbf{Ratio bin range}               & Types 1-4: &[0.06, 0.99] &(94) \\
        (\textbf{number of ratio bins})          & Type 5: &[0.06, 0.98] &(93) \\
        \midrule
        \textbf{Taller bar height range} &
        Types 1-5 &[6, 85]& (80)  \\
        \textbf{(number of height samples)}\ \ \
        & \\
        \bottomrule
    \end{tabular} }
    \vspace{-10pt}
\end{table}

\subsection{Problem Framing and Sampling Methods}
\label{sec:samplingRegimeMethod}

\noindent
\paragraph{Framing ratio estimation task for \cnnn.} We frame the ratio estimation task as training a predictive model $F$ that can take an input image $x$ and output a predicted ratio $\hat{y}$.

\paragraph{Design principles.} We carefully controlled evaluation conditions, assessing a \cnns's ability and behavior by its responses to different sampling inputs by clearly defining the data domain. Given that models (observers) were compared in our experiments, we ensured a fair comparison by configuring conditions so that all observers took the same tests. In other words, test and validation sets were randomly chosen and held constant across different training conditions to ensure that models were evaluated using the same dataset for comparability. This control is crucial, as Cleveland and McGill~\cite{cleveland1984graphical} observed in their position-length experiment that human performance differed when judging ratios in different ratio bins.
\rvision{Because our focus is on data sampling, we keep appearance-based variables, such as line width, color, and gaps between bars, constant (See~\sm \autoref{fig:bar_appearance} for values used for these variables).} The detailed steps below use ratio sampling as an example.

\paragraph{\textcolor{cyan}{Step 1}: Test and Validation Samples.}
\label{sec:step1} We selected $94\times 20\%$ = 19 ratio bins for each of the test \purpleDiamond and validation \grayDiamond sets.

\paragraph{\textcolor{cyan}{Step 2}: Selection of Training Data Samples.} \label{sec:step2} We assembled four sampling methods, each of which subsampled $m=28$ ratios from 56 ratio bins, \ie, half of the remaining 60\% ratio bins. Here, $m$ could be set to any other numerical proportion of $56$. We chose it for simplicity.

\begin{itemize}
    \setlength\itemsep{0em}
    \item \textbf{IID}: The \textit{baseline} sampling method used to train \iidmodel. Training \greenDiamond and test \purpleDiamond samples follow the same simple random sampling.

    \item \textbf{{\cov}}: The \textit{distant-to-train} condition used to prepare data to train \covmodel. Samples were chosen sequentially by selecting $rb$ \greenDiamond that was the most distant from those that had been chosen so far. Thus, the first two $rb$ selected were $rb_{1}$ and $rb_{94}$, the minimum and maximum of available ratio bins.
    The third $rb$ was the most distant ratio bin from $rb_{1}$ and $rb_{94}$, and was close to the center of the sampling domain. This process continued by selecting the next $rb$ as the most distant ratio bin from all previously selected ratio bins, until $m=28$ ratio bins were selected. In a sense, this sampling method encouraged the model to utilize the full data domain whenever possible.

    \item \textbf{{\adv}}: The \textit{distant-to-test} condition used to prepare training data to supervise \advmodel.
    The training samples \greenDiamond were as distant from the \textit{test} samples \purpleDiamond as possible. This is aimed to evaluate an adversarial condition, when the same number of samples as \iid and \cov were chosen from the same domain, thus \textit{looked similar}, but distanced themselves from the test set.
    We can see that these samples \greenDiamond form local clusters, as adjacent points had similar distances to the test set.

    \item \textbf{\mini}:
    The \textit{out-of-distribution} condition used to train \oodmodel.
    We sorted all available 56 ratio bins in ascending order, and used the smallest 28 ratio bins to assemble a case of a biased training set.
    We called this OOD because the training set shifted to a different distribution from the test.
    \ifdetail \tochange{This is a specific choice for OOD. Why did you make it, and what other alternatives might you consider? \jcc{I think an interesting one would be to take the other end to see if the result was symmetrical. we did additional exp.}}
    \fi
\end{itemize}
\ifdetail \tochange{This is repetitive and should go a lot earlier to explain the basic setup. \jcc{condensed now}}
\fi \ifdetail \tochange{This sentence probably should go to the beginning of this section. Overall, this section needs to be rewritten for better clarity, including justifications of the parameter choices.\jcc{I moved it up.}}
\fi

\paragraph{\textcolor{cyan}{Step 3}: Behaviors Under Limited Resources.}
\label{sec:limitedresources} Given this sampling configuration, we evaluated \textit{stability under limited resources} by reducing the number of training samples by half each time. Starting with $30\%$ of available ratio bins for training, we reduced this proportion to $15\%$, $7.5\%$, and $3.75\%$, resulting in 14, 7, or 3 ratio bins, following the labeled order along the axis in \autoref{fig:ratioHeightSampling}.

\paragraph{Rendering Images.} To generate an image, we first randomly select an $rb$ in the training/validation/test set and then randomly select a ($h,H$) pair that yields a ratio within the selected bin, where $h$ and $H$ correspond to the shorter and taller target bar heights, respectively. For example, if $rb=0.42$ was selected, both (19, 45) and (16, 38) are valid options for rendering the image. See \sm~\autoref{fig:ratioHeightPairs} for the full set of ($h,H$) pairs for each ratio bin. The heights of other non-target bars in the image are assigned randomly. See \sm\autoref{fig:morebars} for examples of rendered images. We generated 60k/20k/20k images for training, validation, and test sets, respectively.

\ifdetail \tochange{Either explain all of the experimental setup, including parameter ranges, up front. Or do it for each study separately. Do not mix these organizational approaches. \jc{Fixed is the current version okay?}}
\fi

\paragraph{Model Configuration.} We used \vgg~\cite{simonyan2015a} and \resnet~\cite{He2015DeepRL}, conducting five runs for each sampling configuration to simulate the process as if five individuals were making multiple independent decisions. The images and their corresponding ground truth ratios served as inputs to supervise the
\cnnn. Since the test set remained consistent across sampling methods, the model inferences were directly comparable. (See \autoref{sec:modelConfig} for detailed model hyperparameter configurations.)

\section{Summary of Studies} The following three sections detail our experimental methodology and findings. The studies are registered at \osflink. \begin{itemize}
    \setlength\itemsep{0em}
    \item \textbf{Study I: \sititle (\autoref{sec:study.outofdistribution}):}
    This study answers the question: \textit{How sensitive are \cnnn to out-of-distribution samples?}
    We controlled the sampling method, bar chart type, and CNN model architecture, evaluating performance using the mean absolute error.
    Our findings indicate that \cnnn could achieve considerable accuracy, but were not robust to the distribution shift. Additionally, \cnnn' preference between position and length was minimal.

    \item \textbf{Study II. \siititle (\autoref{sec:study.stability}):}
    This study answers the question: \textit{How resilient are \cnnn to limited input resources?}
    We controlled the sampling method and downsampling level, assessing models based on mean absolute error and models' intra-consistency.
    We found that downsampling the training set degraded model inference outcomes, and \cov showed unexpected advantages that led to higher accuracy and reliability than \iid.

    \item \textbf{Study III. \shumanaititle (\autoref{sec:study.humanai}):}
    This study answers the question: \textit{Can \cnnn outperform humans?}
    We controlled the observer (humans vs. CNNs) and bar chart type. Model performance was assessed using the midmean log absolute error for comparisons with Cleveland and McGill~\cite{cleveland1984graphical} and the mean absolute error otherwise. We also measure the human-\cnns inter-consistency.
    Results show that \cnnn had much lower errors and uncertainty than humans, and the errors of \cnnn were also different from humans'.
\end{itemize}

\section{Study I: Model Robustness to Distribution Shift}
\label{sec:study.outofdistribution}

\noindent This section reports the second contribution of this paper: answering \textit{how models respond when training and test sets belong to different distributions.} We conducted two experiments, each involving either ratio or bar height sampling, by varying three independent variables one at a time: the sampling method (4 levels), the visualization type (5 levels), and \cnnn (2 levels) in a full-factorial design. The configuration led to $4\times5\times2\times 5~(\mathrm{runs})=200$ runs in each experiment.

\subsection{Design}

\subsubsection{Training Data Sampling}
\paragraph{Exp 1 (ratio sampling robustness)} sampled the ratio bins as described in~\autoref{sec:samplingRegimeMethod}, \autoref{fig:ratioHeightSampling}.
\paragraph{Exp 2 (height sampling robustness)} applied the same sampling process to the 80 unique heights~(\autoref{fig:heightSamplingSM}) as was used for ratio sampling, where 24/16/16 unique heights were in the training/validation/test sets applying the same ratio sampling process to the 80 bar heights.

\subsubsection{Model Robustness Hypotheses}
\label{sec:exp1hypo}
\noindent We had the following working hypotheses:

\begin{itemize}
    \setlength\itemsep{0em}
    \item \textbf{{{\sititle}}.H1} (\textit{Accuracy and Adversarial hypotheses}).
    Models will achieve high accuracy with \iid and \cov sampling, but not with \adv and \mini sampling.

    \item \textbf{{{\sititle}}.H2} (\textit{Out-of-distribution hypothesis}).
    Models trained in \mini will have reduced accuracy in regions where the training and test sets do not overlap. \ifdetail \tochange{It is not clear why this hypothesis is justified. Explain.}
    \fi

    \item \textbf{{{\sititle}}.H3} (\textit{Chart type hypothesis}):
    We will not observe the effect of visualization types on inference accuracy.
    \ifdetail \tochange{This is a strange hypothesis. What are you assuming here?}
    \fi
\end{itemize}

Several reasons led to these hypotheses. Both \iid and \cov encouraged diversity, with \iid being commonly used due to its ability to produce accurate inferences. \cov also increased diversity by enforcing a greater separation between training samples. We did not expect \adv and \mini to perform well on average, as \adv represented an adversarial case, and \mini involved a distribution shift in the sampling domain. For the same reason, in H2, we did not expect accuracy gains for test values that were absent from the training data domain. H3 could be supported because \cnnn were largely affected by data sampling input, rather than human perception limits.

\subsection{Analysis Method}
\label{sec:anal_method}

\noindent Each \cnns model produced 20K inferences. Given the large number of inferences, we made several decisions when analyzing the data. First, we aggregated predictions by ($h,H$) pairs, treating each aggregation as an independent trial for downstream statistical analysis. This was necessary to avoid false discoveries resulting from the large number of predictions. Second, the error metric was mean absolute error (MAE), defined as $MAE =\lvert\text{model\;inference}\; \hat{y}-\text{ground\;truth}\;y\rvert$, consistent with the evaluation metric used in human observer experiments by Talbot et al.~\cite{talbot2014four}. We did not use the logarithmic mean as in the C-M study~\cite{cleveland1984graphical} in this and the next sections, because taking the logarithm could disproportionately emphasize smaller errors. We also did not use mid-mean to trim $50\%$ data in these first two studies for practical considerations: we were interested in \cnnn' behaviors in real-world uses, where determining when to trim data may not always be feasible.

We examined the main effect of independent variables on prediction errors and performed statistical analysis using the General Linear Model (GLM). When a significant main effect was observed, we performed a post-hoc analysis using Tukey's Honest Significant Difference (HSD) test to assess the significance of differences across the levels in a treatment condition. Finally, we calculated the effect size using Cohen’s $\eta^{2}$~\cite{Cohen1969StatisticalPA} to measure the practical significance: ``no effect'' ($\eta^{2}<0.01$), ``small effect'' ($\eta^{2}\in [0.01, 0.06)$), ``\textit{medium effect}'' (\textit{$\eta^{2}\in [0.06, 0.14)$}), and ``\textbf{large effect}'' ($\eta^{2}\ge 0.14$).

\subsection{Results}
\label{sec:s1result}

\begin{figure}[!t]
    \centering
    \begin{subfigure}
        [T]{\columnwidth}
        \includegraphics[width=\columnwidth]{
        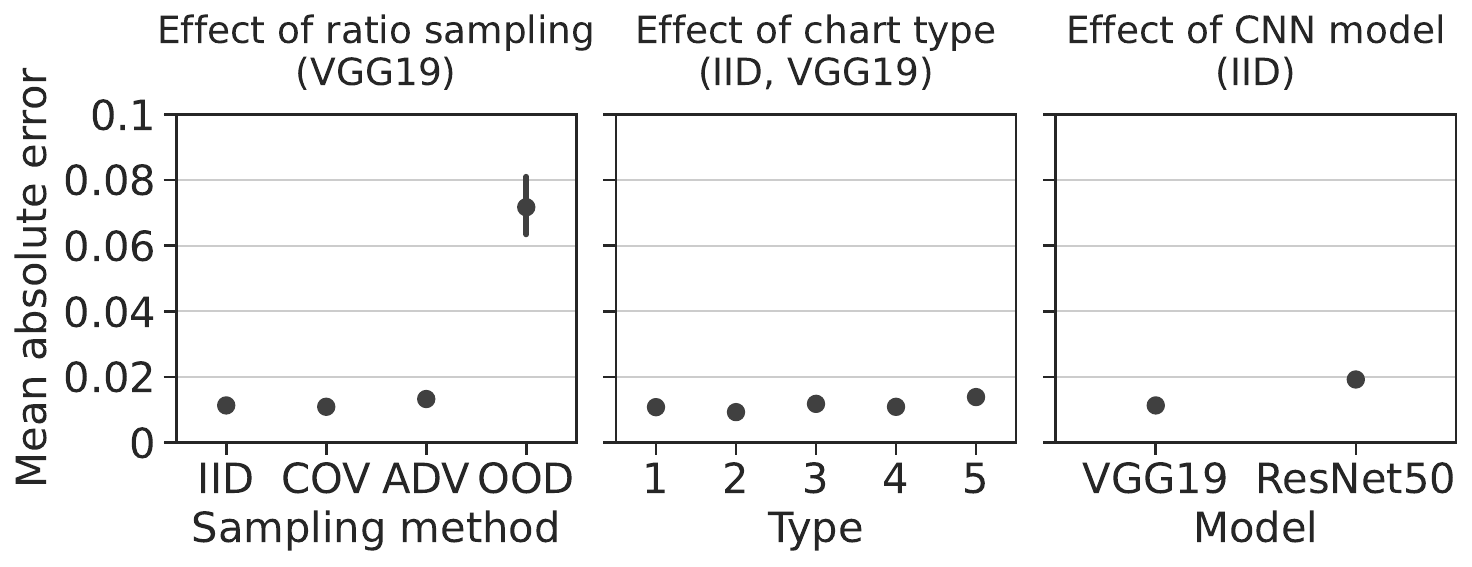
        }
        \put(-230, 5){\footnotesize \circledchar{a}} \put(-145, 5){\footnotesize \circledchar{b}}
        \put(-70, 5){\footnotesize \circledchar{c}}
    \end{subfigure}

    \begin{subfigure}
        [T]{\columnwidth}
        \includegraphics[width=\columnwidth]{
        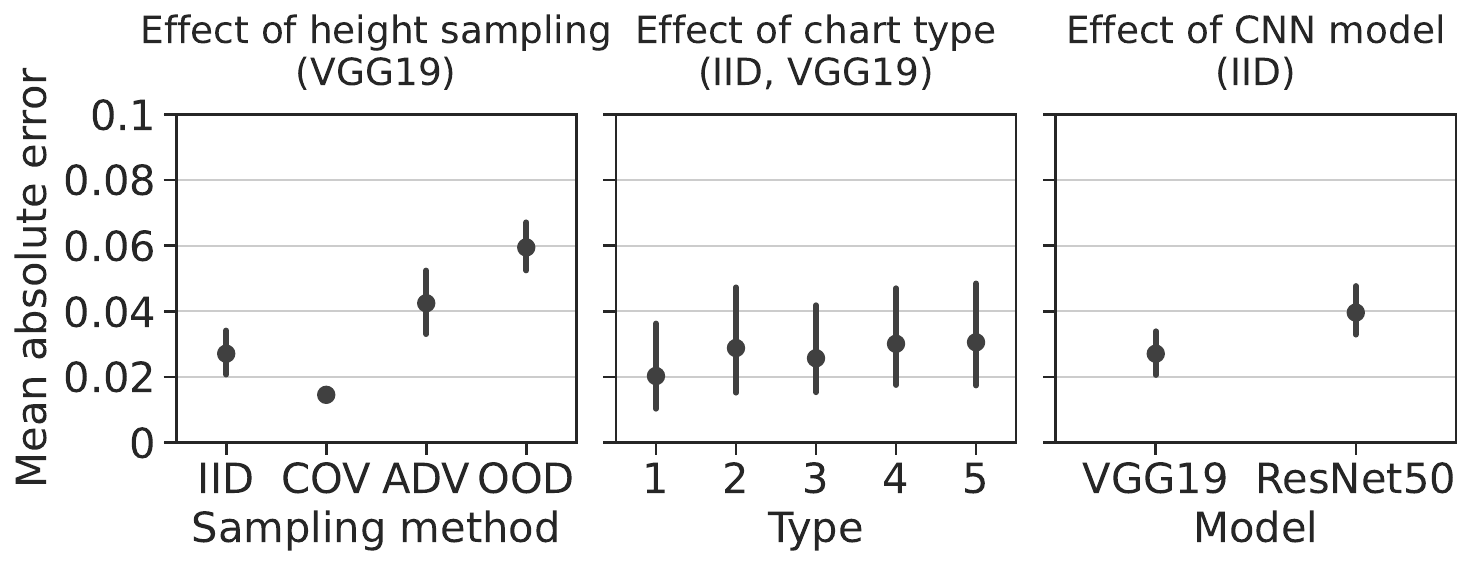
        }
        \put(-230, 5){\footnotesize \circledchar{d}} \put(-145, 5){\footnotesize \circledchar{e}}
        \put(-70, 5){\footnotesize \circledchar{f}}
        \label{fig:height_sampling_overall_summary}
    \end{subfigure}
    \caption{\textbf{\sititle.} Effect of sampling method, chart type, and
    \cnns model. \circledchar{a}-\circledchar{c}. Ratio sampling experimental results. \circledchar{d}-\circledchar{f}. height sampling experimental results. Error bars show $95\%$ confidence intervals.
    \ifdetail \tochange{the chart title should be the effect of sampling method. The sampling target (ratio sampling or height sampling) goes to (). The same issue with all other figures.}
    \fi
    \textbf{\ul{Observations.}}
    (1) \circledchar{a} and \circledchar{d}, \mini had much larger mean inference errors than other sampling methods.
    (2) \circledchar{b} and \circledchar{e}, chart types followed the general trend of humans, where Type 1 had smaller errors
    than Type 5 in ratio sampling.
    (3) \circledchar{c} and \circledchar{f}, \vgg had smaller errors than \resnet. }
    \label{fig:samplingMainEffects}
\end{figure}

\begin{table}[!t]
    \caption{\textbf{\sititle.summary statistics.} $a>b$ means $a$ is statistically
    more accurate than $b$.
    Medium effect size
    ($\eta^{2}\ge0.06$) is in \textit{italic}, and the large one ($\eta^{2}\ge0.14$) is in \textbf{bold}.
    \ifdetail \tochange{I don't know what a Tukey group is. You need to explain and justify your statistical metrics and evaluation in the main body of the text. Also, the text in the tables is impossible to read for me. Choose larger font sizes.}
    \fi }
    \label{tab:study1_stat_main} \resizebox{\columnwidth}{!}{    \addtolength{\tabcolsep}{-0.4em}
    \begin{tabular}{llrrl}
        \hline
        Variable                                 & \multicolumn{1}{c}{$F$} & \multicolumn{1}{c}{$p$} & \multicolumn{1}{c}{$\eta^2$} & \multicolumn{1}{c}{HSD}               \\ \hline
        \multicolumn{5}{l}{Study I Exp 1: Ratio sampling}                                                                                                                   \\
        \textbf{Sampling (\vgg)}  & $F_{(3, 2048)} = 174.6$ & $< \mathbf{0.001}$      & $\mathbf{0.20}$              & $\mathrm{(COV, IID, ADV)>(OOD)}$      \\
        \textbf{Type (IID, \vgg)} & $F_{(4, 509)} = 9.0$    & $< \mathbf{0.001}$      & $\mathit{0.07}$              & $\mathrm{(1, 2, 4)>(1, 3, 4)>(3, 5)}$ \\
        \textbf{Model (IID)}                     & $F_{(1, 1022)} = 337.1$ & $< \mathbf{0.001}$      & $\mathbf{0.23}$              & $\mathrm{(\vgg)>(\resnet)}$           \\ \hline
        \multicolumn{5}{l}{Study I Exp 2: Height sampling}                                                                                                                  \\
        \textbf{Sampling (\vgg)}  & $F_{(3, 1591)} = 29.4$  & $< \mathbf{0.001}$      & $0.05$                       & $\mathrm{(COV, IID)>(ADV)>(OOD)}$     \\
        Type (IID, \vgg)          & $F_{(4, 395)} = 0.3$    & $0.882$                 & $< 0.01$                     &                                       \\
        \textbf{Model (IID)}                     & $F_{(1, 794)} = 5.7$    & $\mathbf{0.017}$        & $<0.01$                      & $\mathrm{(\vgg)>(\resnet)}$           \\ \hline
    \end{tabular} }
\end{table}

\begin{figure}[!t]
    \hspace{-5px}
    \begin{subfigure}[T]{0.49\columnwidth}
        \includegraphics[height=135pt]{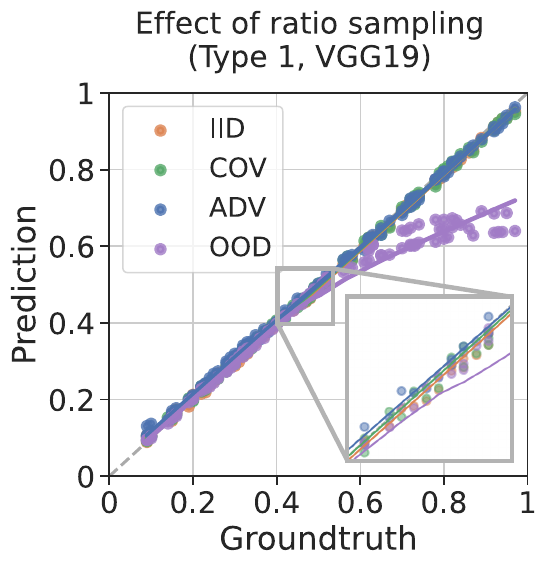}
        \put(-117, 5){\footnotesize \circledchar{a}}
    \end{subfigure}
    \begin{subfigure}[T]{0.49\columnwidth}
        \includegraphics[height=135pt]{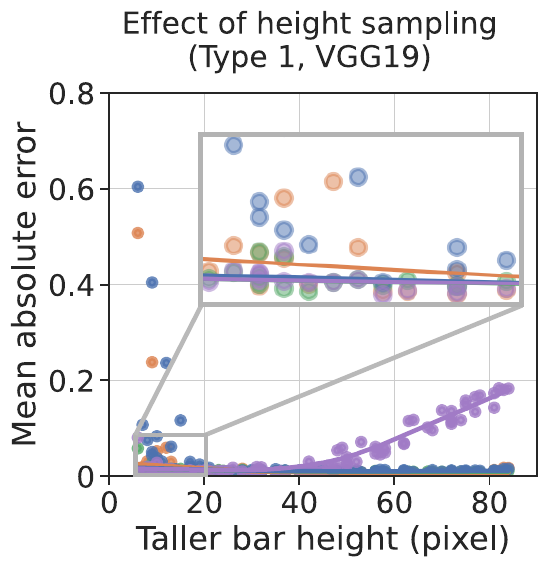}
        \put(-117, 5){\footnotesize \circledchar{b}}
    \end{subfigure}
    \caption{\textbf{\sititle.} Case-by-case analysis for Type 1 charts running \vgg.
    Regression lines show locally weighted scatterplot smoothing (LOWESS)~\cite{Cleveland1979RobustLW}.
    }
    \ifdetail
    \jc{JC. explain what lowest regression line.}
    \fi
    \vspace{-10px}
    \label{fig:distributionShiftDetail}
\end{figure}

\noindent
\textbf{Overview.} We collected a total of 4 million predictions from 200 models in each experiment ($200\; \mathrm{models}\times 20\mathrm{K}\; \mathrm{predictions}$).
\autoref{fig:samplingMainEffects} and \autoref{tab:study1_stat_main} show the mean errors and the summary statistics. See additional analyses in \sm~\autoref{fig:sm_sampling} and \sm~\autoref{tab:sm_study1_stat}.
\textbf{Sampling method is a significant main effect.}
\textbf{\sititle.H1 was supported for height sampling but not for the ratio sampling experiment.} For the ratio sampling experiment, \adv (0.013) was in the same Tukey group as \cov (0.011) and \iid (0.011)~(\autoref{fig:samplingMainEffects}\circledchar{a}), meaning that we did not observe a statistical difference among them. One reason for the lack of significance for \adv could be that it contained sufficient unique samples across the data domain.

For height sampling,
\cov (0.015) and \iid (0.027) had lower errors than \adv (0.042) and \mini (0.059)~(\autoref{fig:samplingMainEffects}\circledchar{d}). Now \adv was in a different Tukey group (\autoref{tab:study1_stat_main}). We suspected that the differences between \iid, \cov, and \adv conditions were caused by errors introduced by the smaller bar heights in~\autoref{fig:distributionShiftDetail}.

\textbf{Striking generalization failure on \mini. \sititle.H2 was supported}. To study the \cnnn' behaviors, we analyzed case-by-case \cnnn' inference accuracy by ratio bins (\autoref{fig:distributionShiftDetail}\circledchar{a}) and by heights (\autoref{fig:distributionShiftDetail}\circledchar{b}) computed from the two experiments correspondingly. In
\autoref{fig:distributionShiftDetail}\circledchar{a}, the diagonal line means that the prediction is a perfect match to the ground truth, and the closer to this line, the smaller the inference errors. \autoref{fig:distributionShiftDetail}\circledchar{b} shows the mean absolute errors against bar heights, lower error indicates better model performance. We observe that knee-shaped curves emerge in the out-of-train regions in \mini in both ratio sampling and height sampling experiments, showing errors that increase almost linearly. Interestingly, the in-train test data points in \oodmodel maintain low errors (0.008 for ratio sampling and 0.025 for height sampling). This result may support that \cnnn' inference accuracy depends on the coverage of input:
\iid did not guarantee sampling of the smallest or the largest values compared to \cov and the minimum values of \mini. Our experiments in the next section validate this conjecture by stress-testing \cnnn with limited training samples.
\ifdetail \tochange{Explain the difference to the results in Cleveland/McGill in more detail. This would be interesting.}
\fi

\textbf{\cnnn had limited ability to read smaller bar heights.}
\ifdetail \tochange{Move this sentence up front where you describe all of the experimental setup and parameters.}
\fi We
\begin{wrapfigure} [11] {r}{0.45\columnwidth}
    \centering
    \vspace{-14pt}
    \hspace{-15pt}
    \includegraphics[width=0.41\columnwidth]{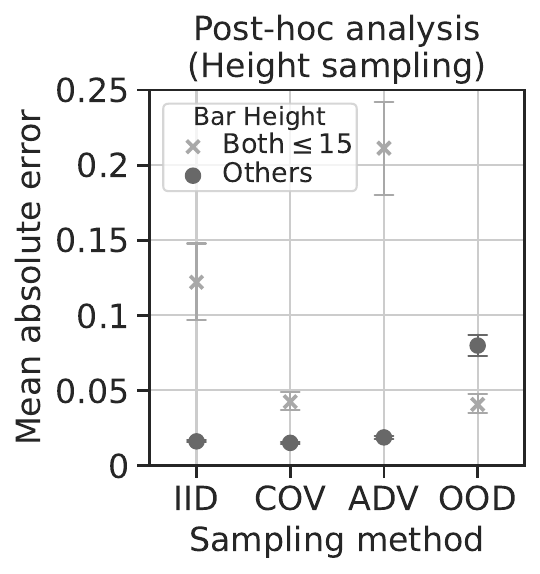}
\end{wrapfigure} performed Tukey's HSD test on the 80 bar height levels and found that smaller bar heights (H$\leq$15 pixels) increased inference errors compared to larger bar heights, independent of sampling methods. This small height effect was also observed in the ratio sampling experiment, where having H$\leq25$ pixels was necessary to lower the inference errors (See \sm~\autoref{fig:ratio_sampling_categorical}).

\ifdetail \tochange{Be more specific: how small do the features have to be in order for the CNN to fail? And why do you claim that this small feature size is "common"? Do you have any supporting evidence for that statement?}
\fi \ifdetail \tochange{What do these results mean in plain English? Do you have any recommendations?}
\fi

\rvision{\textbf{The effect of types varied by experiments.} Sampling effect H3 is partially supported. Type was a significant main effect in ratio sampling only. We see that \cnnn' differences in inference accuracies between types are much smaller than those of humans.}

\textbf{Model was also a significant main effect.} Specifically, \vgg outperformed \resnet in our experiments. We suspect that this performance difference can be attributed primarily to the way we configure the network, rather than to any inherent advantages or disadvantages of these architectures. Given these findings, we focus our subsequent discussion on results obtained for Type 1 and \vgg, and leave other types and \resnet in
\sm~\autoref{fig:sm_sampling}.\begin{figure*}
    \centering
    \includegraphics[width=\textwidth,trim={0 5pt 0 0},clip]{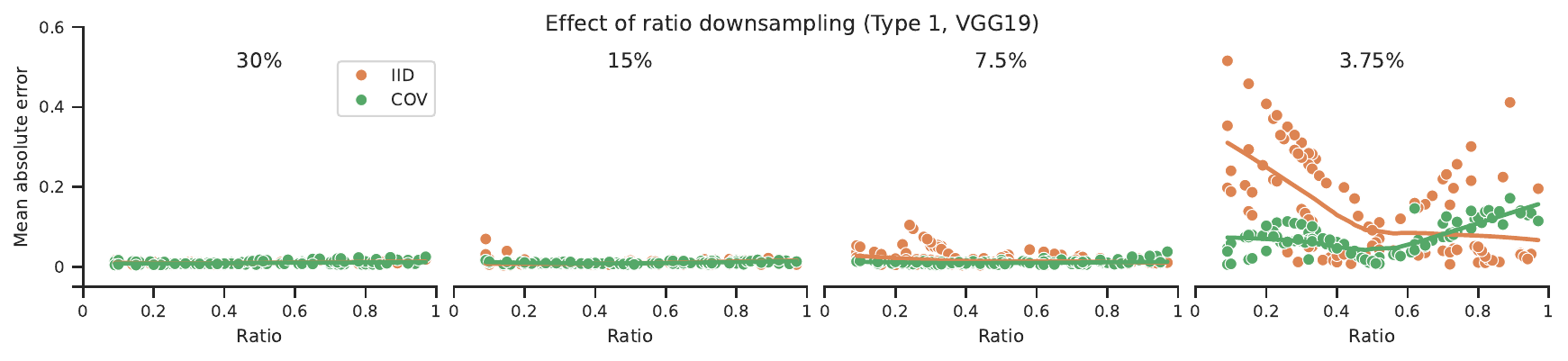}
    \caption{\textbf{\siititle.} Ratio downsampling case-by-case analysis for Type 1, \vgg at four downsampling levels.
    \textbf{\ul{Observations.}} As the number of training samples decreases, \cov exhibited stronger stability compared to \iid. While \iid initially perform similarly to \cov, its errors increased substantially under further downsampling. This was due to the formation of small clusters, causing many test samples to fall out-of-distribution and leading \iid to behave more like \mini. }
    \vspace{-10px}
    \label{fig:downsamplingByMethodLevel}
\end{figure*}
\section{Study II: {Downsampling to Test Model Stability }}
\label{sec:study.stability}

\noindent This section reports results of the \textit{sampling stability}, which examines how the number of unique training samples impacts inference errors. \ifdetail
\tochange{Why is this a good thing to study? Refer back to the goals.} \fi

Understanding model stability can help reduce training costs and guide the selection of more resilient sampling strategies. We also ran two experiments, each downsampling ratio bins or bar heights. We varied two independent variables, one at a time: the sampling method (4-levels) and the downsampling level (3-levels) in a full-factorial design. The configuration led to $4\times3=12$ conditions for each experiment, and running 5 times led to a total of $12\times5=60$ models in each experiment. Besides \textit{accuracy}, we quantified \textit{intra-consistency} to measure the variability of the inference results obtained using the same sampling method at different downsampling levels. Higher consistency indicates better model stability.

\subsection{Design}
\subsubsection{Training Data Sampling }
\noindent We used the same test sets as Study I. The implementation of downsampling was straightforward, described in \autoref{sec:limitedresources}. The training samples were a subset of those used in Study I,
\autoref{sec:study.outofdistribution}. The number of training ratio bins was 14, 7, or 3, and the number of heights was 12, 6, or 3, at downsampling levels of 15\%, 7.5\%, or 3.75\% respectively, taken in order from our sampling configurations in~\autoref{fig:ratioHeightSampling} and~\autoref{fig:heightSamplingSM}. \ifdetail \tochange{Why are these good parameter ranges? Explain and justify your choices. }
\fi

\subsubsection{Model Stability Hypotheses}

\noindent We had the following working hypotheses:
\ifdetail \tochange{Why are these reasonable based on the prior literature? Make the connection. \jcc{there are mostly intuition at the time of writing.}}
\fi

\begin{itemize}
    \setlength\itemsep{0em}

    \item \textbf{\siititle.H1} (\textit{downsampling accuracy hypothesis}):
    Decreasing the number of unique samples in the training set will increase \cnnn' inference errors.

    \item \textbf{\siititle.H2} (\textit{downsampling stability hypothesis}):
    The four sampling methods will not be equally robust.
\end{itemize}

These hypotheses could be supported because \cov and \iid remained diverse, while \mini could introduce greater distribution shifts towards the smaller values. Under constrained resources, \adv also introduced out-of-distribution cases, further reducing its reliability.

\begin{figure}[!t]
    \centering
    \hspace{-10pt}
    \begin{subfigure}
        [T]{0.49\columnwidth}
        \includegraphics[height=120pt]{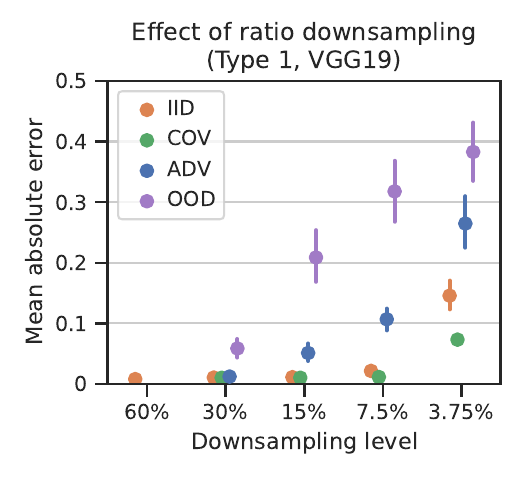}
        \put(-115, 9){\footnotesize \circledchar{a}}
        \label{fig:downsampling_ratio_summary}
    \end{subfigure}
    \begin{subfigure}
        [T]{0.49\columnwidth}
        \includegraphics[height=120pt]{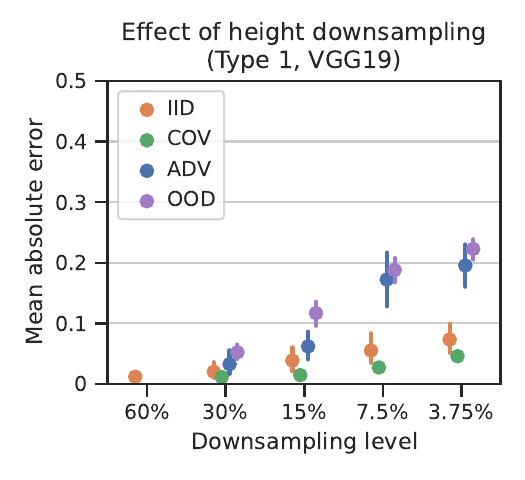}
        \put(-115, 9){\footnotesize \circledchar{b}}
        \label{fig:downsampling_height_summary}
    \end{subfigure}
    \caption{\textbf{\siititle.} \textbf{Effect of downsampling on model stability} for
    \circledchar{a} ratio and \circledchar{b} height sampling experiments.
    \textbf{\ul{Observations.}} (1) \cov was the most resilient to downsampling with limited training resources.
    (2) \adv was not stable and its errors increased as quickly as \mini.
    }
    \label{fig:downsampling_overall}
\end{figure}

\begin{table}[!t]
    \caption{\textbf{\siititle. Summary statistics}.
    }
    \label{tab:study2_stat_main} \resizebox{\columnwidth}{!}{    \addtolength{\tabcolsep}{-0.4em}
    \begin{tabular}{@{}lllll@{}}
        \toprule Variable                                               & \multicolumn{1}{c}{$F$} & \multicolumn{1}{c}{$p$} & \multicolumn{1}{c}{$\eta^{2}$} & \multicolumn{1}{c}{HSD}                   \\
        \midrule \multicolumn{5}{l}{Study II Exp 1: Ratio downsampling}  \\
        \textbf{Level}                                                  & $F_{(3, 1641)}= 150.2$  & $\boldsymbol{<0.001}$   & $\boldsymbol{0.17}$            & $\mathrm{(30\%)>(15\%)>(7.5\%)>(3.75\%)}$ \\
        \textbf{15\%}                                                   & $F_{(3, 408)}= 68.2$    & $\boldsymbol{<0.001}$   & $\boldsymbol{0.33}$            & $\mathrm{(COV, IID, ADV)>(OOD)}$          \\
        \textbf{7.5\%}                                                  & $F_{(3, 408)}= 112.1$   & $\boldsymbol{<0.001}$   & $\boldsymbol{0.45}$            & $\mathrm{(COV, IID)>(ADV)>(OOD)}$         \\
        \textbf{3.75\%}                                                 & $F_{(3, 408)}= 56.6$    & $\boldsymbol{<0.001}$   & $\boldsymbol{0.29}$            & $\mathrm{(COV)>(IID)>(ADV)>(OOD)}$        \\
        \midrule \multicolumn{5}{l}{Study II Exp 2: Height downsampling} \\
        \textbf{Level}                                                  & $F_{(3, 1273)}= 70.6$   & $\boldsymbol{<0.001}$   & $\mathit{0.12}$                & $\mathrm{(30\%)>(15\%)>(7.5\%)>(3.75\%)}$ \\
        \textbf{15\%}                                                   & $F_{(3, 316)}= 21.6$    & $\boldsymbol{<0.001}$   & $\boldsymbol{0.17}$            & $\mathrm{(COV, IID)>(IID, ADV)>(OOD)}$    \\
        \textbf{7.5\%}                                                  & $F_{(3, 316)}= 34.1$    & $\boldsymbol{<0.001}$   & $\boldsymbol{0.24}$            & $\mathrm{(COV, IID)>(OOD, ADV)}$          \\
        \textbf{3.75\%}                                                 & $F_{(3, 316)}= 46.7$    & $\boldsymbol{<0.001}$   & $\boldsymbol{0.31}$            & $\mathrm{(COV, IID)>(OOD, ADV)}$          \\
        \bottomrule
    \end{tabular} }
\end{table}

\subsection{Analysis Method}

\noindent We used the same method to analyze inference errors as described in Study I. In additionally to mean absolute error, we further measured the intra-consistency of model inferences between downsampling levels by the Pearson correlation coefficient~\cite{Cohen1969StatisticalPA}. A higher Pearson correlation indicates greater intra-consistency and stability for a sampling method in response to limited resources.

\begin{figure}[!t]
    \includegraphics[height=61pt]{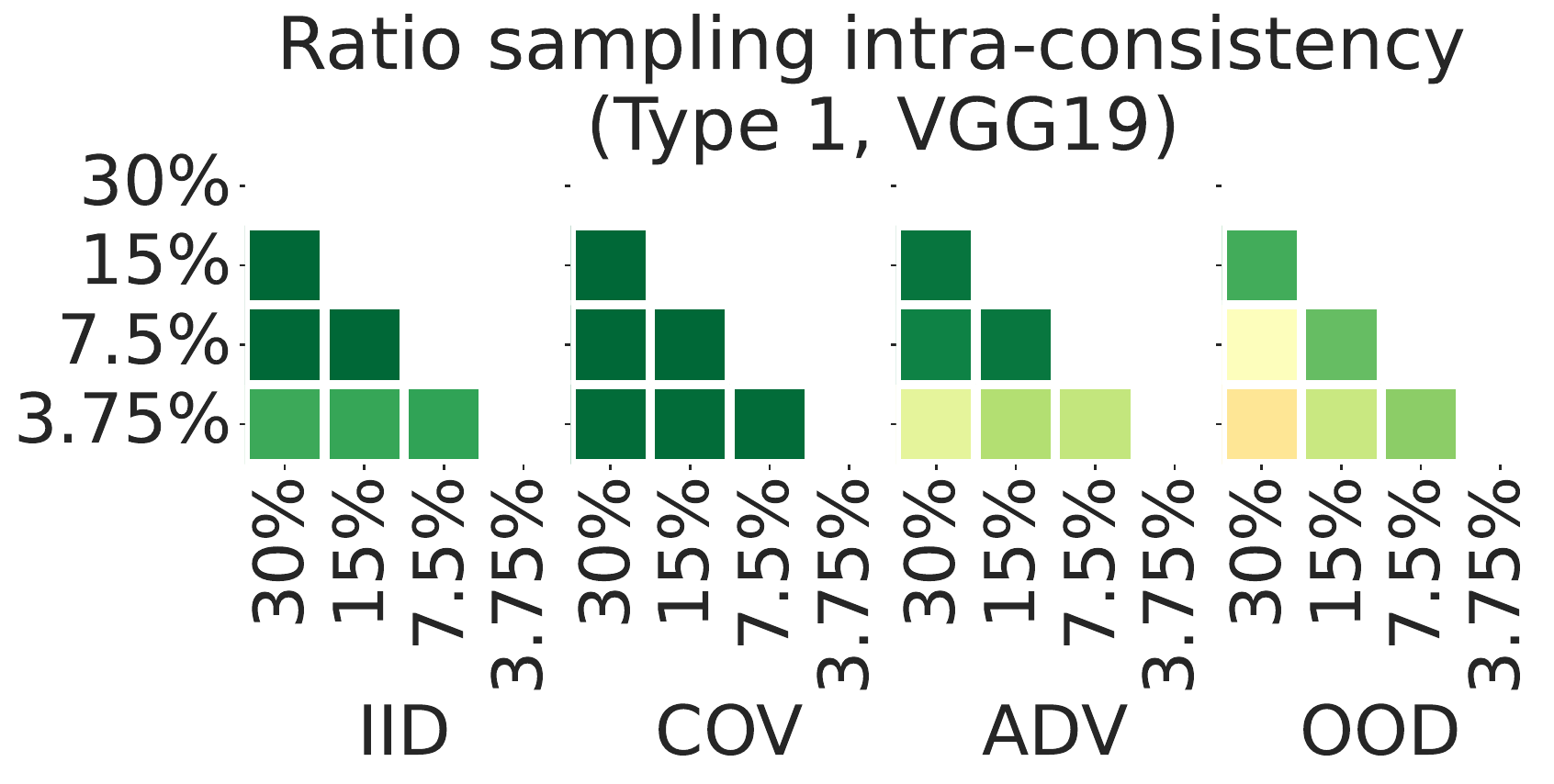}
    \includegraphics[height=61pt, trim=120 0 0 0, clip]{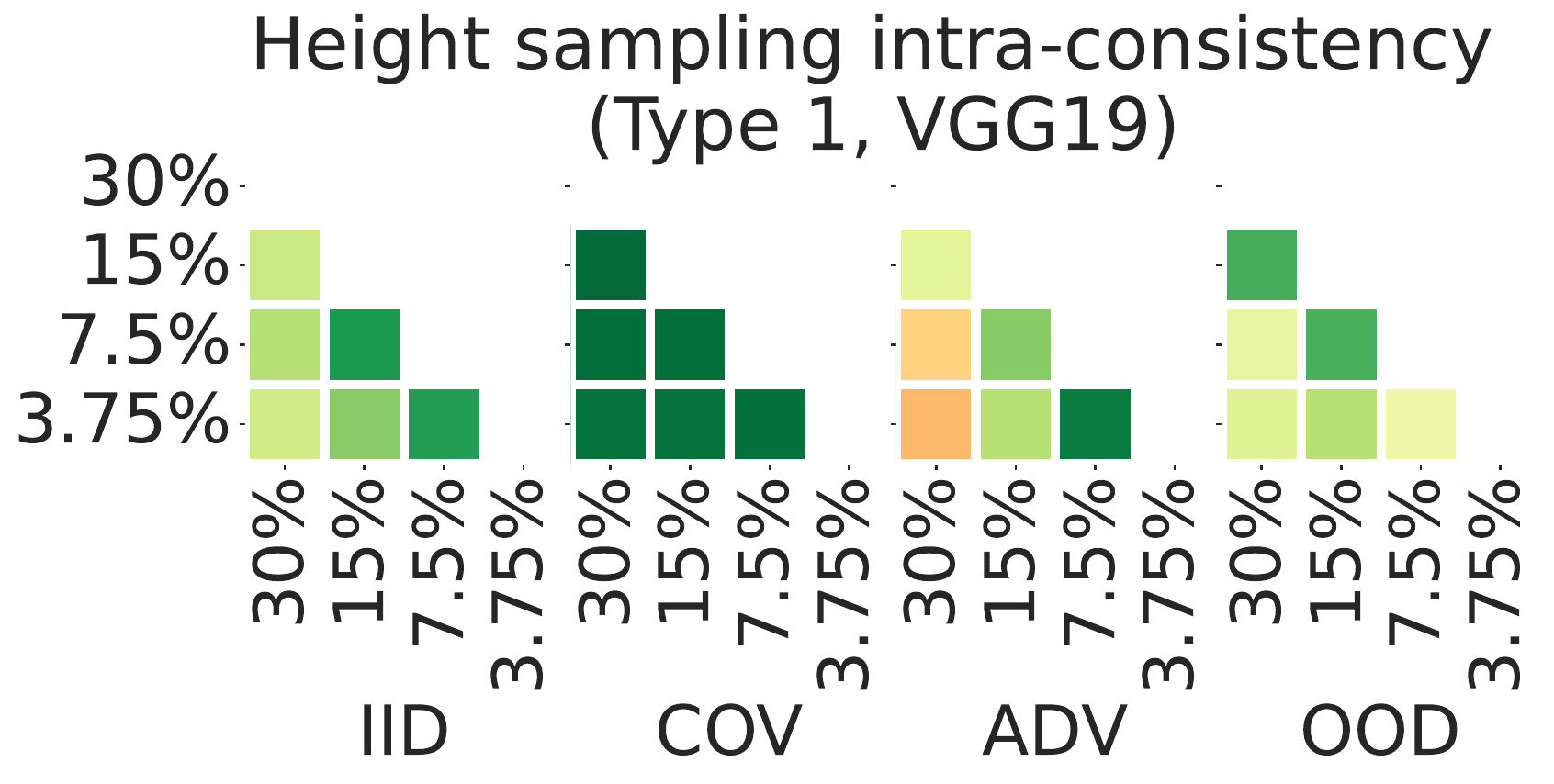}
    \includegraphics[height=61pt]{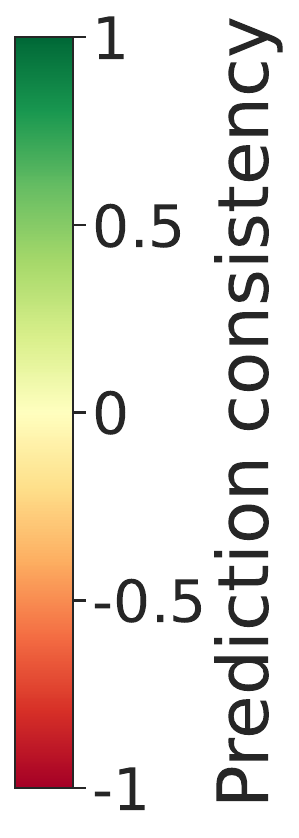}
    \caption{\textbf{\siititle.} Intra-consistency computes the correlation of model inferences between different downsampling levels.
    \textbf{\ul{Observations.}}
    \covmodel had the highest consistency for both ratio and height sampling experiments.
    \ifdetail
    \tochange{do you mean training method correlation?? describe observations from the drawing. Also, the text is too small. only need to show half of the figure. make the two figures appear in the same row.\jc{move this figure to the bottom of the last page.. and so does Figure 7. so they can be compared. Figure 7 please use a smaller figure column size so the blocks looked the same as Figure 6.}}
    \fi
    }
    \label{fig:downsamplingConsistency}
\end{figure}

\subsection{Results}
\ifdetail \tochange{Make this a subsection title.} \fi

\noindent
\textbf{Overview.} We collected a total of 1.2 million predictions ($60\; \mathrm{models}\times 20\mathrm{K}\; \mathrm{predictions}$) for each of the ratio and height stability experiments.
\autoref{fig:downsampling_overall} and \autoref{tab:study2_stat_main} show the mean errors and the summary statistics. Both hypotheses were supported. (More analyses in \sm~\autoref{fig:sm_ratio_downsampling_vgg}.)

\textbf{Reducing the number of unique training samples increased inference errors, with \covmodel exhibiting half the errors of \iidmodel.} \siititle.H1 was supported. The downsampling level was a significant main effect on errors. In the ratio sampling experiment, inference errors for \iidmodel increased more than tenfold, from 0.011 at a downsampling level of 30\% to 0.146 at 3.75\%. For \covmodel, errors rose about sevenfold, from 0.010 to 0.073.
\covmodel and \iidmodel were in the same Tukey group until the $3.75\%$ downsampling condition (\autoref{tab:study2_stat_main}). At $15\%$ downsampling level, \adv (0.051) was in the same group as \iid (0.011) and \cov (0.010), but its error increased sharply ($>10\times$) between 7.5\% and 3.75\%. These results suggested that \cov was the most stable sampling method among the four. The same trend was observed in the bar height experiment, where \covmodel again produced the lowest errors~(\autoref{fig:downsampling_overall}).

\ifdetail \tochange{Why? \jcc{see addition exploratory exp and discussion}} \fi

\textbf{\cov is the most stable sampling method.} {Stability.H2 was supported.} We assessed model stability using intra-consistency, calculated as the correlation of model inferences across different downsampling levels. As shown in~\autoref{fig:downsamplingConsistency}, \iidmodel and \covmodel had the largest intra-consistency ($r>0.99$) between the 30\%, 15\%, and 7.5\% levels in the ratio downsampling experiment. For \iidmodel, consistency dropped to $r=0.72$ when including the 3.75\% level, whereas \covmodel maintained a high consistency of $r=0.98$. Intra-consistencies were much lower for \advmodel ($r=0.59$) and \oodmodel ($r=0.32$) on average. In the height downsampling experiment, \covmodel also showed the highest intra-consistency ($r=0.97$), compared to \iidmodel ($r=0.48$). \advmodel ($r=0.21$) had even lower consistency than \oodmodel ($r=0.34$). These results indicate that the COV sampling method is the most stable across different downsampling levels, maintaining high intra-consistency even at extreme reductions.

\autoref{fig:downsamplingByMethodLevel} provides a case-by-case analysis of the ratio downsampling experiment, further illustrating the stability of \cov. \mini errors increased almost linearly as test cases moved further out-of-distribution. Both \iid and \adv also exhibited \mini-like behavior when training samples were limited. In contrast, \cov enforced in-distribution conditions for all test cases, preventing errors from rising linearly; instead, they remained bounded by the evenly distributed training samples. This is most evident at level 3.75\%, where only three samples were used for training. The errors of \cov were stable, and lower errors occurred when test samples were close to the training samples---specifically, at the leftmost, rightmost, and middle positions.

\ifjccom \jc{number?}
\fi
\ifjccom \jc{number?} \fi
\ifjccom \jc{report the mean intra-consistency?} \fi

\ifjccom \jc{show rs for these} \fi
\ifdetail \tochange{Do you have an explanation why this may be the case? \jcc{see below. I think it is the data span.}}
\fi
\section{Study III: Human vs. \cnns: {A Comparative Study}}
\label{sec:study.humanai}

\noindent Given the relatively small errors in the first two studies when applying \iid and \cov, we proceeded to compare \cnnn and humans and answer the question: \textit{to what extent do \cnnn and humans differ?} For the human observer, the bar chart type was the only independent variable considered. For the \cnns observer, we introduced a new baseline condition (IID-large) that utilized the entire remaining 60\% of ratio bins or heights without downsampling. This corresponded to 56 ratio bins in the ratio sampling experiment or 48 height bins in the height sampling experiment. We denoted this condition as ``IID-large''. We also reused the \iid and \mini conditions from Study I~\autoref{sec:study.outofdistribution}. We did not use \cov since its error was bounded by the error of \iid.

\subsection{Design}

\subsubsection{Human Experiment Data Collection}
\label{sec:human_data_collect} We adapted the position-length experiment in Cleveland and McGill~\cite{cleveland1984graphical} using the 10 integer bar heights of $10, 12, 15, 18, 22, 26, 32, 38, 46, 56$. Exhausting all pairs generated 45 height combinations, resulting in 37 unique ratios across 23 ratio bins: 0.18, 0.26, 0.27, 0.38, 0.39, 0.45, 0.46, 0.47, 0.48, 0.55, 0.56, 0.57, 0.58, 0.67, 0.68, 0.69, 0.7, 0.8, 0.81, 0.82, 0.83, 0.84, 0.85.

We conducted the experiment on the LabIntheWild~\cite{labinthewild} crowdsourcing platform (see the experiment user interface in \sm~\autoref{fig:labinthewide_ui}). Participants were instructed to make a quick division between the heights of the shorter marker bar and the taller marked bar. They first completed 5 training trials---one for each bar chart type---with immediate feedback provided after each response. After that, each participant completed 25 trials, consisting of 5 chart types with 5 trials per type in random order.

\subsubsection{Human-Machine Comparison Hypothesis}
\noindent We had the following working hypotheses: \begin{itemize}
    \setlength\itemsep{0em}

    \item \textbf{\shumanaititle.H1} (\textit{accuracy hypothesis}): Our
    \cnnn' errors will be significantly lower than humans'.
\end{itemize}

This hypothesis could be supported based on our observation comparing our results in the first two experiments to those in the human experiments of Cleveland and McGill~\cite{cleveland1984graphical}. \cnnn appeared to have much smaller errors compared to those of human observers.

\subsection{Analysis Method}
\noindent We used mid-means of the log absolute error (MLAE)~\cite{cleveland1984graphical} to measure accuracy, in order to compare our results to those published in literature, \ie, Cleveland and McGill~\cite{cleveland1984graphical}, Heer and Bostock~\cite{heer2010crowdsourcing}, and Haehn et al.~\cite{haehn2018evaluating}. Here $\mathrm{MLAE}=\mathrm{log}_{2}(|\mathrm{model\;inference}-\mathrm{ground\;truth} |+1/8)$. The height sampling experiments used mean absolute error to compare to that of Talbot et al.~\cite{talbot2014four}. Finally, we evaluated human-\cnns inter-consistency using Pearson correlation analyses.

\subsection{Results}
\label{sec:human.results}

\begin{figure}[!t]
    \centering
    \begin{subfigure}
        {\columnwidth}
        \centering
        \includegraphics[width=\columnwidth]{
        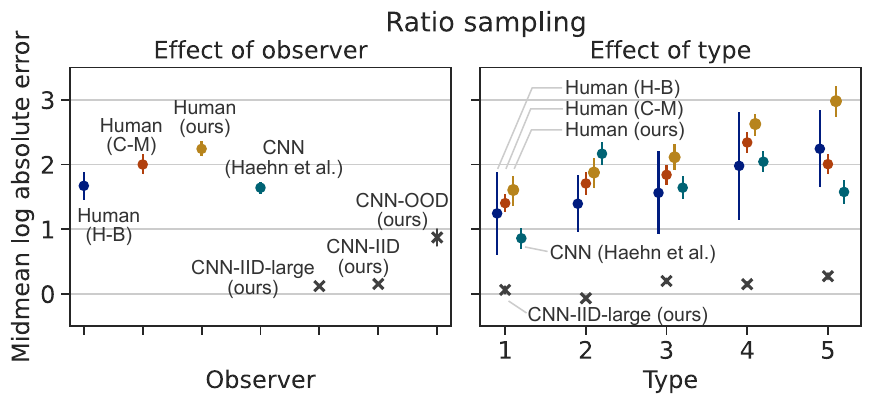
        }
        \put(-235, 5){\footnotesize \circledchar{a}} \put(-115, 5){\footnotesize \circledchar{b}}
        \label{fig:human_ratio_sampling_overall_summary}
    \end{subfigure}
    \caption{\textbf{\shumanaititle. Ratio sampling.} \circledchar{a} Comparison of the average errors across five bar chart types for seven observers: (1) Cleveland and McGill (C-M)~\cite{cleveland1984graphical}, (2) Heer and Bostock (H-B)~\cite{heer2010crowdsourcing},
    (3) Haehn et al.
    ~\cite{haehn2018evaluating}, (4) our human experiment, and (5)-(7) our three \cnnn (IID-large, \iid, and \mini).
    \circledchar{b} Per-type comparison among observers.
    Error bars are 95\% confidence intervals.
    \textbf{\ul{Observations.}}
    (1) From \circledchar{a}, our human results are comparable to those of C-M~\cite{cleveland1984graphical} and Heer and Bostock~\cite{heer2009sizing}. Our \cnnn lowered errors greatly compared to that of
    Haehn et al~\cite{haehn2018evaluating}.
    (2) From \circledchar{b}, \cnnn didn't have strong preference for certain types, unlike humans.
    }
    \vspace{-10px}
    \label{fig:samplingHumanMainEffects}
\end{figure}

\begin{figure}[!t]
    \centering
    \begin{subfigure}[T]{\columnwidth}
        \centering
        \includegraphics[width=\columnwidth]{
        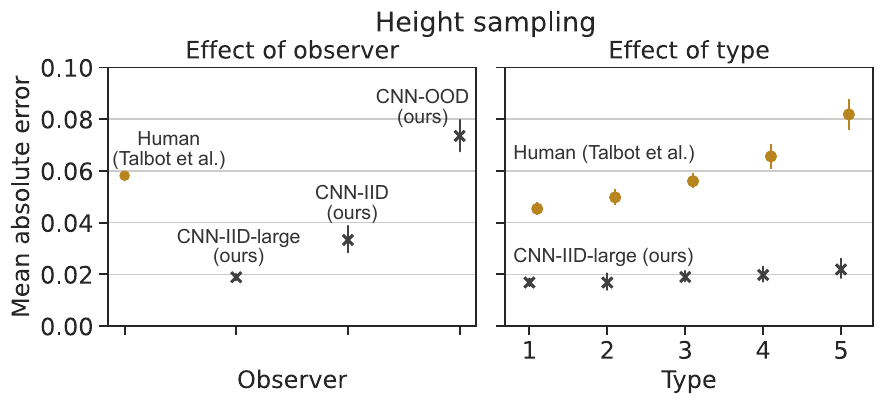
        }
        \put(-235, 5){\footnotesize \circledchar{a}} \put(-115, 5){\footnotesize \circledchar{b}}
        \label{fig:human_height_sampling_overall_summary}
    \end{subfigure}
    \caption{\textbf{\shumanaititle. Height sampling.}
    Mean absolute errors between Talbot et al.~\cite{talbot2014four} and our
    \cnnn.
    Error bars are 95\% confidence intervals.
    \textbf{\ul{Observations.}} (1) From \circledchar{a} \cnnn outperformed humans for all but
    \mini-sampling conditions. (2) From \circledchar{b} chart type had more influence on humans than on \cnnn. }
    \label{fig:samplingHumanMainEffectsHeight}
    \end{figure} \begin{table}[!t]
    \caption{\textbf{\shumanaititle. Summary statistics.}
    }
    \label{tab:study_human_stat_main} \resizebox{\columnwidth}{!}{    \addtolength{\tabcolsep}{-0.4em}
    \begin{tabular}{@{}lllll@{}}
        \toprule Variable                                           & \multicolumn{1}{c}{$F$} & \multicolumn{1}{c}{$p$} & \multicolumn{1}{c}{$\eta^{2}$} & \multicolumn{1}{c}{HSD}                   \\
        \midrule \multicolumn{5}{l}{Study IV Exp 1: Ratio sampling}  \\
        \textbf{Observer}                                           & $F_{(3, 3327)}= 334.3$  & $< \mathbf{0.001}$      & $\mathbf{0.22}$                & $\mathrm{(\text{IID-large}, IID)>(OOD)>(Human)}$  \\
        \textbf{Type (human)} & $F_{(4, 249)} = 24.6$   &	$< \mathbf{0.001}$   &	$\mathbf{0.28}$   &	$\mathrm{(1, 2, 4)>(1, 3, 4)>(3, 4, 5)}$ \\
        \midrule \multicolumn{5}{l}{Study IV Exp 2: Height sampling} \\
        \textbf{Observer}                                           & $F_{(3, 7278)}= 151.2$  & $< \mathbf{0.001}$      & $\mathit{0.06}$                & $\mathrm{(\text{IID-large})>(IID)>(Human)>(OOD)}$ \\
        \textbf{Type (human)}   &	$F_{(4, 4882)} = 50.9$   &	$< \mathbf{0.001}$   &	$0.04$   &	$\mathrm{(1, 2)>(2, 3)>(3, 4)>(5)}$\\
        \bottomrule
    \end{tabular} }
    \end{table} \begin{figure}[!t]
    \centering
    \begin{subfigure}
        [T]{0.49\columnwidth}
        \includegraphics[height=120pt]{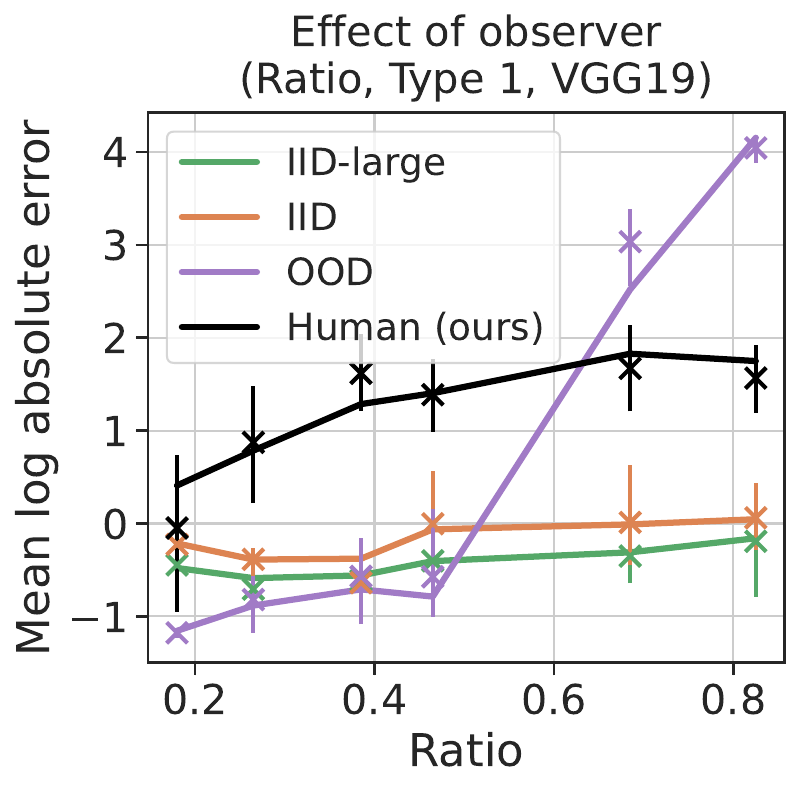}
        \put(-115, 5){\footnotesize \circledchar{a}}
    \end{subfigure}
    \begin{subfigure}
        [T]{0.49\columnwidth}
        \includegraphics[height=120pt]{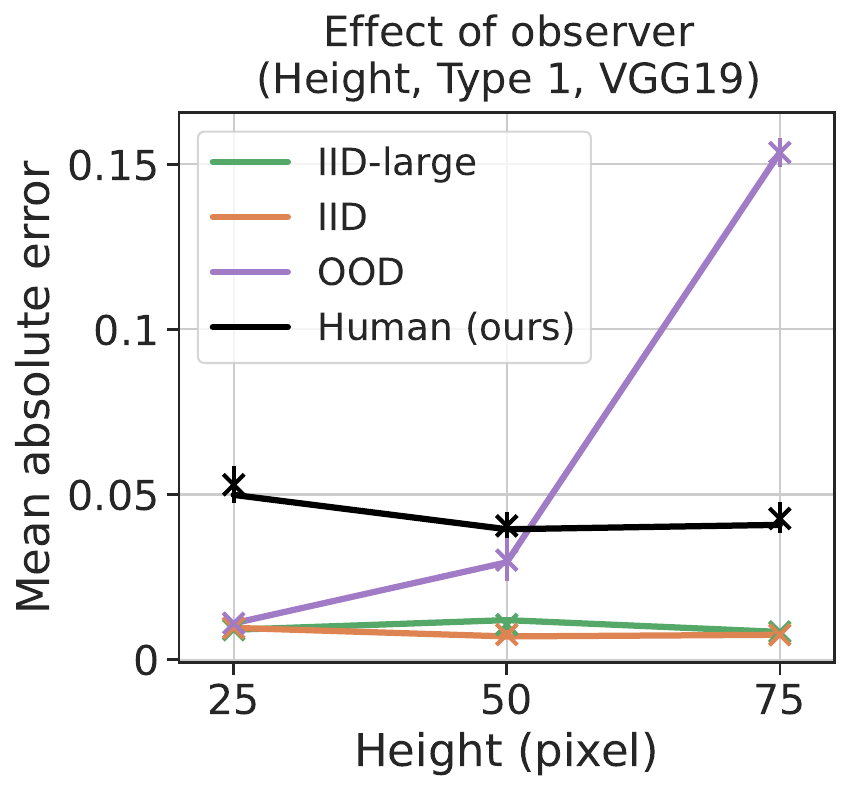}
        \put(-115, 5){\footnotesize \circledchar{b}}
    \end{subfigure}
    \caption{\textbf{\shumanaititle.} Case-by-case analysis of inference errors between humans and \cnnn.
    \textbf{\ul{Observations.}}
    \cnnn can outperform humans in general, but if not trained properly, they may perform worse than humans on out-of-distribution tests.
    }
    \label{fig:humanAICaseByCase}
\end{figure}

\noindent
\textbf{Overview.} We collected human observers' results from 1,575 trials from 63 participants, who spent an average of 10.0 seconds per trial.
\autoref{fig:samplingHumanMainEffects}, \autoref{fig:samplingHumanMainEffectsHeight}, and \autoref{tab:study_human_stat_main} show the mean errors and the summary statistics. Overall, our human experiment results showed slightly higher error than Cleveland and McGill's (\autoref{fig:samplingHumanMainEffects}\circledchar{a}), but the trend across the five chart types remained consistent with the findings of Cleveland and McGill~\cite{cleveland1984graphical} and Heer and Bostock~\cite{heer2010crowdsourcing}~(\autoref{fig:samplingHumanMainEffects}\circledchar{b}). (\sm\autoref{fig:cmLITWCompare} has more details.)
\textbf{\cnnn had lower errors than humans}. {\shumanaititle.H1} was supported. The observer was a main effect. For ratio sampling, human errors (0.056) were about four times that of \cnnn' \iid-large (0.013) ~(\autoref{fig:samplingHumanMainEffects}\circledchar{a}), and for height sampling, human errors (0.058) were about three times of \cnnn' \iid-large (0.019)~(\autoref{fig:samplingHumanMainEffectsHeight}\circledchar{a}).

\textbf{Chart type was also a significant main effect.} The differences between types were observed in human experiments (\autoref{fig:samplingHumanMainEffects}\circledchar{b} and \autoref{fig:samplingHumanMainEffectsHeight}\circledchar{b}). In addition,
\cnnn had smaller uncertainty (95\% CI for Type 1, Baseline, ratio: 0.002, height: 0.002) compared to human observers (95\% CI, ratio: 0.007, height: 0.004).

\begin{figure}[!t]
    \centering
    \includegraphics[height=65pt]{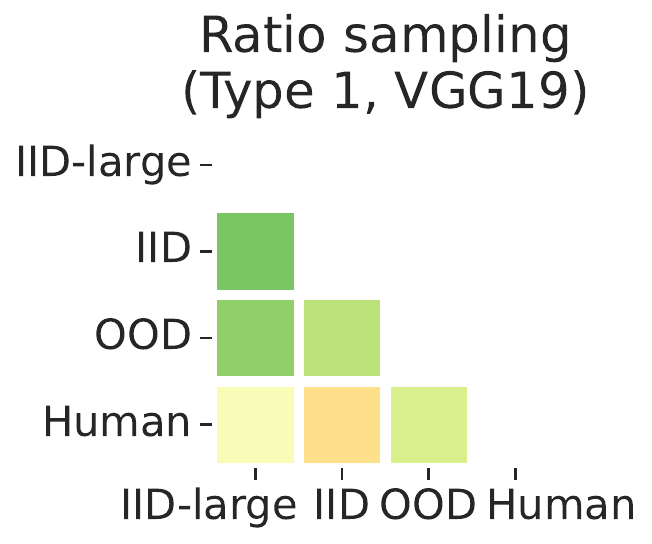}
    \includegraphics[height=65pt]{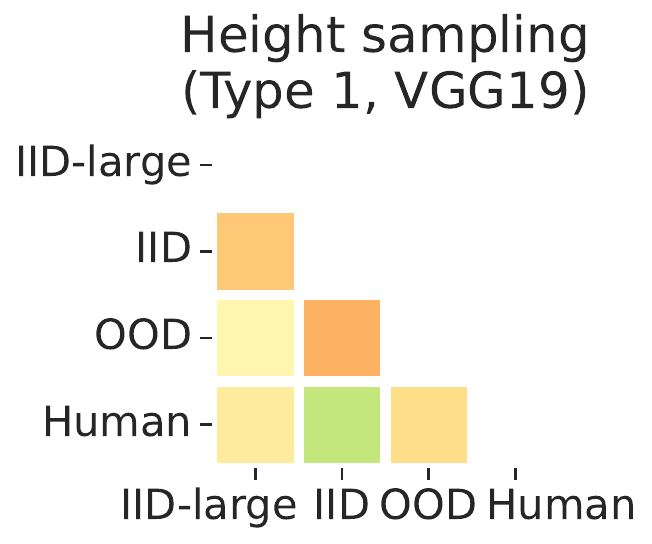}\hspace{10pt}
    \includegraphics[height=65pt]{
    Pictures/StudyIIRatioInnerCorrelationSmallCbar.pdf
    }
    \caption{\textbf{\shumanaititle.}
    Inter-consistency between humans and \cnnn.
    \textbf{\ul{Observations.}}
    The inter-consistency between humans and \cnnn is relatively low.
    }
    \label{fig:humanCNNCorrelation}
\end{figure}

\textbf{Human-\cnns inter-consistency is lower than the intra-consistency between \cnnn~} (\autoref {fig:humanCNNCorrelation}). In ratio sampling: $r=0.01, p>0.1$; and in height sampling: $r=-0.02, p>0.1$. This suggested that human errors differed from those in \cnnn.

\section{General Discussion}

\noindent This section presents explanations for our findings, reflects on approaches for quantifying \cnnn' capabilities, and provides practical recommendations for training data preparation.

\begin{figure}[!t]
    \centering
    \begin{subfigure}
        [T]{0.492\columnwidth}
        \includegraphics[height=120pt]
        {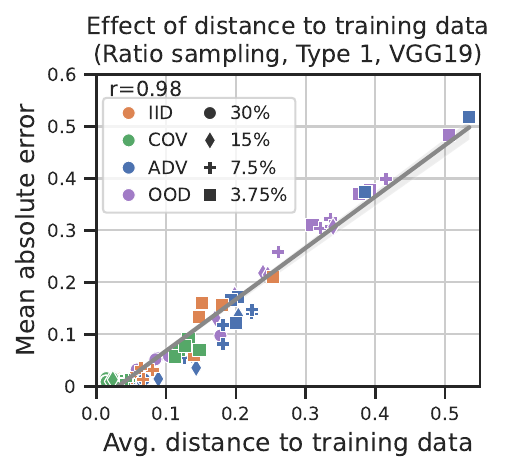} \put(-117, 5){\footnotesize \circledchar{a}}
        \label{fig:downsampling_overall_cover_ratio}
    \end{subfigure}
    \begin{subfigure}
        [T]{0.492\columnwidth}
        \includegraphics[height=120pt]
        {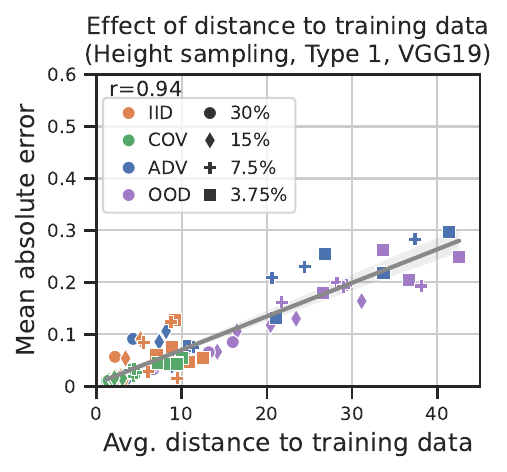} \put(-117, 5){\footnotesize \circledchar{b}}
        \label{fig:downsampling_overall_cover_height}
    \end{subfigure}
    \caption{\textbf{\siititle.} Inference errors vs. \distanceName.
    Each point represents a run and there were 80 runs (5 runs$\times$ 4 sampling methods $\times$ 4 downsampling levels).
    \ul{\textbf{Observations.}}
    We observed a strong correlation ($r=$0.98 and 0.94) between the \distanceName and the inference error. Compared to other sampling methods, \cov \textcolor[HTML]{57a86a}{$\sbullet[1]$} tends to distribute samples more evenly, resulting in a smaller \distanceName and smaller inference errors.
    }
    \vspace{-10px}
    \label{fig:downsampling_distance}
\end{figure}

\begin{figure}[!t]
    \centering
    \begin{subfigure}
        [T]{0.492\columnwidth}
        \includegraphics[height=120pt]{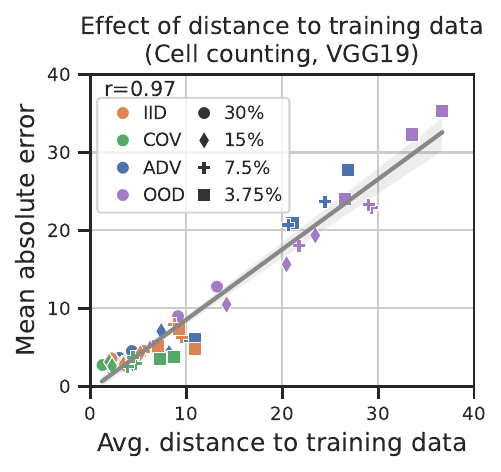} \put(-117, 5){\footnotesize \circledchar{a}}
        \label{fig:downsampling_overall_distance_ratio}
    \end{subfigure}
    \begin{subfigure}
        [T]{0.492\columnwidth}
        \includegraphics[height=120pt]{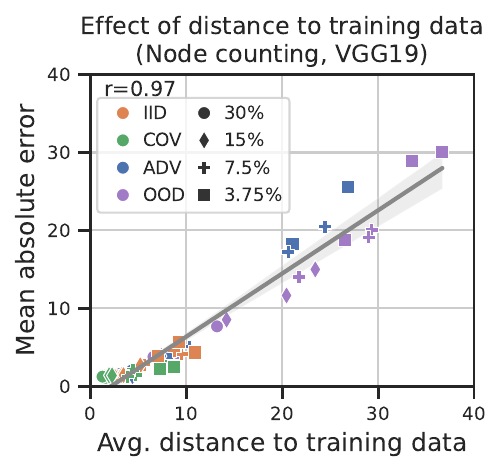} \put(-117, 5){\footnotesize \circledchar{b}}
        \label{fig:downsampling_overall_distance_height}
    \end{subfigure}
    \caption{\rvision{\textbf{Reuse of our stability results for sampling training data counts.}}
    Inference errors vs. \distanceName for cell and node counting tasks.
    \ul{\textbf{Observations.}}
    \rvision{We observed
    a strong correlation between the \distanceName and the model inference error in both experiments (cell: $r=0.97$, node: $r=0.97$),
    suggesting that the effect of \distanceName can be  generalized.}
    \rvision{See \sm~\autoref{sec:cell_node_count} for study configurations.}}
    \vspace{-10px}
    \label{fig:downsampling_distance_cellnode}
\end{figure}

\subsection{The Effect of Training-Test Distances \rvision{and Reusability}}
\label{sec:discussionTrainTestDistance} One of the most intriguing findings in our study was the stability of the COV sampling method, particularly when the total number of training samples was limited (\autoref{fig:downsampling_overall}). We hypothesize that this stability stems from the key property of \cov: it ensures a broad coverage of the data space while maintaining an even distribution of training samples. By maintaining a well-spread distribution of training samples, COV effectively minimizes large gaps in the training space, ensuring that test samples are likely to have nearby training samples for reference.

To validate this hypothesis, we analyzed the relationship between inference error and the \distanceName---the average distance from each test sample to its closest training sample. As shown in ~\autoref{fig:downsampling_distance}, there was a strong positive correlation in both ratio ($r=0.98$) and height ($r=0.94$) experiments, confirming that models performed more accurately when test samples were closer to their nearest training neighbors. Among all sampling strategies evaluated, \cov consistently exhibited the lowest \distanceName at equivalent downsampling levels, further supporting its advantage in stability.

\rvision{ To examine the reusability of this stability result,} we investigated whether the impact of \distanceName extends beyond the ratio estimation task. We observed a similar pattern in two additional challenging real-world counting tasks. \rvision{The first is counting cells in fluorescence-light microscopy images rendered using the method of Lehmussola et al.~\cite{lehmussola2007computational}. It required \cnnn to identify cells based on continuous color, textures, and shapes. The second is counting network nodes in node-link diagrams rendered using high-fidelity graph simulations by Lancichinetti et al.~\cite{lancichinetti2009benchmarks}. We manipulated the community size and required \cnnn to perceive the shape of the diagram to make a prediction. The results in
\autoref{fig:downsampling_distance_cellnode} suggest that the effect of \distanceName can be generalized to new tasks. (\sm~\autoref{sec:cell_node_count} has experiment configuration details.) }
\subsection{Implications to Training Data Preparation \rvision{Concerning Reliability, Stability, and Cost-effectiveness}}
\label{sec:discussion_data_prep}

\noindent The results of our study suggest that the training sampling choices were a significant main effect on
\cnnn' inference errors. Using a few human ratio samples may disadvantage \cnnn, as exposure to a wider range of ratio samples could improve their capabilities. Implicitly mixed sampling conditions in the training could potentially increase inference uncertainty, because inference errors from an \mini-trained model could be much larger than those from an \iid-trained model. On the other hand, conditions that benefit \cnnn could also appear. When training and test samples are close neighbors, the learning algorithm can gain an advantage, leading to higher accuracy.

From a \textbf{reliability} perspective, we found that \cnnn could achieve considerable accuracy. However, they are not robust to the train-test distribution shift. In-distribution tests remained accurate, but errors increased linearly for out-of-distribution cases~(\autoref{fig:distributionShiftDetail}).
\rvision{From a \textbf{stability} perspective, models should try to cover the data range as much as possible, especially when resources are limited or when sampling more data is too expensive~(\autoref{fig:downsampling_distance}).} Under limited resources, the most commonly used \iid sampling method does not guarantee stability~(\autoref{fig:downsamplingConsistency}). The lack of robustness in the \adv condition also suggests that chart reading \cnnn could also be easily fooled, \rvision{which could lead to serious consequences for high-stakes domains.}
\rvision{From a \textbf{cost-effective} perspective, we found that using \cov can substantially reduce the training time: on average 69.7\%, 82.6\%, and 96.4\% less time compared to \iid, \adv, and \mini, respectively~(\sm~\autoref{fig:training_history}).}

These results imply that we should always \rvision{include more samples and ensure a broader training coverage when possible. This practice helps prevent adversarial conditions, in which training and test samples share the same domain but are distant from each other, a scenario that can severely degrade the accuracy of \cnnn.} Finally, we suggest enhancing transparency \rvision{to mitigate misuse and misinterpretation}. A practical measure is to incorporate a model card~\cite{Mitchell2018ModelCF}, \rvision{providing information on how training samples are selected, along with their ranges and distributions, image size, and the number of unique samples.}

\subsection{Comparison to the Results of Haehn et al.~\cite{haehn2018evaluating}}

Haehn et al.~\cite{haehn2018evaluating} stated, ``\textit{In other experiments, such as the position-length experiment, \cnnn cannot complete the task.}'' However, our Study III in~\autoref{sec:study.humanai} demonstrated that \cnnn can outperform humans. We provide an explanation for this mismatch from a data sampling perspective. In our experimental setup, the heights of target bar samples ranged in [5, 85] pixels, whereas Haehn et al. followed Cleveland and McGill’s approach, using only 10 discrete heights. As a result, our configuration led to a substantially larger set of unique ($h, H$) pairs (see~\sm~\autoref{fig:ratioHeightPairs}). Although the smaller \distanceName in Haehn et al.’s experiment (as shown in \autoref{fig:othersSamplingExample}) provided great advantages to \cnns observers, their limited variation in training ($h, H$) pairs was among the reasons that hindered models' inference accuracy.

\rvision{From the attribution study perspectives, their approach of random assignment of the C-M test samples into the training, validation, and test sets made it hard to analyze the influence of the training set on large errors and uncertainty in model inference outcomes.} Their design and analyses may have introduced large variations in test conditions as well \rvision{(see~\sm~\autoref{fig:samplingRatioDH})}. In contrast, our study explicitly controlled training and test sets, providing an interpretable perspective.

\rvision{These sampling methodological differences highlight the impact of sampling strategies on algorithmic performance.}
\rvision{ Our results aligned with computer vision studies that \cnnn' behavior constraints differ from humans'~\cite{geirhos2018imagenettrained}. These spatial tasks seem to be manageable for \cnnn when we have a good knowledge of data preparation. Directly adopting experimental designs from human studies may introduce a methodological bias, leading to models that may not generalize well~\cite{funke2021five}. Therefore, comparing models trained under such methodological biases against humans, especially without a fixed and identical test set, may not constitute a fair evaluation. }

\subsection{The effect of Small Stimuli}\label{sec:discussionSmallStimuli}
\begin{figure}[!t]
    \centering
    \begin{subfigure}
        [T]{0.49\columnwidth}
        \includegraphics[height=130pt]{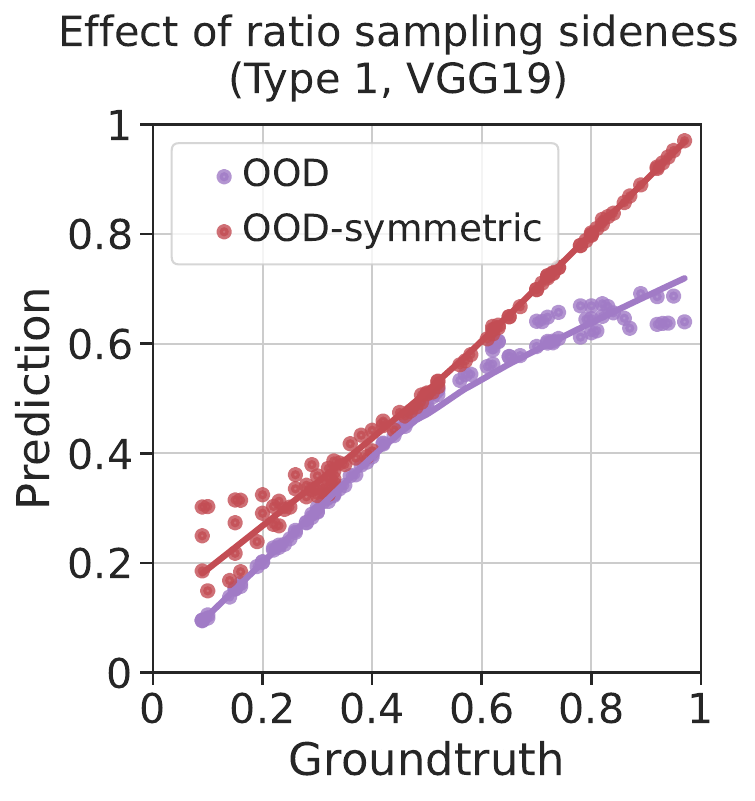}
        \put(-115, 5){\footnotesize \circledchar{a}}
    \end{subfigure}
    \begin{subfigure}
        [T]{0.49\columnwidth}
        \includegraphics[height=130pt]{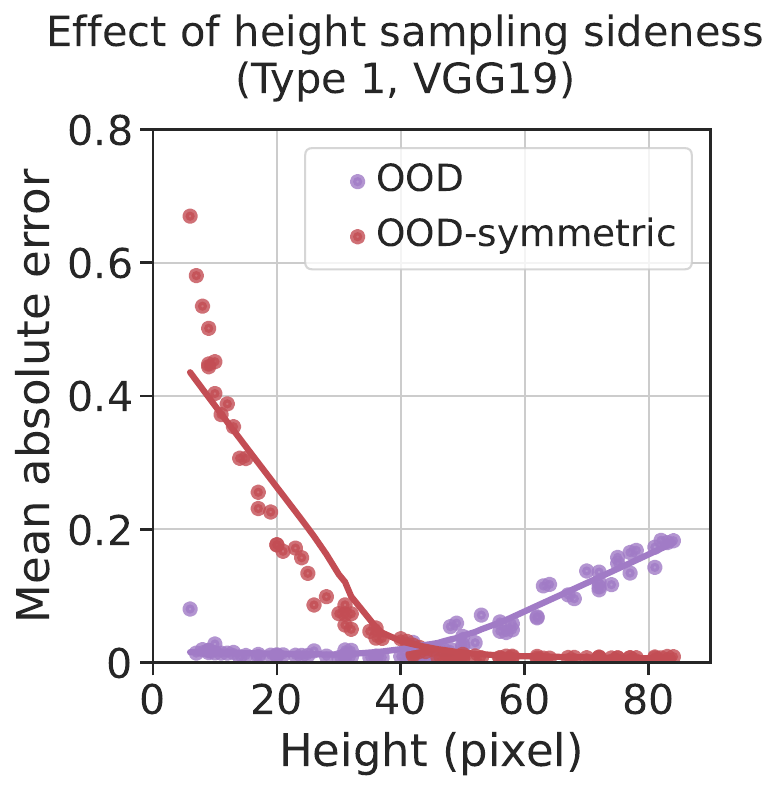}
        \put(-115, 5){\footnotesize \circledchar{b}}
    \end{subfigure}
    \caption{\textbf{Impact of smaller bar heights on \mini.} Comparison of inference
    errors between \mini and \mini-symmetric (sampling from the opposite
    direction). \textbf{\ul{Observations.}}
    Data sampling symmetric conditions (\mini vs. \mini-symmetric) exhibited asymmetric errors in \circledchar{b} height sampling only.
    }
    \vspace{-10px}
    \label{fig:sideness}
\end{figure}

\noindent Our findings from Studies I and II showed that small bar heights ($\le~15$ pixels for height sampling and $\le25$ pixels for ratio sampling) led to increased errors \rvision{for \iid and \adv, but not for \mini}~(\autoref{fig:distributionShiftDetail}). The first explanation was that
\mini encountered more images with short bars during training, as its smaller sampling domain resulted in a smaller \distanceName within the sampling regions. A second explanation could be the effect of sideness: we only selected the left side with small bar heights in our experiments and did not test \mini using regions with the largest bar heights. If similar errors occurred in the larger-side sampling, it would suggest that the error was not due to bar height itself but rather a consequence of our design choice related to the sideness. We thus examined the other side of the bar height or ratio samples under \mini conditions to train models. We observed an asymmetry in the height sampling conditions (\autoref{fig:sideness}), suggesting that smaller heights indeed led to larger errors.

\rvision{This behavior contrasts with human perception that follows Weber's Law~\cite{Householder1940WeberLT}, making ratio judgments independent of the absolute magnitude of the stimuli.
\sm \autoref{sec:smVGGResNetComparison} provides explanations to the short height effect from \cnnn' architectural perspective. To improve this condition, one can increase the input image size so that smaller bar heights are proportionally increased, or use networks and modules that are designed for small-size elements~\cite{li2017perceptual,Lin2016FeaturePN}. }

\subsection{Limitations and Future Work}
\ifdetail \tochange{Add future work.} \fi

\noindent In this work, we designed new attribution study method to quantify algorithmic behaviors, by explicitly considering training-test discrepancies in four sampling strategies. There are other sophisticated sampling methods, \eg, those following target data distribution~\cite{baptista2025approximation} could also be more effective that are worth future investigation.
\rvision{ We've developed a new attribution study method that quantifies algorithmic behaviors, offering foundational scientific insights into effective training data sampling for visualization tasks. While this method advances our understanding of algorithmic quantification, it remains a stepping stone, not a standalone solution for real-world challenges. Addressing the intricate contexts and patterns of practical analytical tasks will necessitate a series of future studies to effectively leverage our quantification methods in quantifying observers. }
\rvision{Finally, fundamental theories to quantifying \cnnn remain scarce. Methods to test the generalizability of the findings to other visualization types, beyond those tested here, are needed to study the alignment between humans and machines specifically designed for visualization uses. }

\section{Conclusions}

\noindent Our sampling regime reveals fundamental insights into \cnnn' graphical perception: while models achieve superhuman accuracy in controlled settings, their brittleness under distribution shifts, especially the adversarial sampling (\adv), suggests the need for rigorous training-test separation to test \cnnn in visualization uses. Coverage-based sampling (COV) emerged as a robust strategy, largely due to its maximized data span and evenly distributed samples, which is crucial for reducing \distanceName without prior knowledge. Smaller visual stimuli (e.g., short bars) were particularly error-prone, suggesting that training protocols must account for mark size. Using our sampling regime, we have disentangled the effects of sampling strategies, emphasizing the need for careful experimentation to understand \cnns model behaviors. This understanding is critical when human observers rely on automated tools in scenarios involving human-AI collaboration---we need to know when to trust AIs and when not to.

Our research findings have yielded new insights into these black-box \cnnn' abilities to interpret charts. \begin{itemize}
    \setlength\itemsep{0em}
    \item
    Quantification Method: it is more informative to report case-by-case accuracy and beyond-average performance metrics. Notably, even in \mini conditions, in-sample data still performed well.
    \item
    Efficiency: Prioritize diverse sampling over large datasets—COV requires $50\%$ fewer samples than \iid for comparable accuracy.
    \item
    Transparency:
    When deploying \cnnn in real-world visualization tasks, model cards should include details on sampling methods, such as data span and train-test distributions.
    \item
    Visual encoding: Sufficiently large visual marks may be necessary to reduce errors.
\end{itemize}

Beyond the recommendations we have made, we hope our sampling regime and the probing methods provide the community with a means to quantify behavior differences between humans and machines.

\ifdetail
\tochange{Complete the sentence.}
\fi

\ifdetail
\tochange{Complete the sentence.}
\fi

\acknowledgments{We thank Justin Talbot, Vidya Setlur, and Anushka Anand for sharing their results, which we have used for human-\cnns comparisons in~\autoref{sec:study.humanai}. We also thank all participants who volunteered in our online experiment for their time and effort.
\rvision{Finally, we thank the reviewers for their feedback, which helped us improve the quality of this paper.} }

\section*{Image Copyrights}

\noindent We, as authors, state that the plots in this paper are and remain under our own copyright, with the permission to be used here. We have also made them available at \osflink and \googleDriveLink.

\bibliographystyle{abbrv-doi-hyperref}
\bibliography{main}

\appendix \clearpage
\noindent\begin{minipage}{\columnwidth}
    \vspace{0.5cm}
    \makeatletter
    \centering
    \sffamily\LARGE\bfseries
    \mytitlea
    \\[1em]
    \large Additional material\\[1em]
    \makeatother
\end{minipage}
\vspace{0.2cm}

\section{Reproducibility \& Access to Code/Models/Data}

In this Appendix, we report experimental details for human and \cnns experiments, reported in this paper. Experiment results and code can be accessed at \osflink or \googleDriveLink. The human behavioral dataset contains 1,575 trials across 63 observers for the ratio sampling experiment, and 4,903 trials across 50 participants for the height sampling experiment.

\section{A Comparison of Sampling Input Distributions and Data Processing Approaches}
\label{sm.samplingComparison}

The sampling values used in the corresponding experiments are marked in ~\autoref{fig:ratioHeightPairs} and sampling values are chosen below.

\textbf{Cleveland and McGill}~\cite{cleveland1984graphical} (\cmX in~\autoref{fig:ratioHeightPairs}). Seven ratios from a geometric sequence evenly spaced on a log scale of 10 bar heights $H_{i}={round}(10\times10^{(i-1)/12})$, where $i=1,..,10$ ($1 0, 12.1, 14.7, 17.8, 21.5, 26.1, 31.6, 38.3, 46.4, 56.2$). The 45 combinations of those 10 heights led to 9 distinct ratios (0.18, 0.22, 0.26, 0.32, 0.38, 0.46, 0.56, 0.68, 0.83). 0.22 and 0.32 were not used because they were too close to 0.18 and 0.26.

\textbf{Heer and Bostock.~\cite{heer2010crowdsourcing}} used the following ratios: 0.18, 0.22, 0.26, 0.32, 0.47, 0.55, 0.56, 0.69, and 0.85, with unknown bar height choices.

\textbf{Talbot et al.~\cite{talbot2014four}} (\talbotX in~\autoref{fig:ratioHeightPairs}) drew bar charts on a 200px$\times$200px canvas. Their three taller bar heights were 50px, 100px, and 150px. Ratios were the same 7 ratios as Cleveland and McGill's~\cite{cleveland1984graphical}: 0.178, 0.215 0.261, 0.316, 0.383, 0.464, 0.562, 0.682, 0.825.

\textbf{Haehn et al.~\cite{haehn2018evaluating}} (\haehnX in~\autoref{fig:ratioHeightPairs}). Used the 10 integer bar heights (10, 12, 15, 18, 21, 26, 32, 38, 46, 56). The simulation of Haehn et al.'s 12 exemplar runs in ~\autoref{fig:samplingRatioDH}.

\textbf{Our replication} also used the 10 integer heights (10, 12, 15, 18, 21, 26, 32, 38, 46, 56). We also removed height combinations [(10, 46), (12, 56)] and [(10, 32), (12, 38), (15, 46), (18, 56)], which will produce ratios 0.18 and 0.32 in the original Cleveland and McGill's experiment if the heights were not rounded. \haehnX in~\autoref{fig:ratioHeightPairs} except for the excluded combinations near ratio 0.18 and 0.32 were used in our experiment.

\section{\cnns Model Configuration}
\label{sec:modelConfig} We trained both \vgg and \resnet on multiple NVIDIA P100 GPUs, using CUDA 11.2, implemented using TensorFlow 2.4 and Python 3.9. We reused the model configuration in Haehn et al.'s~\cite{haehn2018evaluating}. We briefly describe the configuration here to make this paper self-containable. We used \vgg or \resnet as the feature extractor, followed by a one-hidden-layer multilayer perceptron (MLP) for regression. The hidden layer had 256 neurons, with ReLU as activation function and 0.5 dropout to prevent overfitting. We trained \cnnn from scratch for at most 100 epochs using mini-batch stochastic gradient descent (SGD) with a batch size of 32. Training stopped early when validation loss did not improve for ten consecutive epochs. The learning rate was 0.0001 and we set Nesterov momentum~\cite{Sutskever2013OnTI} to 0.9 to accelerate convergence by leveraging past gradients and reducing oscillations, making updates smoother and more stable. During training, we used mean square error (MSE) as the loss function.

\section{Additional Study Details}
\label{sec:sm_addition}

\subsection{\rvision{Architectural Influences on Model Accuracy}}
\label{sec:smVGGResNetComparison} In ~\autoref{sec:s1result} we found that \vgg outperformed \resnet in terms of accuracy. This disparity may stem from \vgg's architecture. With its regular stacking of convolutional layers, is well-suited for capturing the local patterns and structural features in the images, such as the bar heights and the marked dot. In contrast, ResNet50's deeper architecture and residual connections, while powerful for semantic information, might introduce unnecessary complexity or dilute the focus on localized features in this simpler setting.

\rvision{We also observed a small bar height effect in~\autoref{sec:s1result}, which may be explained by how \cnnn are implemented. CNNs (\eg, \vgg) are optimized for classification tasks, which use pooling layers to reduce spatial resolution. While this helps with object recognition by creating abstract features, it causes crucial information from small features, like short bars, to be lost in deeper layers~\cite{liu2021survey}. }

\subsection{Additional Details about the Human Experiment}
\label{sec:smExploratory}

\paragraph{Ratio and Height Sampling}
\label{sec:smHumanDataPreperation} We used the same 10 heights as Cleveland and McGill, rounded to the nearest integer: 10, 12, 15, 18, 21, 26, 32, 38, 46, and 56. In Cleveland and McGill's original experiment, the 45 combinations of these 10 heights yielded 9 ratios, of which two ratios (0.22 and 0.32) were not used. In our experiment, since the heights were rounded, we obtained 9 ratio groups, which was the same as Haehn et al. shown in \autoref{fig:othersSamplingExample} in the main text. Unlike Haehn et al., we also excluded the ratio groups corresponding to the ratios 0.22 and 0.32 that Cleveland and McGill omitted, preserving 39 height pairs in 7 ratio groups.

\paragraph{Image Generation and Rendering} We drew bar charts on 100$\times$100 pixel resolution saved in a vector format to ensure proper display at various screen resolutions on participants' devices. The minimal monitor resolution allowed for the experiment is 720P. The image height was scaled to either one-third of the browser window height or 300 pixels, whichever was larger.

\paragraph{Experiment Procedure and Instruction} Prior to beginning the experiment, participants filled in a demographic questionnaire, and their gender, age, education level, and nationality were collected. Participants were trained on the ratio estimation task. To familiarize participants with the task requirements, they completed 5 practice trials, one for each chart type, using randomly generated heights distinct from the 10 heights used in formal experiment. The practice trial interface is illustrated in \autoref{fig:labinthewide_ui}. During practice, participants were given immediate feedback after making their judgments. In the formal experiment, for each chart type, the 5 trials were randomly selected from the 7 possible ratio groups. Each participant completed 25 trials, consisting of 5 trials for each of the five bar chart types, in random order. The trials were presented with all chart types intermixed. No feedback was provided during the formal experiment. Panning and zooming were not allowed. Participants filled in their answers in the blank, and clicked the ``Next'' button to proceed to the next task.

\paragraph{Outlier Removal} Seven out of 63 participants were outliers and excluded from the study, as their errors exceeded three times the interquartile range.

\subsection{Additional Configuration Details for Reuse}
\label{sec:cell_node_count}

We measured whether the effect of \distanceName was generalizable (main text \autoref{sec:discussionTrainTestDistance}). We introduced two
\rvision{visualization types: continuous color and texture and node-links for two new counting tasks}: cell counting and node counting, where data were generated in high-fidelity simulations. We selected these two tasks not only because they represent classic problems in biology and network analyses, \rvision{but also because they utilize new visualization types distinct from the position/length in ratio estimation task: texture atop of shapes and node-links.}
\noindent
\rvision{Akin to sample ratios for bar charts, here we sampled cell counts and node counts.} The cell counts ranged from 80 to 159 (\autoref{fig:SM.IL.stimuli_example_cell}), and the node count from 20 to 99 (\autoref{fig:SM.IL.stimuli_example_node}), both yielding 80 unique samples. We followed the same proportion configuration used in the ratio estimation task to configure the training, validation, and test sets: 20-20-30$\%$ (16-16-24 samples) for test-validation-training, after applying the four sampling methods to sample the sets. The downsampling experiment reduced the number of unique training samples to 15\%, 7.5\%, and 3.75\%, or 12, 6, and 3 samples. In both cell and node counting experiments, we generate 6K/2K/2K images for training, validation, and test sets, respectively.

\setcounter{figure}{21}
\begin{figure*}

    \vspace{25pt}
    \centering
    \begin{tikzpicture}
        \node[anchor=south west, inner sep=0] (image) at (0,0) {\includegraphics[trim={0 0 0 0},clip, width=\textwidth]{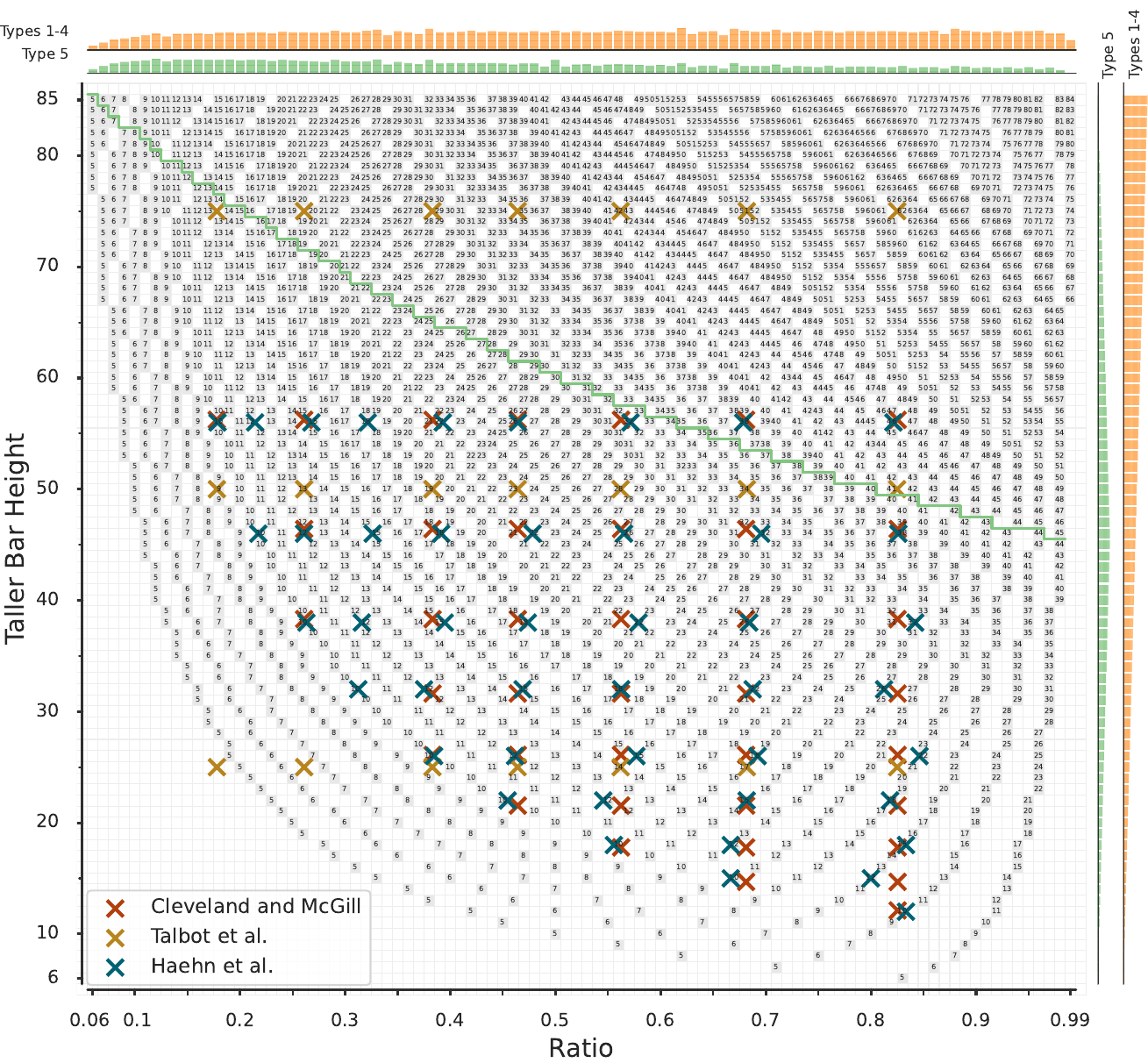}};
        \begin{scope}[x={(image.south east)}, y={(image.north west)}]
            \draw[red, thick] (0.406,0.07) rectangle (0.416,0.915);
            \draw[cyan, thick] (0.08,0.1795) rectangle (0.947,0.190);
            \draw[draw={rgb,255:red,100; green,100; blue,100}, thick] (0.837,0.775) rectangle (0.847,0.785);
            \draw[->, draw={rgb,255:red,100; green,100; blue,100}, thick] (0.847,0.785) to[out=45, in=-45] (0.925,1.) node[anchor=west] {};
        \end{scope}
    \end{tikzpicture}

    \vspace{-517pt}
    \raggedleft
    \begin{subfigure} {0.15\textwidth}
        \includegraphics[width=40pt, height=40pt]{ 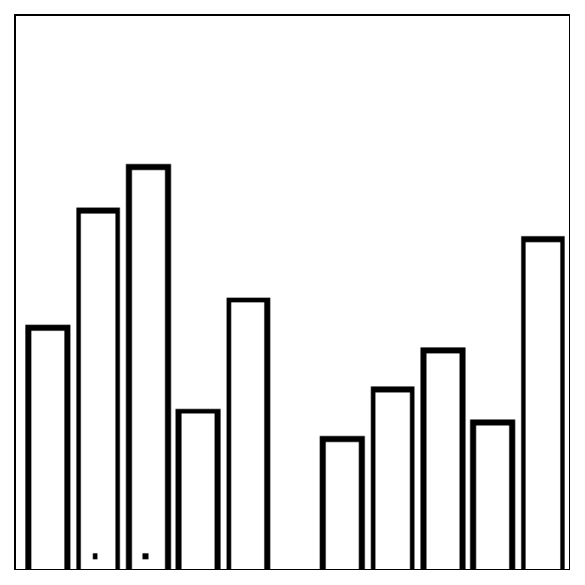 }
        \put(-42, 43){\footnotesize $r=0.89=\frac{65(h)}{73(H)}$}
    \end{subfigure}
    \vspace{467pt}

    \caption{\textbf{A comparison of sampling configurations for bar chart ratio estimates given a chart image resolution of $100\times100$ \rvision{(see ``Rendering Images'' in \autoref{sec:samplingRegimeMethod}}). }
    \textbf{(1) Sampling domain.} the numbers inside of \grayCell are shorter-bar heights $h$, the y-axis represents taller-bar heights $H$, and ratios on the x-axis = $h / H$, where $h\in[5, 84]$ and $H\in[6, 85]$. As a result, there are 80 unique integer $hs$ and 80 unique integer $Hs$. All pairs of integer heights $\in[5, 85]$ yield $C^{2}_{81}=3,240$ combinations or gray cells \grayCell, which represent the sampling domain.
    \textbf{(2) Chart type constraints}. Types 1 to 4 (adjacent, aligned stacked, separated, and unaligned stacked bars) of Cleveland and McGill charts can use the entire sampling domain. In contrast, Type 5 (divided bars) can only use the sampling domain under \customLabelZigzag{77C482}, because the two target bars are stacked on the same side limiting the heights to that $h+H\leq100$.
    \textbf{(3) Experimental conditions}. Clevaland and McGill~\cite{cleveland1984graphical} used 7 \textit{ratios} from 39 $(h, H)$ \cmX pairs; Talbot et al.~\cite{talbot2014four} used three \textit{human test bar heights} from 21 $(h, H)$ \talbotX pairs; Haehn et al.~\cite{haehn2018evaluating} used 43 integer $(h, H)$ \haehnX pairs for \cnnn' test and training sets.
    \textbf{(4) Sampling bins}. Our experiments used the whole domain. For sampling purposes, these ratios are grouped into 94 ratio bins ($rb$) and $rb\in[0.06, 0.99]$. Each column grid is a unique ratio bin and each row grid is a unique integer bar height. A ratio bin contains many ($h, H$) pairs. For example, $rb=0.42$ can be generated from ($h, H$)=(19, 45), (16, 38), and others inside of \redVerticalBox. Each taller bar height can also contain a few ratios. For example, $H=16$ can produced $rb=0.5$, $rb=0.62$ and two others \cyanHorizontalBox. }
    \label{fig:ratioHeightPairs}
\end{figure*}

\begin{figure*}[!thp]
    \centering
    \includegraphics[width=0.9\textwidth]{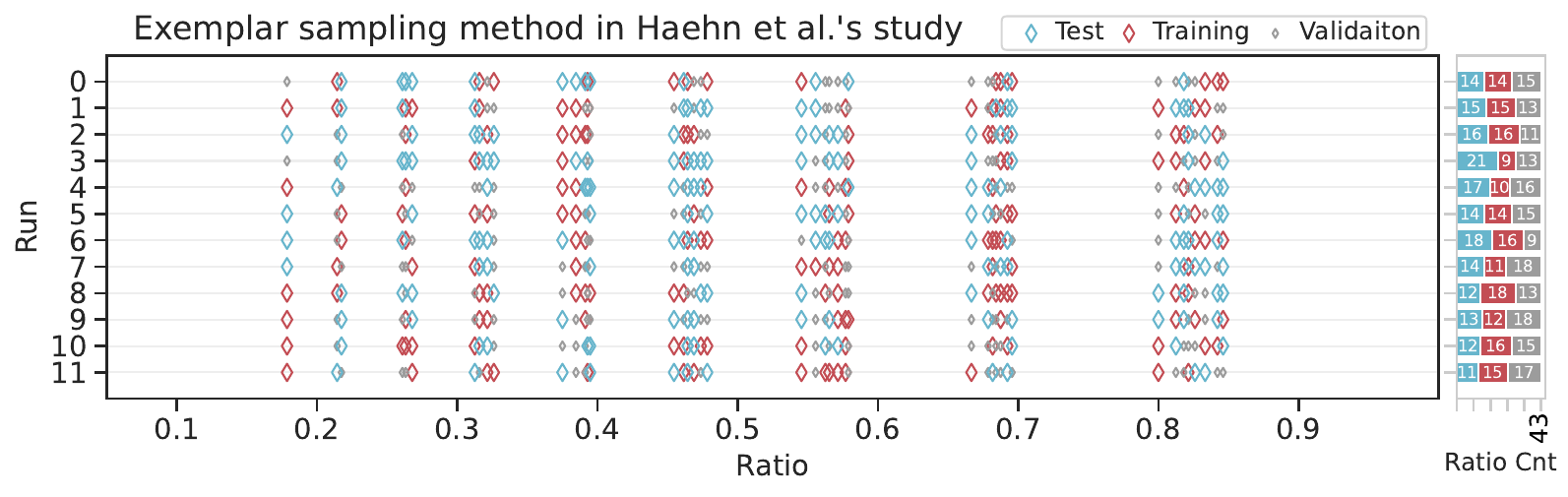}
    \caption{ {\textbf{Twelve exemplar sampling distributions of training, validation, and test sets in Haehn et al.~\cite{haehn2018evaluating}.}
    \rvision{\textbf{Observation.} the proportion of training samples could differ between runs (see main text \autoref{sec:relatedwork_sampling}). } }}
    \label{fig:samplingRatioDH}
\end{figure*}

\begin{figure*}[!t]
    \centering
    \begin{subfigure}{0.8\textwidth}
        \includegraphics[width=\textwidth]{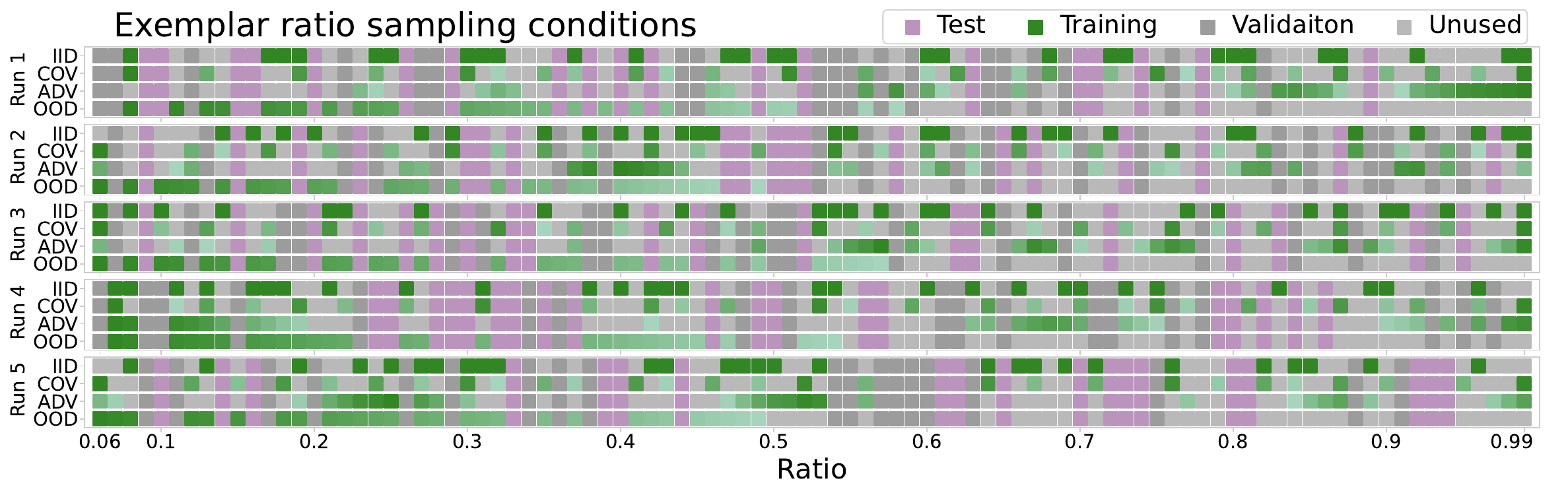}
        \put(-250, 0){\footnotesize (a)}
        \label{fig:robusness3RunsCNNRatio}
    \end{subfigure}

    \begin{subfigure}{0.8\textwidth}
        \includegraphics[width=\textwidth]{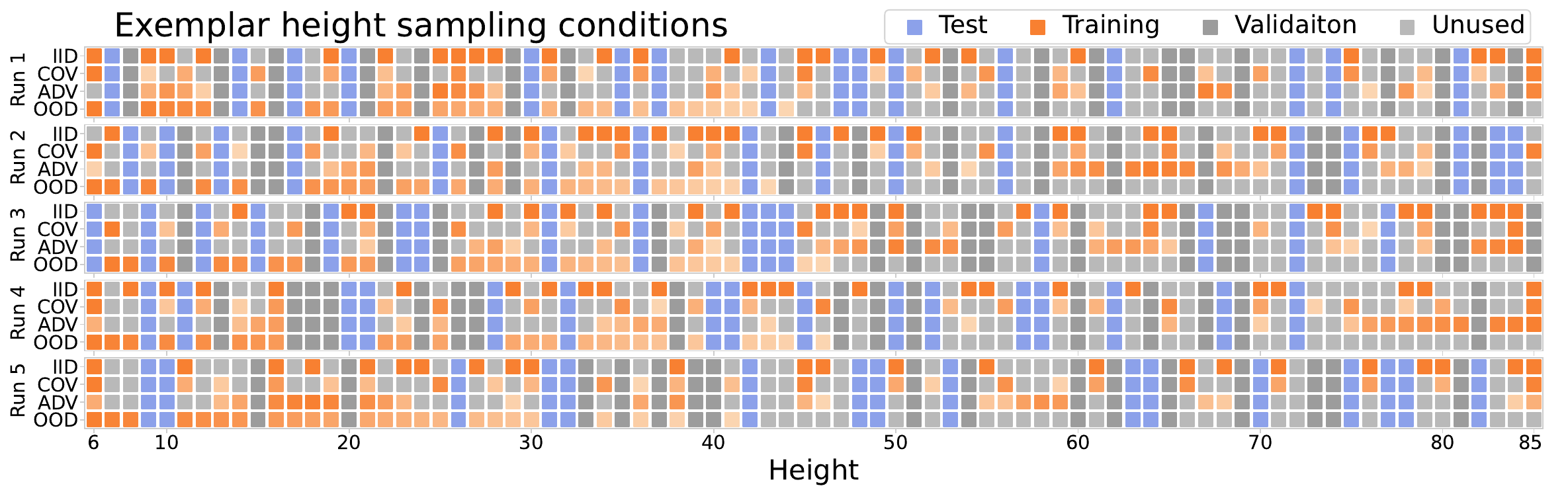}
        \put(-250, 0){\footnotesize (b)}
    \end{subfigure}
    \caption{\textbf{\sititle.} (a) Ratio bins and (b) height samples used for our Study I \rvision{(see main text \autoref{sec:study.outofdistribution})}. }
    \label{fig:robusness3RunsCNN}
\end{figure*}

\begin{figure*}[!t]
    \centering
    \includegraphics[width=0.8\textwidth]{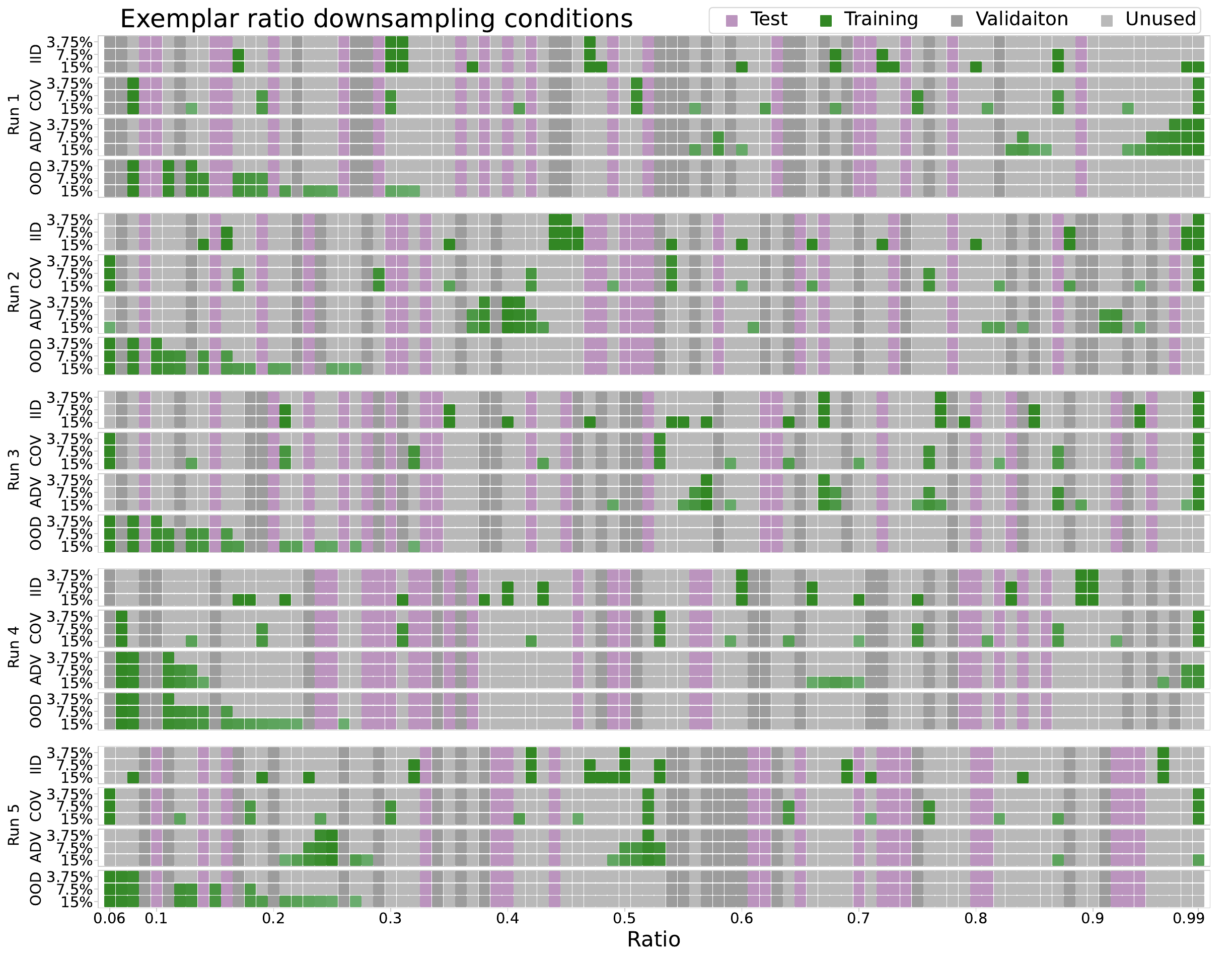}
    \put(-250, 4){\footnotesize \circledchar{a}}

    \includegraphics[width=0.8\textwidth]{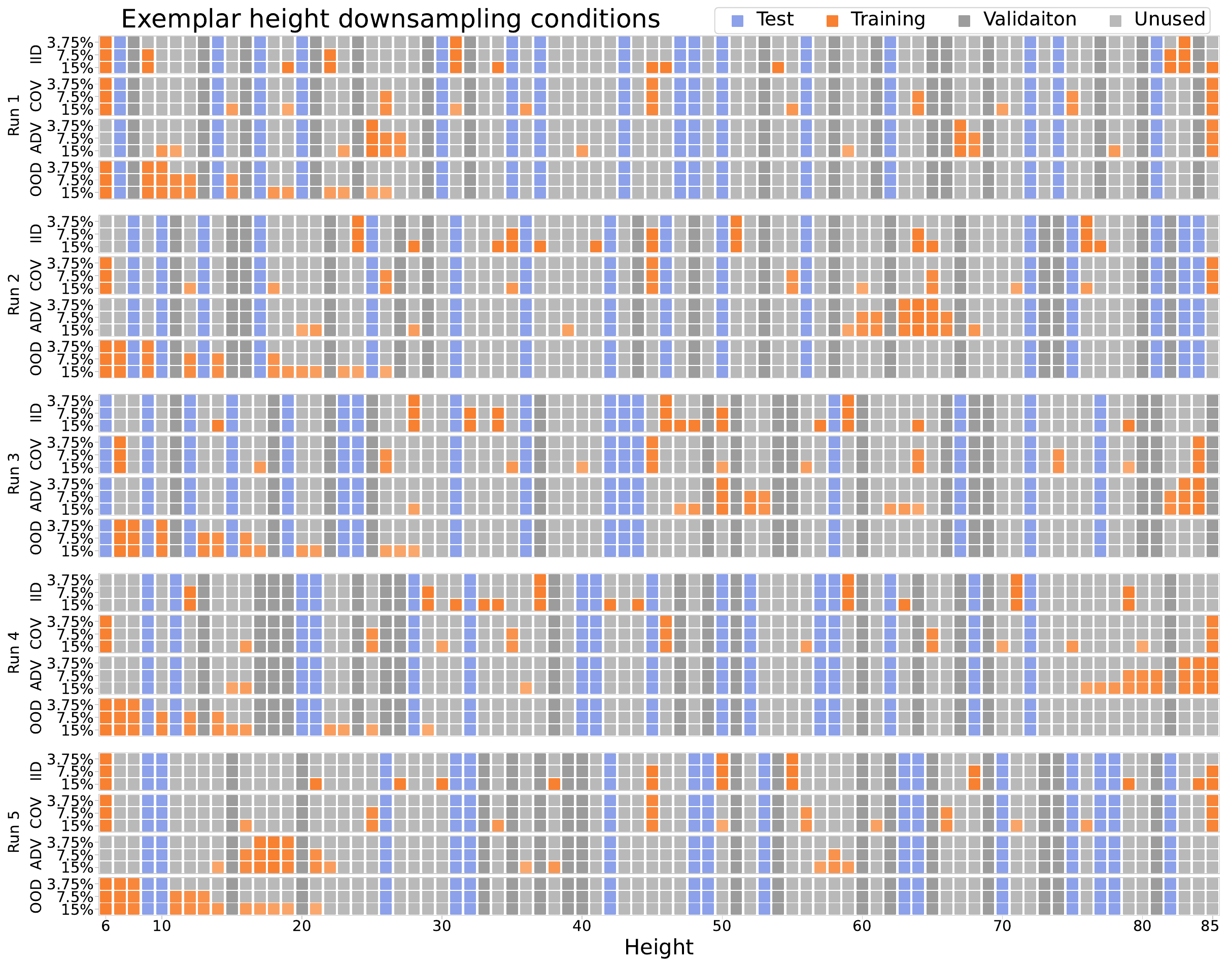}
    \put(-250, 4){\footnotesize\circledchar{b}}
    \caption{
    \textbf{\siititle.} \circledchar{a} Ratio bin and \circledchar{b} height samples used for Study II \rvision{(see main text \autoref{sec:study.stability})}.
    \textbf{\ul{Observations.}} (1) The data span for \cov remained stable across all downsampling. In contrast; (2) \adv had a large reduction in data span compared to the other methods, resulting in sampling conditions closer to \mini; (3) Similarly, \iid assembled \mini when using fewer samples. }
    \label{fig:stability3RunsCNN}
\end{figure*}

\begin{figure*}[!tb]
    \centering
    \vspace{-30px}
    \hspace{-30px}
    \begin{subfigure}[H]{0.4\columnwidth}
        \includegraphics[height=127px]{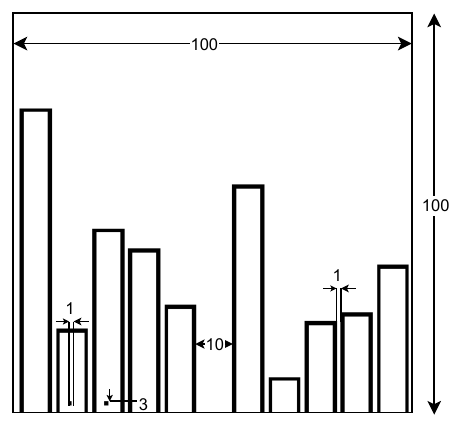}
    \end{subfigure}
    \hspace{120px}
    \begin{subfigure}[H]{0.4\columnwidth}
        \includegraphics[height=120px]{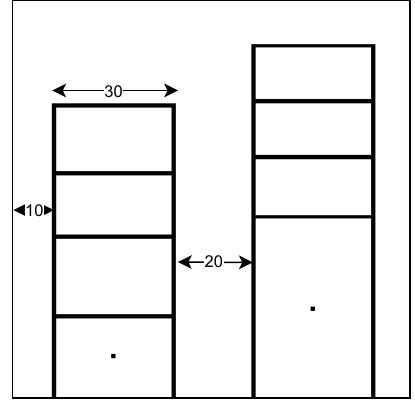}
    \end{subfigure}
    \caption{\rvision{\textbf{Bar Chart Appearance Design} (see main text \autoref{sec:samplingRegimeMethod}). \textit{Left.} Grouped bar charts (Types 1, 3): each bar has a width of 8 pixels. There's a 1-pixel gap between bars within the same group, and an 10-pixel gap separating different groups. A 1$\times$1 pixel dot marks the target bar, positioned 3 pixels from the bottom and horizontally centered. \textit{Right.} Stacked bar charts (Types 2, 4, 5): each bar has a width of 30 pixels. The gap between stacked groups is 20 pixels. The target bar is indicated by a $1\times1$ pixel dot placed at its center.}}
    \vspace{-25px}
    \label{fig:bar_appearance}
\end{figure*}

\begin{figure*}[!tb]
    \centering
    \includegraphics[width=0.75\textwidth]{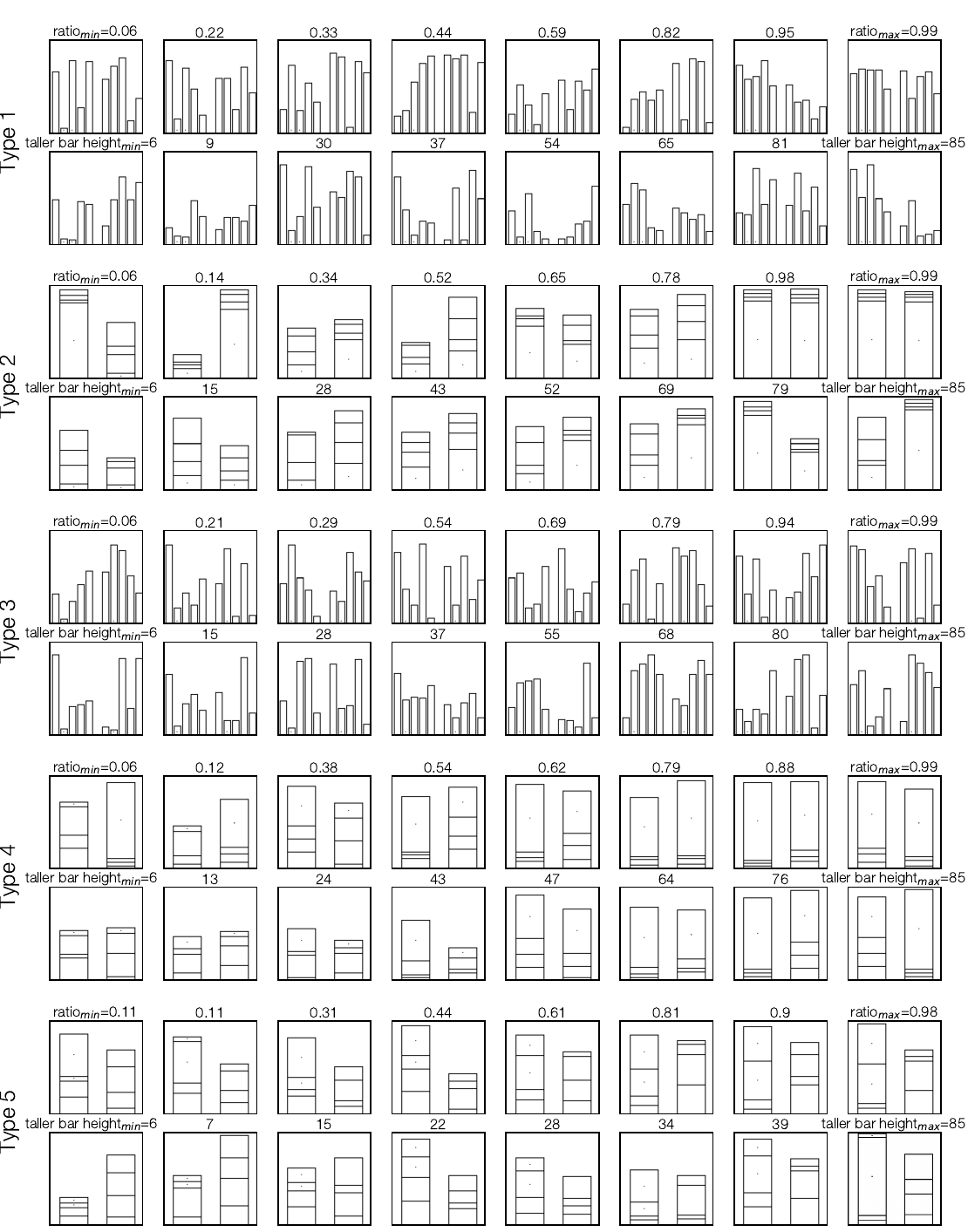}
    \caption{\textbf{Types 1-5 bar chart examples. } Here we show some examples of different ratio and height samples generated in the experiments. \rvision{(See the ``Rendering Images'' paragraph in \autoref{sec:samplingRegimeMethod}}). The black border was not rendered when the images were sent to the \cnnn. }
    \label{fig:morebars}
\end{figure*}

\setcounter{table}{5}
\begin{figure*}
    \centering

    \includegraphics[width=0.9\textwidth]{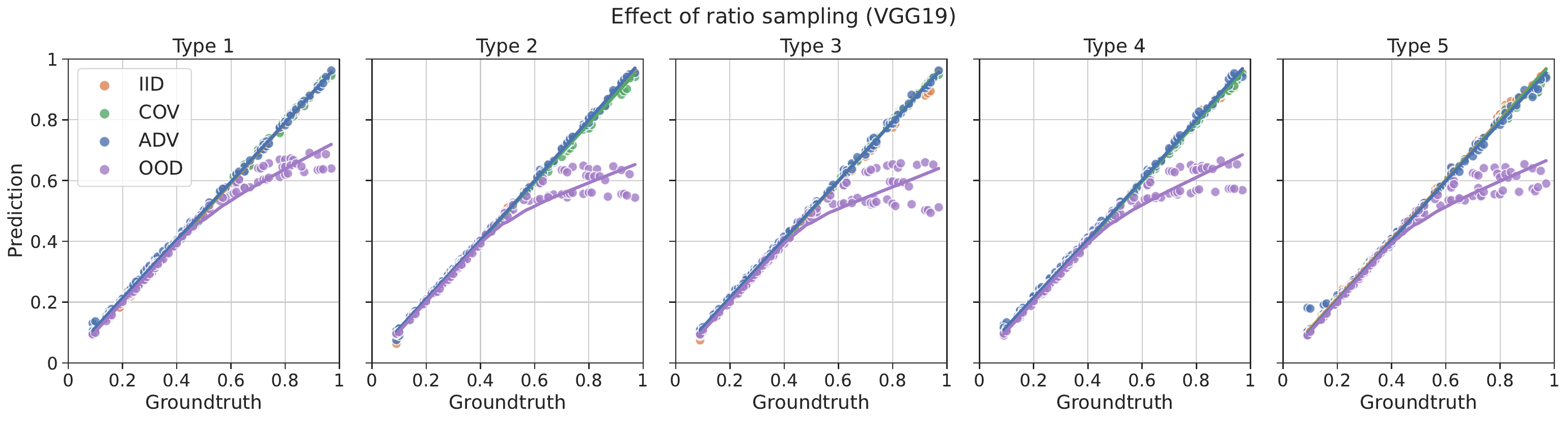}
    \includegraphics[width=0.9\textwidth]{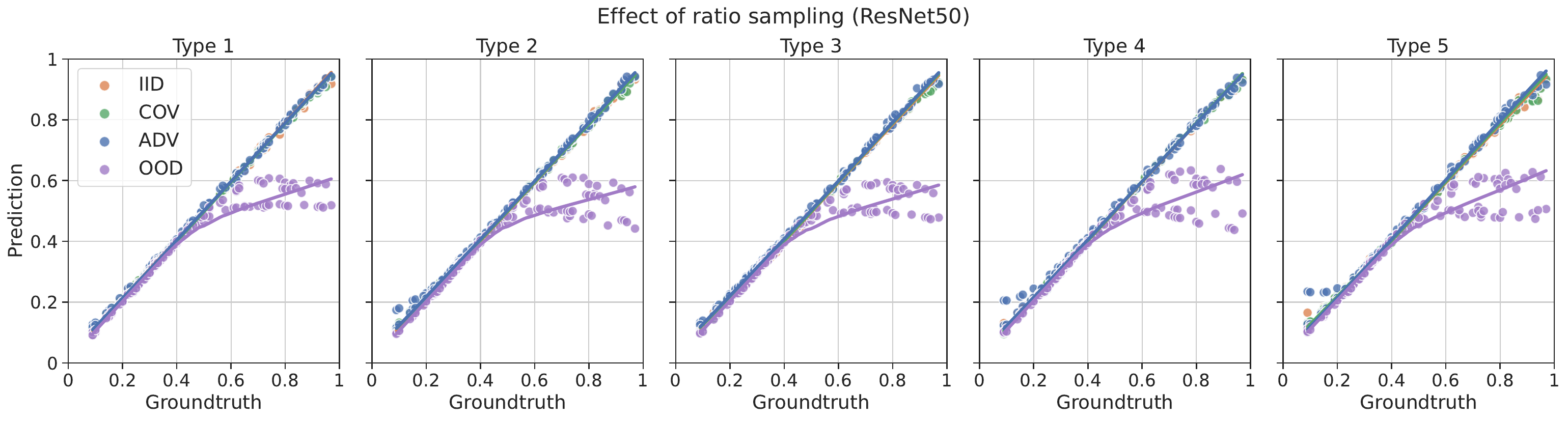}

    \includegraphics[width=0.9\textwidth]{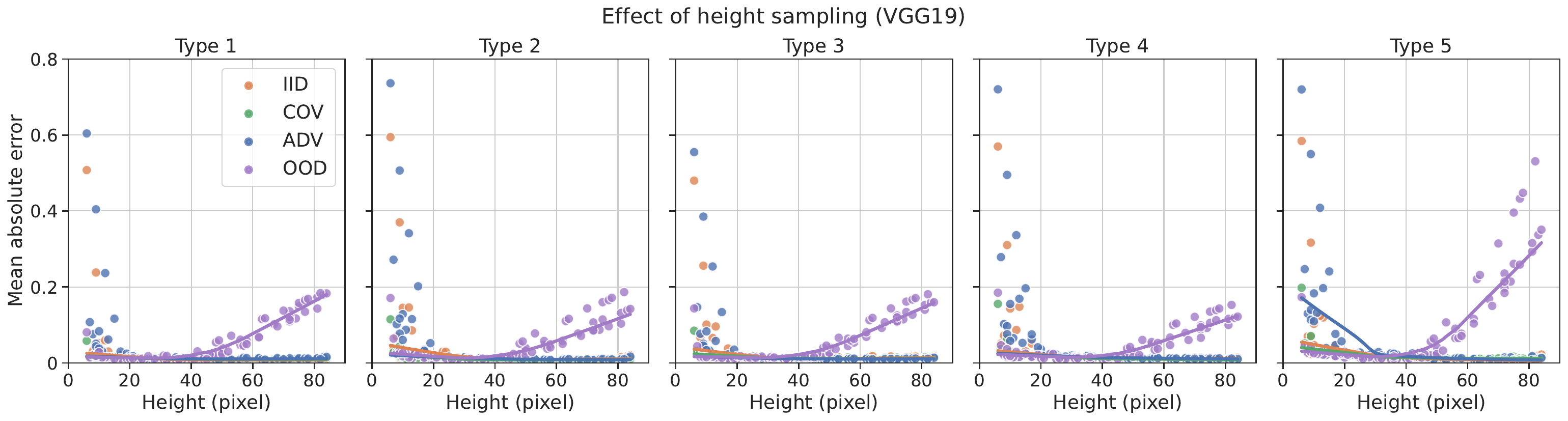}
    \includegraphics[width=0.9\textwidth]{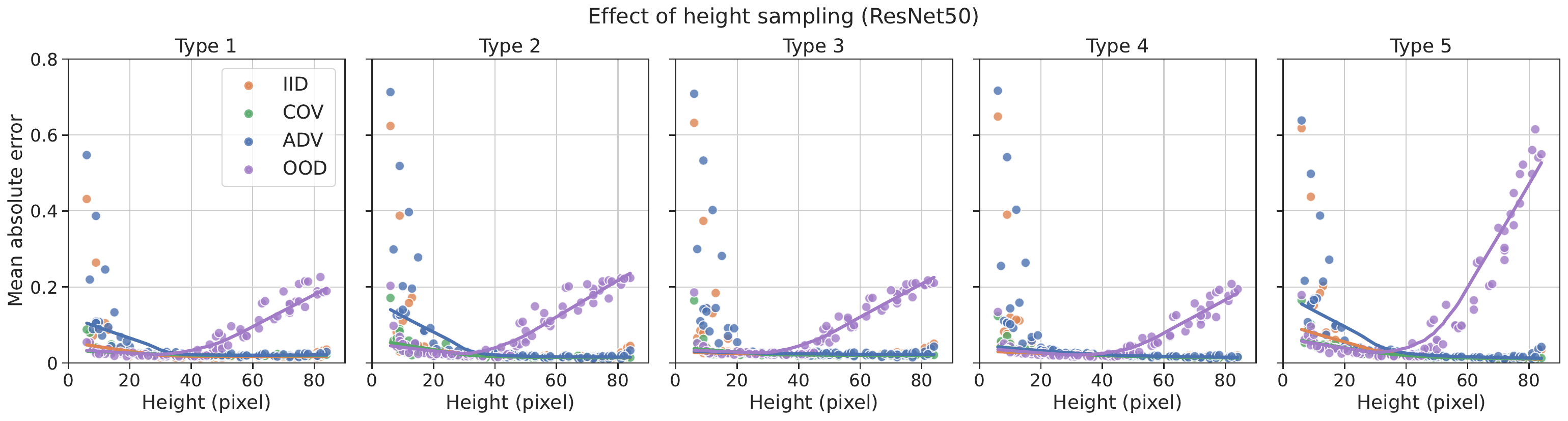}
    \caption{\textbf{\sititle.} Case-by-case analysis additional results (see main text~\autoref{sec:s1result}). \textbf{\ul{Observations.}} (1) \vgg and \resnet had similar behaviors. (2) Smaller bar heights led to larger inference errors. }
    \label{fig:sm_sampling}
\end{figure*}

\begin{table*} [!t]
    \caption{\textbf{\sititle.} Summary Statistics. }
    \label{tab:sm_study1_stat}
    \resizebox{\textwidth}{!}{ \begin{tabular}{@{}lllllllllll@{}}
        &  & \multicolumn{4}{c}{\vgg} &                         & \multicolumn{4}{c}{\resnet}     \\
        \cmidrule(lr){3-6} \cmidrule(l){8-11} Experiment (effect)   &  & \multicolumn{1}{c}{$F$}  & \multicolumn{1}{c}{$p$} & \multicolumn{1}{c}{$\eta^{2}$} & \multicolumn{1}{c}{HSD}               & \multicolumn{1}{c}{} & \multicolumn{1}{c}{$F$} & \multicolumn{1}{c}{$p$} & \multicolumn{1}{c}{$\eta^{2}$} & \multicolumn{1}{c}{HSD}                     \\
        \midrule \multicolumn{11}{l}{Study I Exp 1: Ratio sampling}  \\
        Type (all sampling)                                         &  & $F_{(4, 2049)}= 0.9$     & $p = 0.438$             & $\eta^{2} < 0.01$              & $\mathrm{(1, 2, 3, 4, 5)}$            &                      & $F_{(4, 2049)}= 0.6$    & $p = 0.650$             & $\eta^{2} < 0.01$              & $\mathrm{(1, 2, 3, 4, 5)}$                  \\
        Type (IID)                                                  &  & $F_{(4, 509)}= 9.0$      & $p < \mathbf{0.001}$    & $\eta^{2} = \mathit{0.07}$     & $\mathrm{(1, 2, 4)>(1, 3, 4)>(3, 5)}$ &                      & $F_{(4, 509)}= 11.2$    & $p < \mathbf{0.001}$    & $\eta^{2} = \mathit{0.08}$     & $\mathrm{(1, 2)>(1, 4)>(3, 4)>(3, 5)}$      \\
        Type (COV)                                                  &  & $F_{(4, 509)}= 2.1$      & $p = 0.076$             & $\eta^{2} = 0.02$              & $\mathrm{(1, 2, 3, 4, 5)}$            &                      & $F_{(4, 509)}= 9.0$     & $p < \mathbf{0.001}$    & $\eta^{2} = \mathit{0.07}$     & $\mathrm{(1, 2, 4)>(3, 4)>(3, 5)}$          \\
        Type (ADV)                                                  &  & $F_{(4, 510)}= 9.3$      & $p < \mathbf{0.001}$    & $\eta^{2} = \mathit{0.07}$     & $\mathrm{(1, 2)>(1, 3, 4)>(3, 4, 5)}$ &                      & $F_{(4, 510)}= 10.6$    & $p < \mathbf{0.001}$    & $\eta^{2} = \mathit{0.08}$     & $\mathrm{(1, 2, 3)>(3, 4)>(4, 5)}$          \\
        Type (OOD)                                                  &  & $F_{(4, 509)}= 0.6$      & $p = 0.653$             & $\eta^{2} < 0.01$              & $\mathrm{(1, 2, 3, 4, 5)}$            &                      & $F_{(4, 509)}= 0.1$     & $p = 0.995$             & $\eta^{2} < 0.01$              & $\mathrm{(1, 2, 3, 4, 5)}$                  \\
        Sampling (all types)                                        &  & $F_{(3, 2049)}= 174.7$   & $p < \mathbf{0.001}$    & $\eta^{2} = \mathbf{0.20}$     & $\mathrm{(COV, IID, ADV)>(OOD)}$      &                      & $F_{(3, 2049)}= 178.6$  & $p < \mathbf{0.001}$    & $\eta^{2} = \mathbf{0.21}$     & $\mathrm{(COV, IID, ADV)>(OOD)}$            \\
        Sampling (Type 1)                                           &  & $F_{(3, 408)}= 33.3$     & $p < \mathbf{0.001}$    & $\eta^{2} = \mathbf{0.20}$     & $\mathrm{(COV, IID, ADV)>(OOD)}$      &                      & $F_{(3, 408)}= 38.7$    & $p < \mathbf{0.001}$    & $\eta^{2} = \mathbf{0.22}$     & $\mathrm{(COV, IID, ADV)>(OOD)}$            \\
        Sampling (Type 2)                                           &  & $F_{(3, 408)}= 36.1$     & $p < \mathbf{0.001}$    & $\eta^{2} = \mathbf{0.21}$     & $\mathrm{(COV, IID, ADV)>(OOD)}$      &                      & $F_{(3, 408)}= 35.9$    & $p < \mathbf{0.001}$    & $\eta^{2} = \mathbf{0.21}$     & $\mathrm{(COV, IID, ADV)>(OOD)}$            \\
        Sampling (Type 3)                                           &  & $F_{(3, 408)}= 35.6$     & $p < \mathbf{0.001}$    & $\eta^{2} = \mathbf{0.21}$     & $\mathrm{(COV, IID, ADV)>(OOD)}$      &                      & $F_{(3, 408)}= 36.5$    & $p < \mathbf{0.001}$    & $\eta^{2} = \mathbf{0.21}$     & $\mathrm{(COV, IID, ADV)>(OOD)}$            \\
        Sampling (Type 4)                                           &  & $F_{(3, 408)}= 36.3$     & $p < \mathbf{0.001}$    & $\eta^{2} = \mathbf{0.21}$     & $\mathrm{(COV, IID, ADV)>(OOD)}$      &                      & $F_{(3, 408)}= 33.0$    & $p < \mathbf{0.001}$    & $\eta^{2} = \mathbf{0.20}$     & $\mathrm{(COV, IID, ADV)>(OOD)}$            \\
        Sampling (Type 5)                                           &  & $F_{(3, 405)}= 34.5$     & $p < \mathbf{0.001}$    & $\eta^{2} = \mathbf{0.20}$     & $\mathrm{(COV, IID, ADV)>(OOD)}$      &                      & $F_{(3, 405)}= 34.2$    & $p < \mathbf{0.001}$    & $\eta^{2} = \mathbf{0.20}$     & $\mathrm{(COV, IID, ADV)>(OOD)}$            \\
        Categorical (Type 1)                                        &  & $F_{(1, 1090)}= 10.4$    & $p < \mathbf{0.001}$    & $\eta^{2} < 0.01$              & $\mathrm{(tall)>(short)}$             &                      & $F_{(1, 1090)}= 10.8$   & $p < \mathbf{0.001}$    & $\eta^{2} = 0.01$              & $\mathrm{(tall)>(short)}$                   \\
        Categorical (Type 2)                                        &  & $F_{(1, 1090)}= 5.2$     & $p = \mathbf{0.023}$    & $\eta^{2} < 0.01$              & $\mathrm{(tall)>(short)}$             &                      & $F_{(1, 1090)}= 11.7$   & $p < \mathbf{0.001}$    & $\eta^{2} = 0.01$              & $\mathrm{(tall)>(short)}$                   \\
        Categorical (Type 3)                                        &  & $F_{(1, 1090)}= 2.3$     & $p = 0.129$             & $\eta^{2} < 0.01$              & $\mathrm{(tall, short)}$              &                      & $F_{(1, 1090)}= 6.0$    & $p = \mathbf{0.014}$    & $\eta^{2} < 0.01$              & $\mathrm{(tall)>(short)}$                   \\
        Categorical (Type 4)                                        &  & $F_{(1, 1090)}= 20.1$    & $p < \mathbf{0.001}$    & $\eta^{2} = 0.02$              & $\mathrm{(tall)>(short)}$             &                      & $F_{(1, 1090)}= 38.3$   & $p < \mathbf{0.001}$    & $\eta^{2} = 0.03$              & $\mathrm{(tall)>(short)}$                   \\
        Categorical (Type 5)                                        &  & $F_{(1, 1062)}= 18.8$    & $p < \mathbf{0.001}$    & $\eta^{2} = 0.02$              & $\mathrm{(tall)>(short)}$             &                      & $F_{(1, 1062)}= 31.2$   & $p < \mathbf{0.001}$    & $\eta^{2} = 0.03$              & $\mathrm{(tall)>(short)}$                   \\
        \midrule \multicolumn{11}{l}{Study I Exp 2: Height sampling} \\
        Type (all sampling)                                         &  & $F_{(4, 1592)}= 4.2$     & $p = \mathbf{0.002}$    & $\eta^{2} = 0.01$              & $\mathrm{(1, 2, 3, 4)>(4, 5)}$        &                      & $F_{(4, 1592)}= 4.4$    & $p = \mathbf{0.002}$    & $\eta^{2} = 0.01$              & $\mathrm{(1, 2, 3, 4)>(2, 3, 5)}$           \\
        Type (IID)                                                  &  & $F_{(4, 395)}= 0.3$      & $p = 0.882$             & $\eta^{2} < 0.01$              & $\mathrm{(1, 2, 3, 4, 5)}$            &                      & $F_{(4, 395)}= 0.3$     & $p = 0.872$             & $\eta^{2} < 0.01$              & $\mathrm{(1, 2, 3, 4, 5)}$                  \\
        Type (COV)                                                  &  & $F_{(4, 395)}= 3.6$      & $p = \mathbf{0.007}$    & $\eta^{2} = 0.04$              & $\mathrm{(1, 2, 3, 4)>(3, 4, 5)}$     &                      & $F_{(4, 395)}= 0.3$     & $p = 0.882$             & $\eta^{2} < 0.01$              & $\mathrm{(1, 2, 3, 4, 5)}$                  \\
        Type (ADV)                                                  &  & $F_{(4, 395)}= 0.7$      & $p = 0.608$             & $\eta^{2} < 0.01$              & $\mathrm{(1, 2, 3, 4, 5)}$            &                      & $F_{(4, 395)}= 0.3$     & $p = 0.896$             & $\eta^{2} < 0.01$              & $\mathrm{(1, 2, 3, 4, 5)}$                  \\
        Type (OOD)                                                  &  & $F_{(4, 395)}= 6.9$      & $p < \mathbf{0.001}$    & $\eta^{2} = \mathit{0.07}$     & $\mathrm{(1, 2, 3, 4)>(5)}$           &                      & $F_{(4, 395)}= 7.9$     & $p < \mathbf{0.001}$    & $\eta^{2} = \mathit{0.07}$     & $\mathrm{(1, 2, 3, 4)>(5)}$                 \\
        Sampling (all types)                                        &  & $F_{(3, 1592)}= 29.5$    & $p < \mathbf{0.001}$    & $\eta^{2} = 0.05$              & $\mathrm{(COV, IID)>(ADV)>(OOD)}$     &                      & $F_{(3, 1592)}= 43.7$   & $p < \mathbf{0.001}$    & $\eta^{2} = \mathit{0.07}$     & $\mathrm{(COV)>(IID)>(ADV)>(OOD)}$          \\
        Sampling (Type 1)                                           &  & $F_{(3, 316)}= 7.2$      & $p < \mathbf{0.001}$    & $\eta^{2} = \mathit{0.06}$     & $\mathrm{(COV, IID, ADV)>(OOD, ADV)}$ &                      & $F_{(3, 316)}= 9.8$     & $p < \mathbf{0.001}$    & $\eta^{2} = \mathit{0.09}$     & $\mathrm{(COV, IID, ADV)>(OOD, ADV)}$       \\
        Sampling (Type 2)                                           &  & $F_{(3, 316)}= 4.4$      & $p = \mathbf{0.005}$    & $\eta^{2} = 0.04$              & $\mathrm{(COV, IID)>(OOD, IID, ADV)}$ &                      & $F_{(3, 316)}= 9.2$     & $p < \mathbf{0.001}$    & $\eta^{2} = \mathit{0.08}$     & $\mathrm{(COV, IID)>(IID, ADV)>(OOD, ADV)}$ \\
        Sampling (Type 3)                                           &  & $F_{(3, 316)}= 6.6$      & $p < \mathbf{0.001}$    & $\eta^{2} = 0.06$              & $\mathrm{(COV, IID, ADV)>(OOD, ADV)}$ &                      & $F_{(3, 316)}= 8.3$     & $p < \mathbf{0.001}$    & $\eta^{2} = \mathit{0.07}$     & $\mathrm{(COV, IID)>(IID, ADV)>(OOD, ADV)}$ \\
        Sampling (Type 4)                                           &  & $F_{(3, 316)}= 3.9$      & $p = \mathbf{0.010}$    & $\eta^{2} = 0.04$              & $\mathrm{(COV, IID)>(OOD, IID, ADV)}$ &                      & $F_{(3, 316)}= 4.0$     & $p = \mathbf{0.008}$    & $\eta^{2} = 0.04$              & $\mathrm{(COV, IID)>(OOD, IID, ADV)}$       \\
        Sampling (Type 5)                                           &  & $F_{(3, 316)}= 10.7$     & $p < \mathbf{0.001}$    & $\eta^{2} = \mathit{0.09}$     & $\mathrm{(COV, IID, ADV)>(OOD)}$      &                      & $F_{(3, 316)}= 16.2$    & $p < \mathbf{0.001}$    & $\eta^{2} = \mathit{0.13}$     & $\mathrm{(COV, IID, ADV)>(OOD)}$            \\
        Categorical (Type 1)                                        &  & $F_{(1, 318)}= 24.5$     & $p < \mathbf{0.001}$    & $\eta^{2} = \mathit{0.07}$     & $\mathrm{(tall)>(short)}$             &                      & $F_{(1, 318)}= 33.7$    & $p < \mathbf{0.001}$    & $\eta^{2} = \mathit{0.10}$     & $\mathrm{(tall)>(short)}$                   \\
        Categorical (Type 2)                                        &  & $F_{(1, 318)}= 60.5$     & $p < \mathbf{0.001}$    & $\eta^{2} = \mathbf{0.16}$     & $\mathrm{(tall)>(short)}$             &                      & $F_{(1, 318)}= 61.0$    & $p < \mathbf{0.001}$    & $\eta^{2} = \mathbf{0.16}$     & $\mathrm{(tall)>(short)}$                   \\
        Categorical (Type 3)                                        &  & $F_{(1, 318)}= 35.4$     & $p < \mathbf{0.001}$    & $\eta^{2} = \mathit{0.10}$     & $\mathrm{(tall)>(short)}$             &                      & $F_{(1, 318)}= 46.9$    & $p < \mathbf{0.001}$    & $\eta^{2} = \mathit{0.13}$     & $\mathrm{(tall)>(short)}$                   \\
        Categorical (Type 4)                                        &  & $F_{(1, 318)}= 68.4$     & $p < \mathbf{0.001}$    & $\eta^{2} = \mathbf{0.18}$     & $\mathrm{(tall)>(short)}$             &                      & $F_{(1, 318)}= 64.4$    & $p < \mathbf{0.001}$    & $\eta^{2} = \mathbf{0.17}$     & $\mathrm{(tall)>(short)}$                   \\
        Categorical (Type 5)                                        &  & $F_{(1, 318)}= 29.5$     & $p < \mathbf{0.001}$    & $\eta^{2} = \mathit{0.09}$     & $\mathrm{(tall)>(short)}$             &                      & $F_{(1, 318)}= 23.4$    & $p < \mathbf{0.001}$    & $\eta^{2} = \mathit{0.07}$     & $\mathrm{(tall)>(short)}$                   \\
        \bottomrule
    \end{tabular} }
\end{table*}

\begin{figure*}[!t]
    \centering
    \includegraphics[height=125pt]{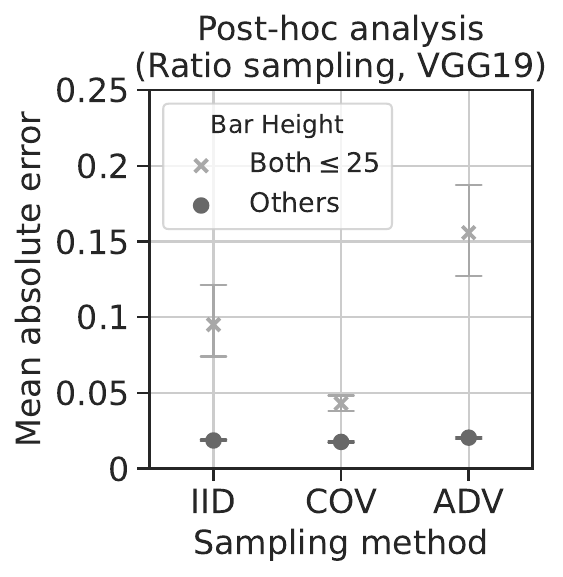}
    \includegraphics[height=125pt]{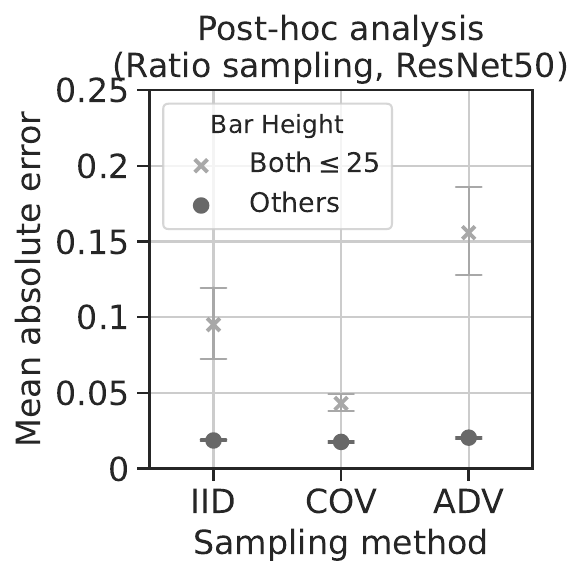}
    \includegraphics[height=125pt]{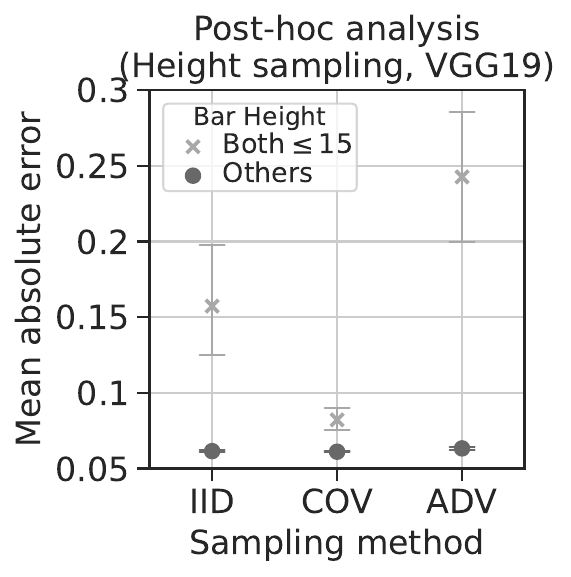}
    \includegraphics[height=125pt]{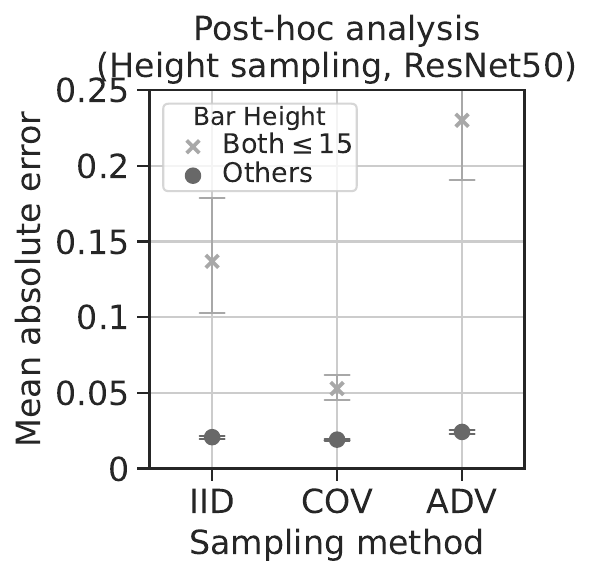}
    \caption{\textbf{\sititle.} Effect of short bar heights. Errors when both bar heights were small vs. other cases \rvision{(see main text \autoref{sec:s1result})}. \textbf{\ul{Observations.}} The error increased when both marked bars' heights were smaller or equal to 15 pixels in height sampling, and when both heights are smaller or equal to 25 pixels in ratio sampling. }
    \label{fig:ratio_sampling_categorical}
\end{figure*}

\begin{figure*}
    \centering
    \begin{subfigure} [T]{0.90\textwidth}
        \includegraphics[width=\textwidth]{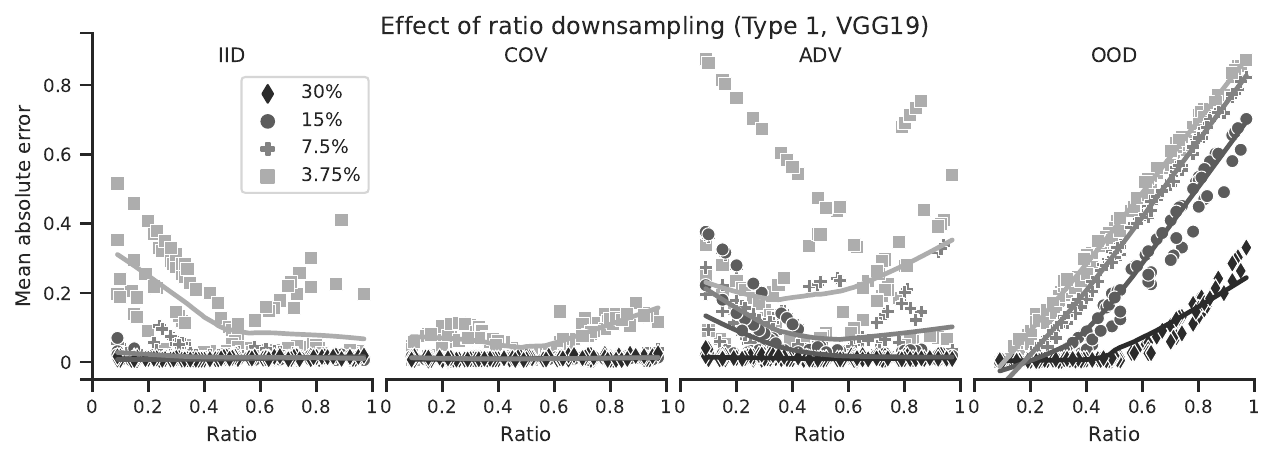}
    \end{subfigure}

    \begin{subfigure} [T]{0.90\textwidth}
        \includegraphics[width=\textwidth]{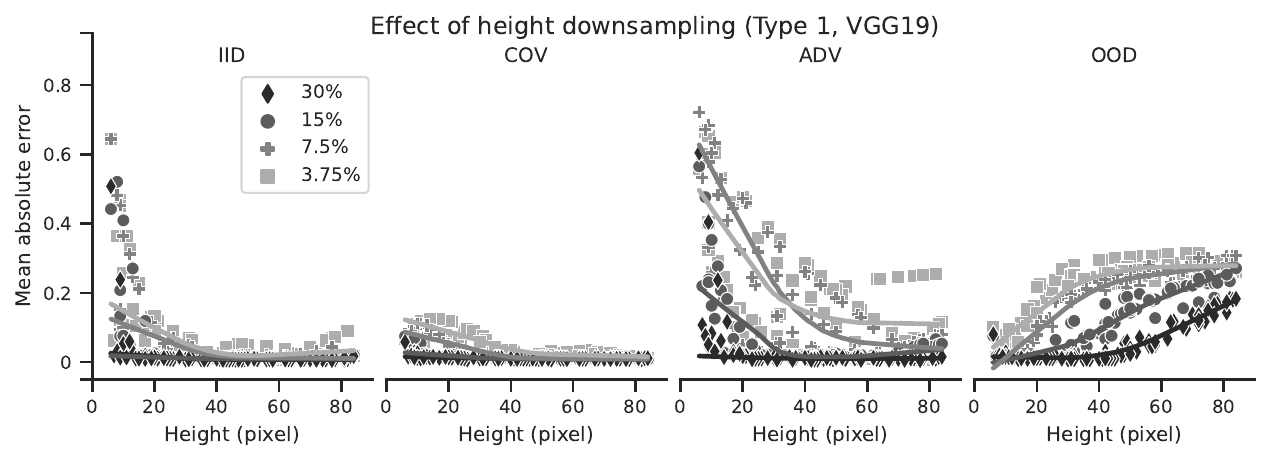}
    \end{subfigure}

    \caption{\textbf{\siititle.} Additional Results. Ratio downsampling results using \vgg~(see main text \autoref{sec:study.stability}). \textbf{\ul{Observations.}} Reducing the number of samples lowered the prediction accuracy for all sampling methods, but \cov was most robust to the downsampling. }
    \label{fig:sm_ratio_downsampling_vgg}
\end{figure*}

\begin{figure*}
    \centering

    \includegraphics[width=\textwidth]{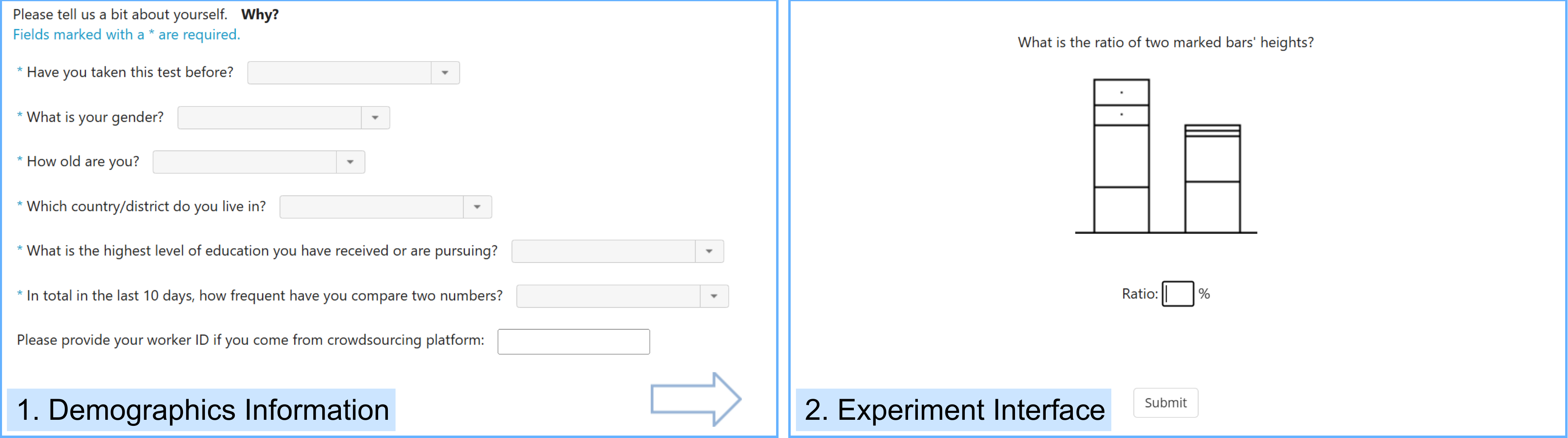}
    \caption{
    \rvision{\textbf{\shumanaititle.}} Human ratio experiment user interface \rvision{(see main text \autoref{sec:human_data_collect})}. Participants were shown bar charts of five types and were required to fill in the percentage ratio of the shorter to the taller marked bar heights. After a valid number between 1 and 99 was filled in, participants can click the ``Submit'' button to move on to the next trial. }
    \label{fig:labinthewide_ui}
\end{figure*}

\begin{figure*}[!htp]
    \centering
    \includegraphics[height=82 px]{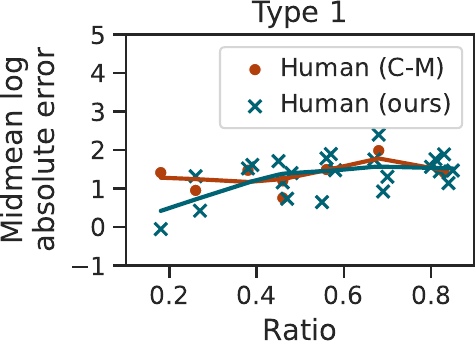}
    \includegraphics[height=82 px, trim={30 0 0 0}, clip]{ 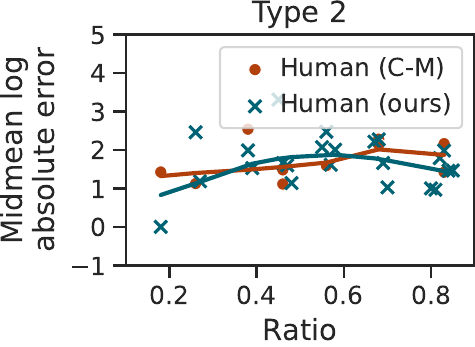 }
    \includegraphics[height=82 px, trim={30 0 0 0}, clip]{ 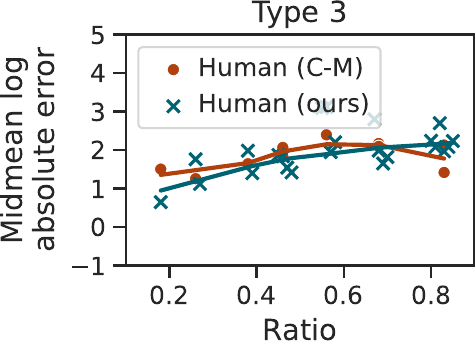 }
    \includegraphics[height=82 px, trim={30 0 0 0}, clip]{ 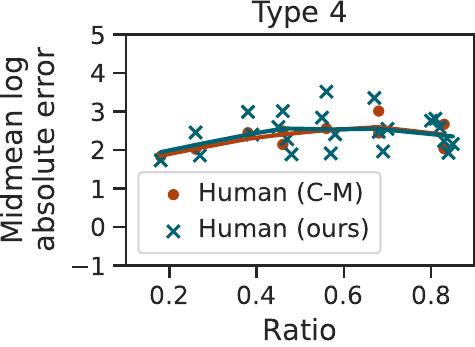 }
    \includegraphics[height=82 px, trim={30 0 0 0}, clip]{ 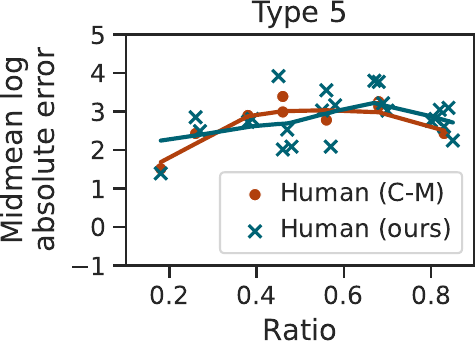 }
    \caption{
    \rvision{\textbf{\shumanaititle.}} Comparison of our position-length experimental results to that of Cleveland and McGill~\cite{cleveland1984graphical} \rvision{(see main text \autoref{sec:human.results})}. The C-M data is a copy of Figure 13 from their position-length experiment because their results are not in the public domain. We used midmean log errors for comparison purposes. The line in each figure depicted the LOWESS regression curves~\cite{Cleveland1979RobustLW} from the data collected in Cleveland and McGill~\cite{cleveland1984graphical} and ours.
    \textbf{\ul{Observations.}} Our human study results aligned well with those of Cleveland and McGill. }
    \label{fig:cmLITWCompare}
\end{figure*}

\begin{figure*}
    \centering
    \includegraphics[width=0.9\textwidth]{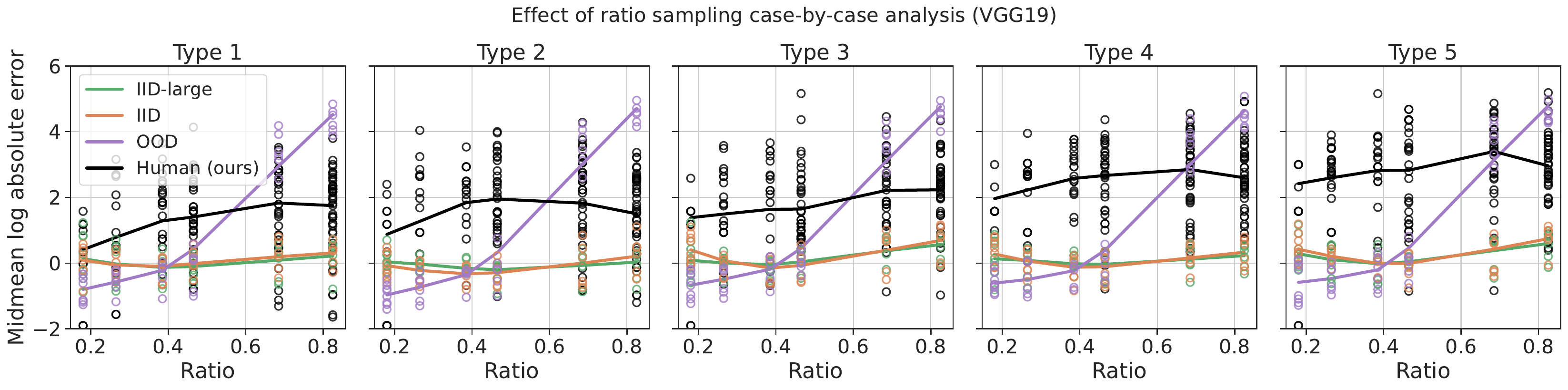}
    \includegraphics[width=0.9\textwidth]{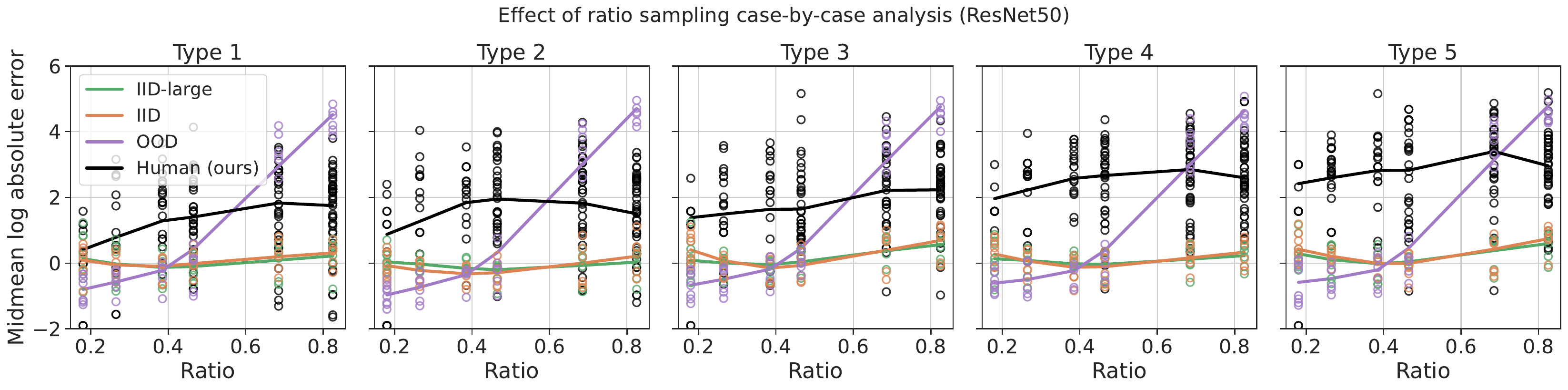}

    \includegraphics[width=0.9\textwidth]{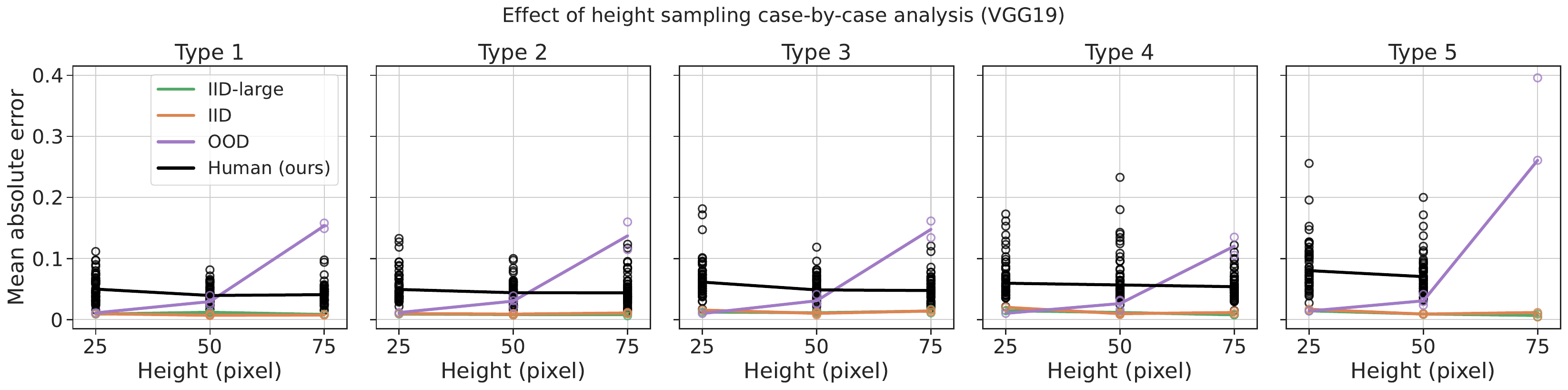}
    \includegraphics[width=0.9\textwidth]{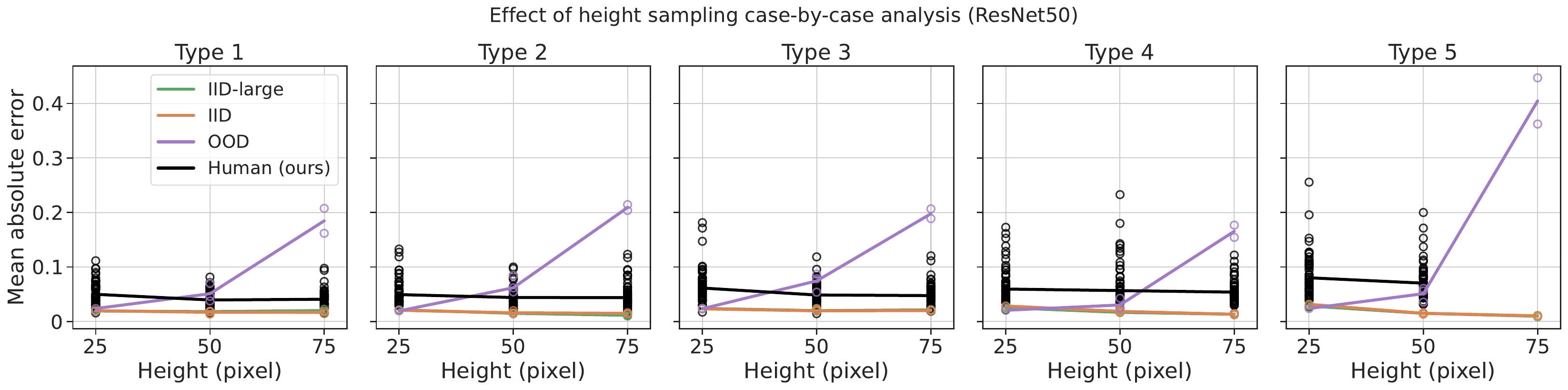}
    \caption{\textbf{\shumanaititle.} Case-by-case analysis of inference errors between humans and \cnnn. \textbf{\ul{Observations.}} \cnnn had lower error than humans except for the \oodmodel's out-of-distribution places. \cnnn' errors were more centered than humans.
    \cnnn had lower errors and narrower error spread compared to humans, except for the out-of-distribution values. }
    \label{fig:smStudyHumanDetail}
\end{figure*}

\usetikzlibrary{calc}
\usetikzlibrary{fit}
\usetikzlibrary{shapes}
\tikzset{boximg/.style={remember picture,red,thick,draw,inner sep=0pt,outer sep=0pt}}

\begin{figure*}[!t]
    \centering
    \begin{subfigure}[T]{\columnwidth}
        \includegraphics[width=1.1\columnwidth]{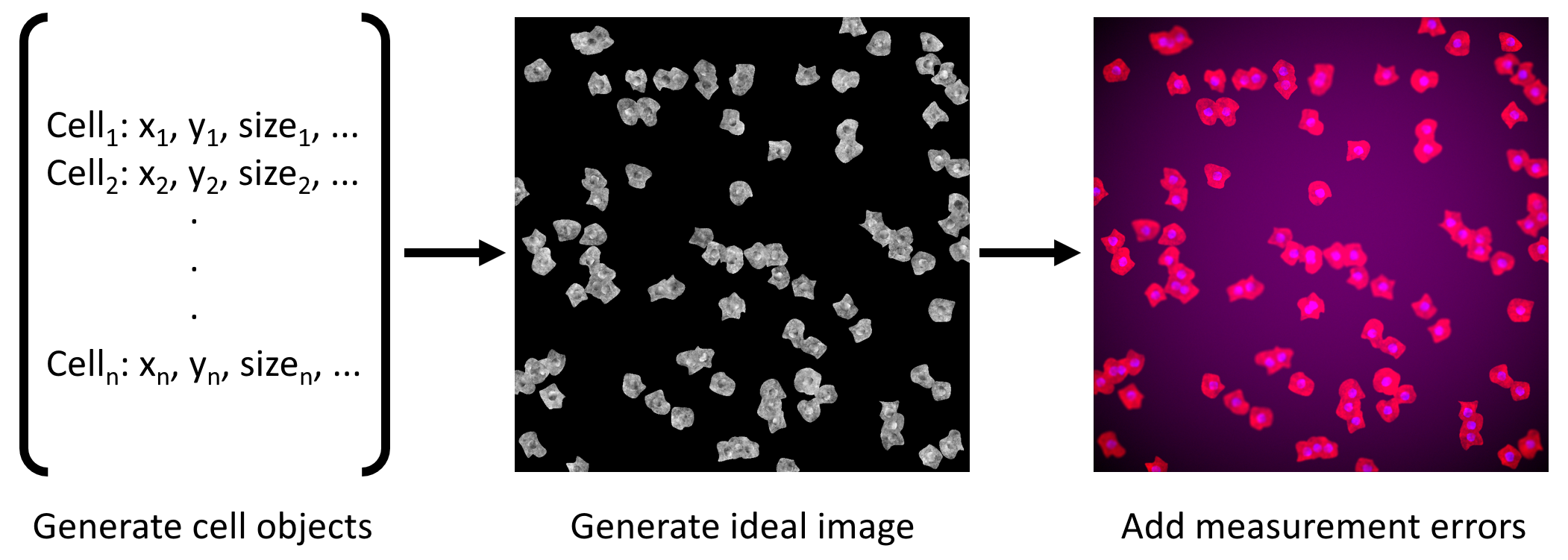}
    \end{subfigure}
    \hfill
    \centering
    \begin{subfigure}[T]{0.33\columnwidth}
        \begin{tikzpicture}[boximg]
            \node[anchor=south west] (img) {\includegraphics[width=\columnwidth]{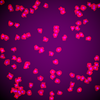}};
            \draw[stealth-, very thick,green] (0.8,1.6) -- ++(0.4,0.6) node[align=center,above,white]{\small Background\\\small Illumination};
            \begin{scope}[x=(img.south east),y=(img.north west)]
                \node [draw,green,thick,fit={(.63,.08) (1,.45)}](Box){};
            \end{scope}
        \end{tikzpicture}
    \end{subfigure}
    \hspace{15pt}
    \begin{subfigure}[T]{0.33\columnwidth}
        \begin{tikzpicture}[boximg]
            \node[anchor=south west] (zoom) {\includegraphics[width=\columnwidth]{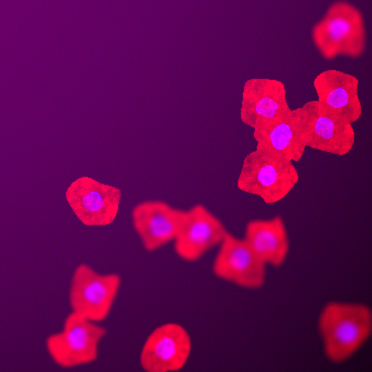}};
            \draw[green,ultra thick] (zoom.south west) rectangle ([yshift=0cm,xshift=0cm]zoom.north east);

            \draw[stealth-, very thick,green] (0.8,1.6) -- ++(0.4,0.6) node[align=center,above,white]{\small Cytoplasm with\\\small unique texture};
            \draw[stealth-, very thick,green] (0.8,1.4) -- ++(0.0,-0.6) node[align=center,below,white]{\small Nuclei};
            \draw[stealth-, very thick,green] (2,0.7) -- ++(0.0,-0.3) node[align=center,below,white ]{\small Blurry cells};
            \draw[stealth-, very thick,green] (2.4,1.75) -- ++(0.0,-0.3) node[align=center,below,white]{\small Overlaps};
        \end{tikzpicture}
    \end{subfigure}

    \begin{tikzpicture}[overlay,boximg]
        \draw[green,thick] (Box.north east) -- ([yshift=0cm] zoom.north west);
        \draw[green,thick] (Box.south east) -- ([yshift=0cm,xshift=0cm] zoom.south west);
    \end{tikzpicture}

    \caption{\textbf{Process of cell image generation and image texture explanation \rvision{ (see the main text \autoref{sec:discussionTrainTestDistance})}.} Left: three steps of cell image generation 1) generate cell objects, 2) generate ideal image, 3) colorize add measurement errors. Right: texture explanation.}
    \label{fig:cell_generation}
\end{figure*}

\begin{figure*}[!htp]
    \centering
    \includegraphics[width=0.9\textwidth]{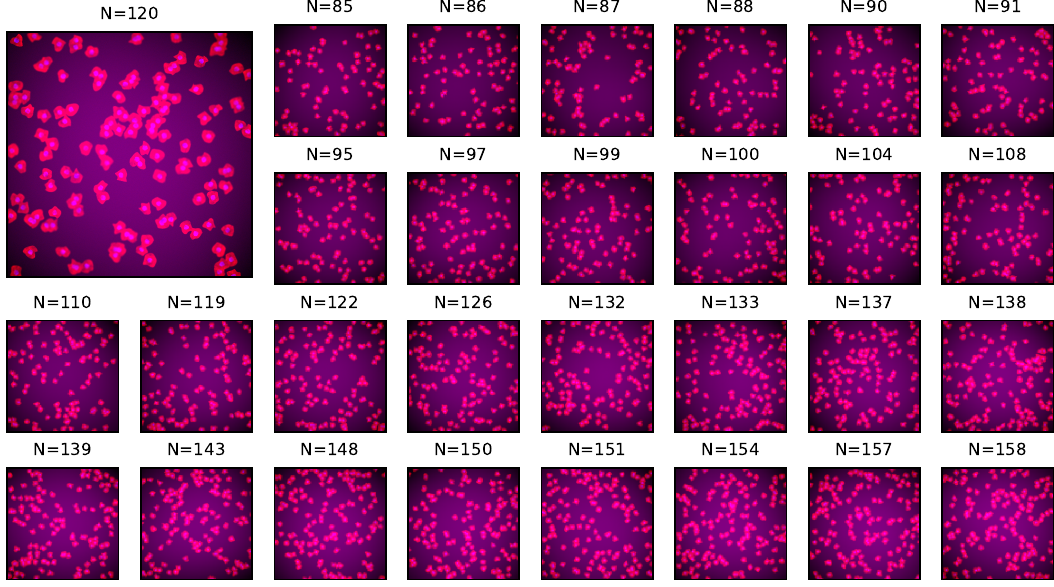}
    \caption{Examples of simulated cell image with fluorescently labeled cytoplasm \rvision{for cell counting task, which requires perceiving textures and shapes of cells (see the main text \autoref{sec:discussionTrainTestDistance}). Continuous color and textures visualization type in fluorescence-light microscopy images were synthesized using SIMCEP~\cite{lehmussola2007computational} at a resolution of 1000$\times$1000 pixels and then downscaled to 100$\times$100 pixels. The simulation accurately replicates diverse cell morphologies with varying shapes and different staining conditions, creating a multichannel representation of individual cells and their intracellular activities. The underlying cell shapes are randomly generated using parametric models, with positions randomly distributed throughout the field. While cytoplasmic overlap is permitted in the simulation, nuclei overlap is prohibited. The simulation also incorporates realistic compression artifacts, introducing blur that mimics actual microscopy imaging conditions. } }
    \label{fig:SM.IL.stimuli_example_cell}
\end{figure*}
\begin{figure*}[!htp]
    \centering
    \includegraphics[width=0.9\textwidth]{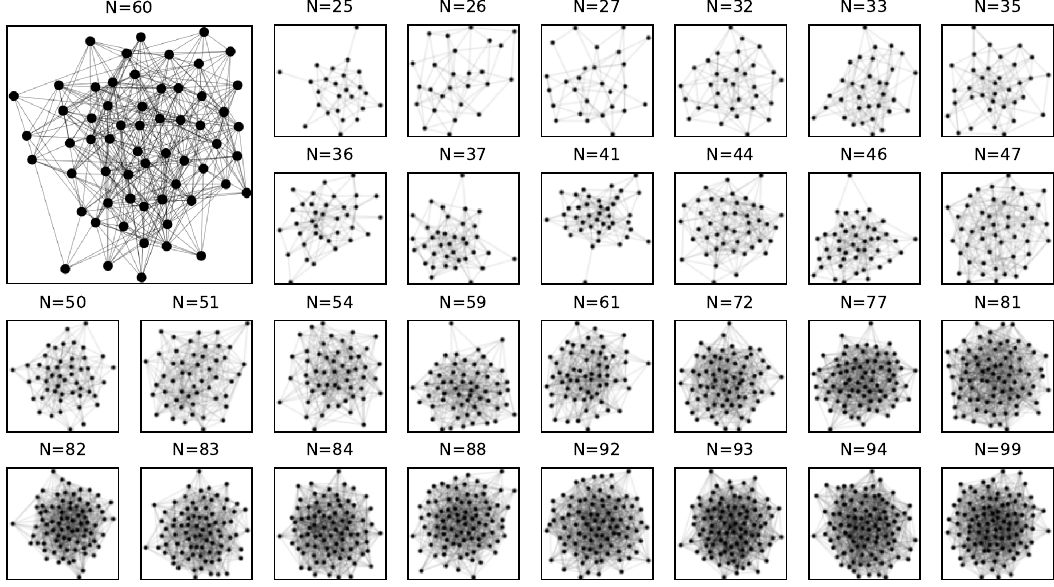}
    \caption{Examples of node link networks that show community structures \rvision{for node counting task, which requires perceiving shapes in a graph structure (see the main text \autoref{sec:discussionTrainTestDistance}). Node-links visualization type in node-link diagrams were generated using Lancichinetti et al.'s benchmark algorithm~\cite{lancichinetti2009benchmarks} at a resolution of 100$\times$100 pixels. The visualization type is node-link diagrams which use point locations and links to show structures in data. This generation method simulates complex hierarchical organizational systems, where the elementary parts of the system and their mutual interactions are nodes and links, respectively. The node density ranges from 0.3 to 0.34. Node positions are determined by a force-directed layout, creating visually distinct community structures. Each node is connected to one or more neighbors with no self-loops or duplicate edges, and edges are rendered with minimal width to emphasize node patterns. } }
    \label{fig:SM.IL.stimuli_example_node}
\end{figure*}

\begin{figure*}
    \includegraphics[width=\textwidth]{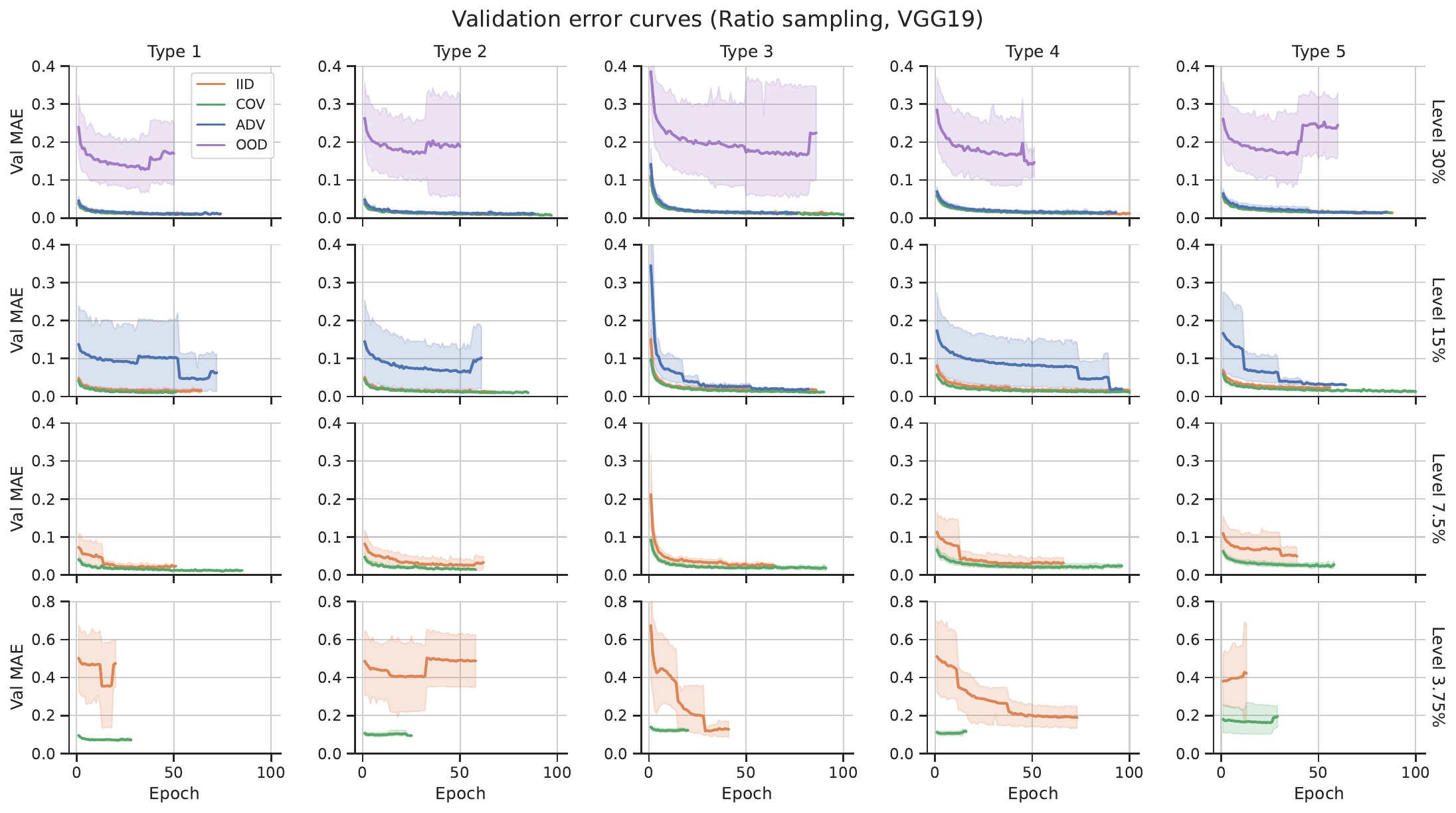}
    \includegraphics[width=\textwidth]{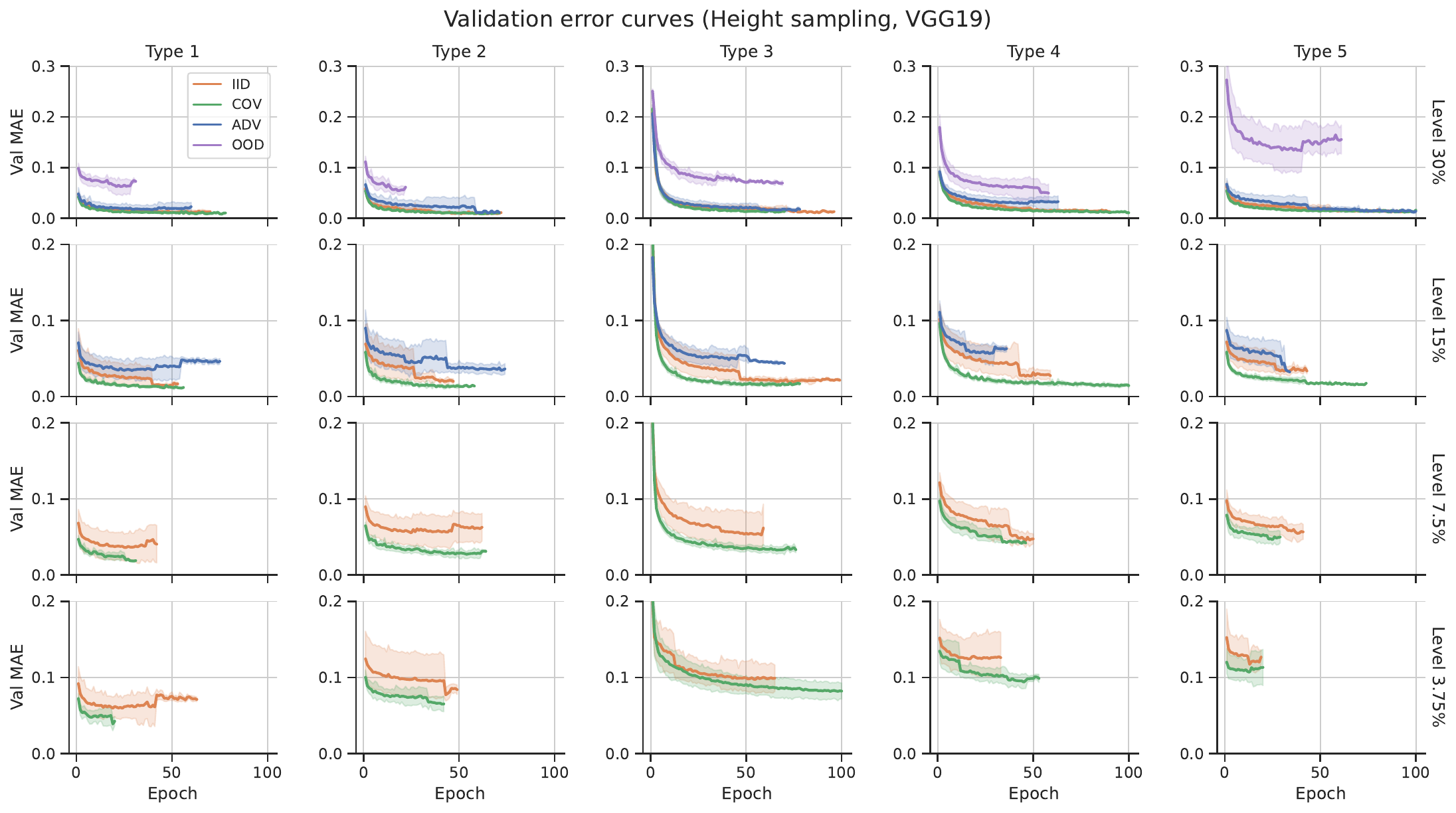}
    \caption{\rvision{\textbf{Validation error as a function of the training epochs for the ratio and height sampling experiments.} This figure analyzes the training efficiency of sampling methods (see main text \autoref{sec:discussion_data_prep}). The performance of the \iid and \cov methods are compared against the \mini and \adv methods. Due to significantly larger error magnitudes, the \mini method is not shown at sampling levels of 15\%, 7.5\%, and 3.75\%, nor is the \adv method at levels of 7.5\% and 3.75\%. Each plotted line represents the mean of five experimental runs, with the corresponding 95\% confidence interval shaded. Abrupt changes in the error curves may appear, which is an artifact of certain runs stopping early.
    \textbf{\ul{Observations.}} (1) In both experiments, \cov converges faster than all other sampling methods, especially at level 7.5\% and 3.75\%. To reach a comparable performance level, models trained with \cov required drastically less training time---on average, 69.7\%, 82.6\%, and 96.4\% less time compared to \iid, \adv, and \mini, respectively. (2) As the number of training samples is reduced, we observe that it is generally harder for models to learn and they tend to stop training earlier. At the 30\% sampling level, it takes an average of 67.04 epochs to train a model, while at the 3.75\% level, this number drops to 29.25---less than half of the original. This indicates that with severely limited data, models fail to generalize effectively, causing the validation error to plateau much sooner and trigger early stopping. }}
    \label{fig:training_history}
\end{figure*}

\end{document}